\documentclass[12pt]{article}
\usepackage[utf8]{inputenc}
\usepackage{booktabs}
\usepackage{mathrsfs}
\usepackage{url}
\usepackage{amsmath}
\usepackage{amsthm}
\usepackage{amssymb}
\usepackage{graphicx}
\usepackage{setspace}
\usepackage{chapterbib}
\usepackage[authoryear]{natbib}
\usepackage[unicode=true,pdfusetitle,
 bookmarks=true,bookmarksnumbered=false,bookmarksopen=false,
 breaklinks=false,pdfborder={0 0 1},backref=false,colorlinks=false]
 {hyperref}

\makeatletter

\providecommand{\tabularnewline}{\\}

\theoremstyle{plain}
\newtheorem{thm}{\protect\theoremname}
\theoremstyle{plain}
\newtheorem{assumption}{\protect\assumptionname}
\theoremstyle{plain}
\newtheorem{lem}{\protect\lemmaname}
\theoremstyle{plain}
\newtheorem{cor}{\protect\corollaryname}
\theoremstyle{definition}
 \newtheorem{example}{\protect\examplename}
\theoremstyle{plain}
\newtheorem{prop}{\protect\propositionname}
\theoremstyle{remark}
\newtheorem{rem}{\protect\remarkname}

\@ifundefined{date}{}{\date{}}
\usepackage{url}
\hypersetup{hidelinks}
\usepackage{yhmath}

\newcommand{\blind}{0}

\usepackage{changepage}

\usepackage[nameinlink,noabbrev]{cleveref}

\crefname{thm}{Theorem}{Theorems}
\Crefname{thm}{Theorem}{Theorems}

\crefname{lem}{Lemma}{Lemmas}
\Crefname{lem}{Lemma}{Lemmas}

\crefname{prop}{Proposition}{Propositions}
\Crefname{prop}{Proposition}{Propositions}

\crefname{cor}{Corollary}{Corollaries}
\Crefname{cor}{Corollary}{Corollaries}

\crefname{assumption}{Assumption}{Assumptions}
\Crefname{assumption}{Assumption}{Assumptions}

\crefname{rem}{Remark}{Remarks}
\Crefname{rem}{Remark}{Remarks}

\crefname{example}{Example}{Examples}
\Crefname{example}{Example}{Examples}

\newenvironment{proofof}[1]
  {\begin{proof}[Proof of \Cref{#1}]}
  {\end{proof}}

\@ifundefined{showcaptionsetup}{}{%
 \PassOptionsToPackage{caption=false}{subfig}}
\usepackage{subfig}
\makeatother

\providecommand{\assumptionname}{Assumption}
\providecommand{\corollaryname}{Corollary}
\providecommand{\examplename}{Example}
\providecommand{\lemmaname}{Lemma}
\providecommand{\propositionname}{Proposition}
\providecommand{\remarkname}{Remark}
\providecommand{\theoremname}{Theorem}

\begin{document}
\global\long\def\spacingset{ \global\long
\global\long\def\baselinestretch{%
}%
\small\normalsize}%
\if0\blind { 
\title{\textbf{PDE-Residual-Informed Kriging}}
\author{\noindent Yu Tan, Shichang Du, Yongxiang Li\thanks{Contact: Yongxiang Li, \protect\href{mailto:yongxiangli@sjtu.edu.cn}{yongxiangli@sjtu.edu.cn},
Department of Industrial Engineering and Management, Shanghai Jiao
Tong University, Shanghai 200240. } \\
 {\small{}Department of Industrial Engineering and Management, \vspace{-4mm}}\\
{\small{}Shanghai Jiao Tong University}}

\maketitle
} \fi

\noindent \if1\blind { \bigskip{}
\vphantom{} \bigskip{}
\vphantom{} \bigskip{}
\textbf{\LARGE{} }{\LARGE\par}
\noindent \begin{center}
\textbf{\LARGE{}PDE-Residual-Informed Kriging}{\LARGE\par}
\par\end{center}

\medskip{}
 } \fi 
\begin{abstract}
\begin{spacing}{1} 

Physics-informed Gaussian process (GP) methods provide a probabilistic
framework for solving partial differential equations (PDEs). Existing
approaches incorporate PDE information in different ways: pointwise
conditioning uses the governing equation only at finitely many selected
locations, globally constrained GPs restrict the admissible functions
to satisfy the PDE exactly, and residual-penalized methods control
the PDE mismatch over the whole domain through an additional penalty
weight. What is still missing is a GP formulation that directly incorporates
global PDE-residual information of the entire domain into its covariance
structure, while retaining model flexibility and standard GP conditioning
without a separate residual-weight parameter. To solve this issue,
this study proposes  PDE-residual-informed Kriging (PRIK) based on
a PDE-residual kernel. The PDE-residual kernel is constructed by equipping
a kernel-section space with an inner product defined by the $L^{2}$
inner product of PDE residuals. This construction makes PRIK yield
the posterior mean with minimum integrated squared PDE residual. It
also provides direct statistical interpretations: the variance-scale
estimate equals the normalized PDE-residual energy (i.e., the integrated
squared PDE residual of the posterior mean divided by the number of
labeled observations), and the profile likelihood balances this residual
energy against model complexity. Under the stated regularity and stability
conditions, we establish $H^{2}$-quasi-optimality and convergence
rates for the posterior mean, together with a nonasymptotic bound
for its PDE-residual norm. We further show that the normalized posterior
kernel converges to a variational Green-type kernel and derive a uniform
nonasymptotic contraction bound for the posterior variance. Numerical
studies provide empirical evidence consistent with the theory and
demonstrate the PDE-solving capability of PRIK.\end{spacing}
\end{abstract}
\begin{spacing}{1}\textit{Keywords:} Gaussian Process, Uncertainty
Quantification, Physics-informed Machine Learning, Partial Differential
Equations, Kernel Methods\end{spacing}

\section{Introduction \label{sec:intro}}

\sloppy 

Physics-informed machine learning (PIML; \citealp{raissi2019physics,karniadakis2021physics})
has become a widely used framework for incorporating scientific knowledge
into data-driven modeling. It has been applied to forward and inverse
problems governed by partial differential equations (PDEs), including
fluid mechanics \citep{raissi2020hidden}, heat transfer \citep{cai2021physics},
wave propagation \citep{rasht2022physics}, and engineering design
problems \citep{sun2023physics}. By embedding physical laws into
learning procedures, PIML provides a way to combine observations with
prior knowledge encoded by the governing equations.

More recently, Gaussian process (GP) methods have received increasing
attention as probabilistic formulations for PDE solving and PIML \citep{chenSolvingLearningNonlinear2021,chenAPIKActivePhysicsInformed2022,pfortner2022physics,liParameterInferenceBased2024,batlleErrorAnalysisKernel2025}.
Conceptually, GP formulations provide a probabilistic perspective
on numerical approximation, connecting PDE solvers with reproducing
kernel Hilbert spaces (RKHSs), optimal recovery, and weighted-residual
methods; practically, the posterior distribution provides uncertainty
quantification for the solution and related quantities, which can
serve as a posteriori or a priori error indicators under suitable
assumptions \citep{chenSolvingLearningNonlinear2021,pfortner2022physics,batlleErrorAnalysisKernel2025}.
These features have motivated a growing literature on GP-based PDE
solving \citep{graepelSolvingNoisyLinear2003,pfortner2022physics}
and related tasks, including learning linear differential equations
from scarce and noisy observations \citep{jidlingLinearlyConstrainedGaussian2017},
solving nonlinear and time-dependent PDEs \citep{NumericalGaussianProcesses,chenSolvingLearningNonlinear2021},
active physics-informed kriging for improving GP prediction \citep{chenAPIKActivePhysicsInformed2022},
and parameter inference for nonlinear PDE systems \citep{liParameterInferenceBased2024}. 

A common approach to incorporate PDE information into GPs is to condition
on virtual observations of the PDE residual at a finite set of interior
points \citep{chenAPIKActivePhysicsInformed2022}. Given a GP prior
for the unknown solution, often called the solution process, one applies
the linear differential operator to the GP to obtain the so-called
PDE process and forms a joint Gaussian distribution for the outputs
of the solution and PDE processes \citep{graepelSolvingNoisyLinear2003,sarkkaLinearOperatorsStochastic2011,chenAPIKActivePhysicsInformed2022}.
Conditioning on the zero PDE residual values at these points and labeled
observations gives the finite-constraint GP predictor \citep{chenSolvingLearningNonlinear2021,chenAPIKActivePhysicsInformed2022,pfortner2022physics}.
We refer to this mechanism as pointwise PDE conditioning. It provides
a flexible probabilistic construction and can be implemented with
a user-specified kernel, but the PDE information is used only through
finitely many pointwise constraints. Consequently, satisfying the
virtual observations at the selected points does not directly quantify
the PDE mismatch over the whole domain and may lead to loss of physics
information away from these points.

This limitation motivates constructions that use PDE information over
the entire domain. One approach is hard enforcement through globally
constrained GPs, which rely on specialized kernels designed so that
the resulting predictions satisfy the PDE globally \citep{swilerSURVEYCONSTRAINEDGAUSSIAN2020}.
Early examples include divergence-free and curl-free kernels for vector-valued
fields \citep{narcowichGeneralizedHermiteInterpolation1994,fuselierRefinedErrorEstimates2007},
which have been further developed \citep{fan2018modeling,drake2021partition,HHDGP2024XuHao,ni2025representing}.
\citet{jidlingLinearlyConstrainedGaussian2017} proposed a transformation-operator
construction for linearly constrained vector-valued GPs, and \citet{lange-hegermannAlgorithmicLinearlyConstrained2018,lange-hegermannLinearlyConstrainedGaussian2021}
developed related algebraic constructions based on Gröbner bases.
Other algebraic approaches include Smith-normal-form-based GPs for
systems of ordinary differential equations \citep{besginowConstrainingGaussianProcesses2022}
and Ehrenpreis--Palamodov GPs for constant-coefficient PDE systems
\citep{harkonenGaussianProcessPriors2023}. These constructions are
attractive because the PDE constraint is built into the prior itself.
However, they require a suitable transformation or parametrizing operator,
whose construction is problem-dependent and available only for restricted
classes of linear PDEs. 

A second approach uses soft enforcement by incorporating a PDE-residual
penalty into the objective function, so that the PDE mismatch is controlled
throughout the domain. These residual-penalized methods typically
define this penalty through the integral of the squared PDE residual,
which we refer to as the integrated squared PDE residual. Unlike globally
constrained GPs, residual-penalized methods allow nonzero PDE residuals
while controlling the overall PDE mismatch across the domain. Similar
ideas appear in statistical spatial modeling, where PDE penalization
and differential regularization have been used to incorporate mechanistic
structure into regression over complex domains \citep{azzimonti2015blood,sangalli2021spatial}.
More recently, physics-informed kernel learning (PIKL) formulated
PIML as a kernel regression problem with a PDE penalty added to the
empirical risk, and the resulting estimator is computed through a
truncated Fourier-series representation \citep{doumechePhysicsinformedMachineLearning2024,doumechePhysicsinformedKernelLearning2025}.
These methods use PDE information over the domain rather than only
at selected points. However, the relative influence of the PDE residual
is controlled by manually tuning parameters that balance physical
consistency against data fidelity and additional regularization.

Existing approaches involve different compromises in how PDE information
is incorporated into a GP. Pointwise conditioning is flexible but
uses the governing equation only at finitely many selected locations.
Globally constrained GPs enforce the PDE throughout the domain, but
require the admissible function class to satisfy the equation exactly
and typically rely on problem-specific constructions. Residual-penalized
methods control the PDE mismatch over the entire domain, but introduce
a separate parameter to balance the PDE consistency against model
fitting. What is still missing is a GP formulation in which global
PDE consistency over the entire domain is an intrinsic part of the
GP model itself, while retaining model flexibility and standard GP
conditioning without a separate residual-weight parameter. 

Motivated by this observation, we propose PDE-residual-informed kriging
(PRIK), which incorporates the governing equation at the level of
GP construction rather than through auxiliary PDE constraints or residual
penalization. The central idea is to use the $L^{2}$ inner product
of PDE residuals to define the inner-product structure of a kernel-section
space and to multiply the resulting reproducing kernel, termed the
PDE-residual kernel, by a variance-scale parameter to construct the
prior covariance kernel of PRIK. PDE consistency over the entire domain
is therefore built into PRIK directly without introducing a separate
residual-weight parameter. Based on the PDE-residual kernel, we derive
the likelihood-based parameter estimation and predictive distribution
of PRIK. We then show that the posterior mean is the interpolation
function with minimum integrated squared PDE residual, while the variance-scale
estimate and profile likelihood admit corresponding PDE-residual interpretations.

The contributions of this study are threefold. First, PRIK incorporates
the PDE directly through the prior covariance kernel without requiring
a problem-specific parameterization of the exact PDE solution space.
It uses the PDE residual over the entire domain rather than at finitely
many auxiliary locations and requires no separate residual-weight
parameter in the noiseless setting.  Second, we provide a unified
statistical interpretation of PRIK in terms of the integrated squared
PDE residual: PRIK yields the interpolation function that minimizes
the integrated squared PDE residual; the variance-scale estimate equals
the integrated squared PDE residual divided by the number of labeled
observations, and the profile likelihood balances this quantity against
the model complexity. Third, we establish theoretical guarantees for
both prediction accuracy and posterior uncertainty as the function
space is refined. Under the stated conditions, the posterior mean
is $H^{2}$-quasi-optimal and achieves the corresponding kernel approximation
rate, with a nonasymptotic bound for its PDE-residual norm. We also
show that the normalized posterior kernel converges to a variational
Green-type kernel and that the posterior variance contracts uniformly.

The rest of the paper is organized as follows. Section \ref{sec:Review}
reviews the problem setting and two related kernel-based approaches.
Section \ref{sec:PRIK} introduces PRIK, including the PDE-residual
kernel, parameter estimation, prediction, and the interpretation.
Section \ref{sec:Theoretical-Properties} further studies the theoretical
properties of PRIK. Section \ref{sec:Numerical-PDE-Solving} presents
numerical studies for PDE-solving examples. The final Section \ref{sec:Conclusion}
concludes, and the technical proofs and additional results on the
PDE-residual Gram matrix are collected in the Appendix.

\section{Review of Previous Work \label{sec:Review}}

\subsection{Problem Setting and Notation}

Let $\Omega\subset\mathbb{R}^{d}$ be a bounded connected domain with
boundary $\mathbb{E}=\partial\Omega$. We consider a target function
$f$ satisfying the known homogeneous linear PDE
\begin{equation}
\mathscr{D}f=0,\quad\text{in }\Omega,\label{eq:Linear-PDE}
\end{equation}
where $\mathscr{D}=\sum_{\left|\boldsymbol{\alpha}\right|\leqslant s}a_{\boldsymbol{\alpha}}\left(\boldsymbol{x}\right)\partial^{\boldsymbol{\alpha}}$
is a linear differential operator of order at most $s$, with known
PDE coefficients $\left\{ a_{\boldsymbol{\alpha}}\left(\boldsymbol{x}\right)\right\} _{\left|\boldsymbol{\alpha}\right|\leqslant s}$.
Here, $\boldsymbol{\alpha}=\left[\alpha_{1},\cdots,\alpha_{d}\right]\in\mathbb{N}_{0}^{d}$
denotes a multi-index, and its order is defined by $\left|\boldsymbol{\alpha}\right|=\sum_{j=1}^{d}\alpha_{j}$.
For a sufficiently smooth function $u$, we write 
\[
\partial^{\boldsymbol{\alpha}}u=\frac{\partial^{\left|\boldsymbol{\alpha}\right|}u}{\partial x_{1}^{\alpha_{1}}\cdots\partial x_{d}^{\alpha_{d}}}.
\]
For a function $u$, we call $\mathscr{D}u$ its PDE residual.

Suppose we observe noiseless data at $\boldsymbol{X}_{l}=\left[\boldsymbol{x}_{1}^{\top},\cdots,\boldsymbol{x}_{N_{l}}^{\top}\right]^{\top}\in\mathbb{R}^{N_{l}\times d}$
with $\boldsymbol{y}_{l}=\left[f\left(\boldsymbol{x}_{1}\right),\cdots,f\left(\boldsymbol{x}_{N_{l}}\right)\right]^{\top}\in\mathbb{R}^{N_{l}}$.
The observation locations may lie in the interior of $\Omega$, on
its boundary $\mathbb{E}$, or both. The objective is to construct
an estimator $\hat{f}$ for $f$ using the labeled data $\left(\boldsymbol{X}_{l},\boldsymbol{y}_{l}\right)$
and the differential equation $\mathscr{D}f=0$. In the following
review, we focus on two kernel-based constructions that incorporate
the PDE information in different ways.

\subsection{PDE-Informed Kriging (PIK) \label{subsec:PIK}}

PDE-informed kriging (PIK) incorporates the PDE through a finite set
of virtual PDE observations \citep{chenAPIKActivePhysicsInformed2022}.
Let $\boldsymbol{X}_{\mathrm{int}}\in\mathbb{R}^{N_{\mathrm{int}}\times d}$
denote a set of interior points of $\Omega$, and the virtual observations
are given by 
\[
\mathscr{D}f\left(\boldsymbol{X}_{\mathrm{int}}\right)=\boldsymbol{0}.
\]
Suppose that $f\sim\mathcal{GP}\left(0,k_{\boldsymbol{\theta}}\left(\cdot,\cdot\right)\right)$,
where $k_{\boldsymbol{\theta}}\left(\cdot,\cdot\right)$ is a sufficiently
smooth base kernel such that $\mathscr{D}f\left(\cdot\right)$ is
well defined. In particular, the covariance functions involving $\left\{ \mathscr{D}_{\boldsymbol{x}}k_{\boldsymbol{\theta}}\left(\cdot,\cdot\right),\mathscr{D}_{\boldsymbol{x}^{\prime}}k_{\boldsymbol{\theta}}\left(\cdot,\cdot\right),\mathscr{D}_{\boldsymbol{x}}\mathscr{D}_{\boldsymbol{x}^{\prime}}k_{\boldsymbol{\theta}}\left(\cdot,\cdot\right)\right\} $
are assumed to be well defined. Here, $\mathscr{D}_{\boldsymbol{x}}$
and $\mathscr{D}_{\boldsymbol{x}^{\prime}}$ indicate that the differential
operator is applied to the first and second arguments of the kernel,
respectively. The PIK method can then be viewed as a cokriging construction
for $f\left(\cdot\right)$ and $\mathscr{D}f\left(\cdot\right)$ \citep{yang2019physics}.
For any prediction point $\boldsymbol{x}\in\Omega$, the joint distribution
is
\[
\left[\begin{array}{c}
f\left(\boldsymbol{X}_{l}\right)\\
\mathscr{D}f\left(\boldsymbol{X}_{\mathrm{int}}\right)\\
f\left(\boldsymbol{x}\right)
\end{array}\right]\sim\mathcal{N}\left(\boldsymbol{0},\left[\begin{matrix}k_{\boldsymbol{\theta}}\left(\boldsymbol{X}_{l},\boldsymbol{X}_{l}\right) & \mathscr{D}_{\boldsymbol{x}^{\prime}}k_{\boldsymbol{\theta}}\left(\boldsymbol{X}_{l},\boldsymbol{X}_{\mathrm{int}}\right) & k_{\boldsymbol{\theta}}\left(\boldsymbol{X}_{l},\boldsymbol{x}\right)\\
\mathscr{D}_{\boldsymbol{x}}k_{\boldsymbol{\theta}}\left(\boldsymbol{X}_{\mathrm{int}},\boldsymbol{X}_{l}\right) & \mathscr{D}_{\boldsymbol{x}}\mathscr{D}_{\boldsymbol{x}^{\prime}}k_{\boldsymbol{\theta}}\left(\boldsymbol{X}_{\mathrm{int}},\boldsymbol{X}_{\mathrm{int}}\right) & \mathscr{D}_{\boldsymbol{x}}k_{\boldsymbol{\theta}}\left(\boldsymbol{X}_{\mathrm{int}},\boldsymbol{x}\right)\\
k_{\boldsymbol{\theta}}\left(\boldsymbol{x},\boldsymbol{X}_{l}\right) & \mathscr{D}_{\boldsymbol{x}^{\prime}}k_{\boldsymbol{\theta}}\left(\boldsymbol{x},\boldsymbol{X}_{\mathrm{int}}\right) & k_{\boldsymbol{\theta}}\left(\boldsymbol{x},\boldsymbol{x}\right)
\end{matrix}\right]\right).
\]
Conditioning on $f\left(\boldsymbol{X}_{l}\right)=\boldsymbol{y}_{l}$
and $\mathscr{D}f\left(\boldsymbol{X}_{\mathrm{int}}\right)=\boldsymbol{0}$
gives the PIK posterior mean.

\subsection{Physics-Informed Kernel Learning (PIKL) \label{subsec:PIKL}}

Physics-informed kernel learning (PIKL) incorporates the PDE information
through a residual-regularized empirical risk rather than through
virtual PDE observations \citep{doumechePhysicsinformedMachineLearning2024,doumechePhysicsinformedKernelLearning2025}.
Suppose that $\Omega\subset\left[-L,L\right]^{d}$ for some $L>0$.
Its objective can be written in the form
\begin{equation}
\hat{f}_{\mathrm{PIKL}}=\underset{u\in\mathbb{F}_{M}}{\arg\min}\left\{ \frac{1}{N_{l}}\left\Vert u\left(\boldsymbol{X}_{l}\right)-\boldsymbol{y}_{l}\right\Vert _{2}^{2}+\lambda\left\Vert u\right\Vert _{H_{\mathrm{per}}^{s}\left(\left[-2L,2L\right]^{d}\right)}^{2}+\mu\left\Vert \mathscr{D}u\right\Vert _{L^{2}\left(\Omega\right)}^{2}\right\} ,\label{eq:PIKL}
\end{equation}
where $\mathbb{F}_{M}\subset H_{\mathrm{per}}^{s}\left(\left[-2L,2L\right]^{d}\right)$
denotes a truncated Fourier space with truncation level $M$, $H_{\mathrm{per}}^{s}\left(\left[-2L,2L\right]^{d}\right)$
denotes the periodic Sobolev space of order $s$, and $\lambda$ and
$\mu$ are tuning parameters controlling the Sobolev regularization
and the PDE-residual penalty, respectively. To obtain a computationally
tractable estimator, PIKL embeds the problem into a periodic Sobolev
space with a truncated Fourier basis. 

This construction differs from PIK in two main respects. PIK keeps
a user-specified base kernel $k_{\boldsymbol{\theta}}\left(\cdot,\cdot\right)$
and incorporates the PDE through local virtual observations $\mathscr{D}f\left(\boldsymbol{X}_{\mathrm{int}}\right)=\boldsymbol{0}$.
PIKL instead uses a global PDE-residual penalty, and derives a finite-dimensional
Fourier-feature estimator. However, neither approach simultaneously
retains the flexibility of a user-specified kernel and incorporates
PDE information over the entire domain without introducing a manually
selected residual weight.

\section{PDE-Residual-Informed Kriging \label{sec:PRIK} }

The construction of PRIK is guided by three requirements. First, the
PDE information should be incorporated through the residual over the
entire domain rather than through residual values at finitely many
selected points. Second, this incorporation should not require an
additional parameter that trades PDE consistency against interpolation
of the observed data. Third, the construction should retain the flexibility
of a user-specified function space and the standard probabilistic
machinery of GP regression. PRIK achieves these requirements by defining
the RKHS norm through the $L^{2}$-norm of the PDE residual. We first
formulate this construction and derive its associated kernel, and
then show how it leads to standard GP inference and a posterior mean
that minimizes the integrated squared PDE residual.

\subsection{Model Formulation\label{subsec:Model-Formulation}}

We first introduce some notation. Let $\mathbb{V}_{p}=\mathrm{span}\left\{ \psi_{1}\left(\cdot\right),\cdots,\psi_{p}\left(\cdot\right)\right\} $
be a $p$-dimensional function space on $\bar{\Omega}=\Omega\cup\mathbb{E}$,
where $\left\{ \psi_{j}\left(\cdot\right)\right\} _{j=1}^{p}$ are
basis functions satisfying $\mathscr{D}\psi_{j}\in L^{2}\left(\Omega\right)$,
$j=1,\cdots,p$. Here, $L^{2}\left(\Omega\right)$ denotes the space
of square-integrable functions on $\Omega$. Moreover, for $u,v\in\mathbb{V}_{p}$,
we assume that the bilinear form 
\[
\left<\mathscr{D}u,\mathscr{D}v\right>_{L^{2}\left(\Omega\right)}=\int_{\Omega}\mathscr{D}u\left(\boldsymbol{\xi}\right)\mathscr{D}v\left(\boldsymbol{\xi}\right)\mathrm{d}\boldsymbol{\xi}
\]
is positive definite on $\mathbb{V}_{p}$ and call it the PDE-residual
inner product. The quantity $\left\Vert \mathscr{D}u\right\Vert _{L^{2}\left(\Omega\right)}=\sqrt{\left<\mathscr{D}u,\mathscr{D}u\right>_{L^{2}\left(\Omega\right)}}$
measures the global PDE mismatch over the domain. We refer to it as
the PDE-residual norm and call $\left\Vert \mathscr{D}u\right\Vert _{L^{2}\left(\Omega\right)}^{2}$
the PDE-residual energy. In this study, minimizing the PDE residual
refers to minimizing the PDE-residual energy, which is equivalent
to minimizing the PDE-residual norm.

The construction of PRIK builds on the fact that noiseless kernel
interpolation selects the interpolation function with minimum RKHS
norm \citep[Theorem 13.2]{wendlandScatteredDataApproximation2004}.
To make this minimum-RKHS-norm principle coincide with minimum PDE-residual
interpolation while incorporating the PDE solely through the covariance
structure, we equip $\mathbb{V}_{p}$ with the PDE-residual inner
product $\left<\mathscr{D}u,\mathscr{D}v\right>_{L^{2}\left(\Omega\right)}$
and let $\tilde{k}_{\boldsymbol{\theta}}\left(\cdot,\cdot\right)$
denote the reproducing kernel of $\mathbb{V}_{p}$ under this inner
product. Hence, $\mathcal{H}_{\tilde{k}}=\mathbb{V}_{p}$ and
\begin{equation}
\left<u,v\right>_{\mathcal{H}_{\tilde{k}}}=\left<\mathscr{D}u,\mathscr{D}v\right>_{L^{2}\left(\Omega\right)}\text{, for }u,v\in\mathbb{V}_{p}.\label{eq:Desired-Properties}
\end{equation}
We then model the target function $f$ in \eqref{eq:Linear-PDE} as
\begin{equation}
f\sim\mathcal{GP}\left(0,\sigma^{2}\tilde{k}_{\boldsymbol{\theta}}\left(\cdot,\cdot\right)\right),\label{eq:PRIK}
\end{equation}
where $\sigma^{2}\geqslant0$ is the variance-scale parameter, $\sigma^{2}\tilde{k}_{\boldsymbol{\theta}}\left(\cdot,\cdot\right)$
is the prior covariance kernel,  and $\boldsymbol{\theta}$ collects
the parameters that determine the function space $\mathbb{V}_{p}$,
such as the scale parameters of a base kernel $k_{\boldsymbol{\theta}}\left(\cdot,\cdot\right)$
in the kernel-section construction $\psi_{j}\left(\boldsymbol{x}\right)=k_{\boldsymbol{\theta}}\left(\boldsymbol{z}_{j},\boldsymbol{x}\right)$
with $\boldsymbol{z}_{j}\in\Omega$. We refer to this GP model as
PDE-residual-informed kriging (PRIK).

Eq. \eqref{eq:Desired-Properties} defines the central construction
of PRIK. The PDE residual is neither imposed at a finite set of auxiliary
locations nor appended to the data-fitting criterion as a separate
penalty. Instead, the PDE-residual norm $\left\Vert \mathscr{D}u\right\Vert _{L^{2}\left(\Omega\right)}$,
which measures global PDE inconsistency, is used directly as the RKHS
norm $\left\Vert u\right\Vert _{\mathcal{H}_{\tilde{k}}}$. Consequently,
once the corresponding reproducing kernel $\tilde{k}_{\boldsymbol{\theta}}\left(\cdot,\cdot\right)$
is constructed, PDE information enters all subsequent GP calculations
through the kernel function itself. The remaining question is therefore
how the PDE-residual inner product can be translated into an explicit
covariance kernel for GP regression.

\subsection{PDE-Residual Kernel Construction \label{subsec:PDE-Residual_Kernel}}

For convenience, we first define the PDE-residual Gram matrix 
\[
\boldsymbol{A}=\left<\mathscr{D}\boldsymbol{\psi}\left(\cdot\right),\mathscr{D}\boldsymbol{\psi}\left(\cdot\right)\right>_{L^{2}\left(\Omega\right)}=\left[\left<\mathscr{D}\psi_{i}\left(\cdot\right),\mathscr{D}\psi_{j}\left(\cdot\right)\right>_{L^{2}\left(\Omega\right)}\right]_{i,j\in\left\{ 1,2,\cdots,p\right\} },
\]
where $\boldsymbol{\psi}\left(\boldsymbol{x}\right)=\left[\psi_{1}\left(\boldsymbol{x}\right),\cdots,\psi_{p}\left(\boldsymbol{x}\right)\right]^{\top}$.
In particular, when $\mathbb{V}_{p}$ is generated from the base kernel
$k_{\boldsymbol{\theta}}\left(\cdot,\cdot\right)$ through $\psi_{j}\left(\boldsymbol{x}\right)=k_{\boldsymbol{\theta}}\left(\boldsymbol{z}_{j},\boldsymbol{x}\right)$
with interior centers $\boldsymbol{Z}=\left[\boldsymbol{z}_{1},\cdots,\boldsymbol{z}_{p}\right]^{\top}$,
we have 
\[
\boldsymbol{A}=\left[\left<\mathscr{D}k_{\boldsymbol{\theta}}\left(\cdot,\boldsymbol{z}_{i}\right),\mathscr{D}k_{\boldsymbol{\theta}}\left(\cdot,\boldsymbol{z}_{j}\right)\right>_{L^{2}\left(\Omega\right)}\right]_{i,j\in\left\{ 1,2,\cdots,p\right\} }.
\]
Thus, the base kernel $k_{\boldsymbol{\theta}}\left(\cdot,\cdot\right)$
determines the finite-dimensional function space $\mathbb{V}_{p}$,
while the PDE-residual Gram matrix $\boldsymbol{A}$ incorporates
the differential operator $\mathscr{D}$ through the PDE-residual
inner products of its basis functions. In addition, they together
determine the PDE-residual kernel $\tilde{k}_{\boldsymbol{\theta}}\left(\cdot,\cdot\right)$
constructed below. 

Since the bilinear form $\left<\mathscr{D}u,\mathscr{D}v\right>_{L^{2}\left(\Omega\right)}$
is assumed to be an inner product on $\mathbb{V}_{p}$ and $\left\{ \psi_{j}\left(\cdot\right)\right\} _{j=1}^{p}$
are assumed to be basis functions, the PDE-residual Gram matrix $\boldsymbol{A}$
is positive definite. Appendix \ref{sec:PDE-Gram} provides explicit
conditions for verifying this assumption in several common kernel--operator
settings. In particular, for Gaussian and Matérn kernel sections $\psi_{j}\left(\boldsymbol{x}\right)=k_{\boldsymbol{\theta}}\left(\boldsymbol{z}_{j},\boldsymbol{x}\right)$
with polynomial-coefficient operators and pairwise distinct interior
centers $\boldsymbol{Z}$, $\boldsymbol{A}$ is positive definite
if and only if 

\[
\mathscr{D}k_{\boldsymbol{\theta}}\left(\cdot,\boldsymbol{z}_{j}\right)\ne0,\text{ for }j=1,\cdots,p.
\]

We now construct the PDE-residual kernel explicitly. For each fixed
$\boldsymbol{\xi}\in\bar{\Omega}$, the kernel section $\tilde{k}_{\boldsymbol{\theta}}\left(\cdot,\boldsymbol{\xi}\right)$
belongs to $\mathbb{V}_{p}$. Since $\left\{ \psi_{1}\left(\cdot\right),\cdots,\psi_{p}\left(\cdot\right)\right\} $
is a basis of $\mathbb{V}_{p}$, it can therefore be represented as

\begin{equation}
\tilde{k}_{\boldsymbol{\theta}}\left(\cdot,\boldsymbol{\xi}\right)=\boldsymbol{\psi}\left(\cdot\right)^{\top}\boldsymbol{c}\left(\boldsymbol{\xi}\right),\label{eq:Reproducing-Kernel}
\end{equation}
where $\boldsymbol{c}\left(\boldsymbol{\xi}\right)\in\mathbb{R}^{p}$.
By the reproducing property of $\tilde{k}_{\boldsymbol{\theta}}\left(\cdot,\cdot\right)$,
we have $\left\langle u,\tilde{k}_{\boldsymbol{\theta}}\left(\cdot,\boldsymbol{\xi}\right)\right\rangle _{\mathcal{H}_{\tilde{k}}}=u\left(\boldsymbol{\xi}\right)$
for $u\in\mathbb{V}_{p}$. Using the PDE-residual inner-product identity
in Eq. \eqref{eq:Desired-Properties}, this condition is equivalently
written as 

\begin{equation}
\left<\mathscr{D}u,\mathscr{D}\tilde{k}_{\boldsymbol{\theta}}\left(\cdot,\boldsymbol{\xi}\right)\right>_{L^{2}\left(\Omega\right)}=u\left(\boldsymbol{\xi}\right).\label{eq:Reproducing-Property}
\end{equation}
Now let $u\left(\cdot\right)=\boldsymbol{\psi}\left(\cdot\right)^{\top}\boldsymbol{\beta}$
for arbitrary $\boldsymbol{\beta}\in\mathbb{R}^{p}$. Substituting
Eq. \eqref{eq:Reproducing-Kernel} into Eq. \eqref{eq:Reproducing-Property}
gives $\boldsymbol{\beta}^{\top}\boldsymbol{A}\boldsymbol{c}\left(\boldsymbol{\xi}\right)=\boldsymbol{\beta}^{\top}\boldsymbol{\psi}\left(\boldsymbol{\xi}\right)$
for every $\boldsymbol{\beta}\in\mathbb{R}^{p}$, and hence $\boldsymbol{c}\left(\boldsymbol{\xi}\right)=\boldsymbol{A}^{-1}\boldsymbol{\psi}\left(\boldsymbol{\xi}\right)$.
Therefore, the reproducing kernel of $\mathbb{V}_{p}$ equipped with
the PDE-residual inner product is 
\begin{equation}
\tilde{k}_{\boldsymbol{\theta}}\left(\boldsymbol{x},\boldsymbol{x}^{\prime}\right)=\boldsymbol{\psi}\left(\boldsymbol{x}\right)^{\top}\boldsymbol{A}^{-1}\boldsymbol{\psi}\left(\boldsymbol{x}^{\prime}\right),\label{eq:PR-Kernel}
\end{equation}
which is positive semidefinite because $\boldsymbol{A}^{-1}$ is positive
definite. We refer to it as the PDE-residual kernel. This kernel can
then be used directly in the standard GP formulas for prediction and
uncertainty quantification.

The matrix $\boldsymbol{A}$ records the integrated interactions among
the PDE residuals of the basis functions, while $\boldsymbol{A}^{-1}$
transforms this residual inner-product structure into the reproducing
kernel of the same function space. Thus, the construction changes
the inner-product structure of the function space, rather than imposing
additional PDE observations or modifying the interpolation constraints.

Importantly, the construction does not prescribe a particular family
of basis functions. The function space $\mathbb{V}_{p}$ can be constructed
using kernel sections $\left\{ k_{\boldsymbol{\theta}}\left(\cdot,\boldsymbol{z}\right):\boldsymbol{z}\in\boldsymbol{Z}\right\} $
\citep{wendlandScatteredDataApproximation2004}, multiscale or heterogeneous
kernel functions \citep{farrellRBFMultiscaleCollocation2013}, splines
\citep{hughes2005isogeometric}, or combinations of these functions.
These choices determine the approximation capability and numerical
properties of the resulting PRIK model. This study mainly focuses
on the kernel-section construction because it is straightforward to
implement and analyze. 

Under the kernel-section construction, the base kernel determines
the approximation space and can encode properties such as smoothness
or periodicity, whereas $\boldsymbol{A}$ incorporates the differential
operator. For example, the Gaussian kernel 
\begin{equation}
k_{\boldsymbol{\theta}}\left(\boldsymbol{x},\boldsymbol{x}^{\prime}\right)=\exp\left(-\sum_{i=1}^{d}\theta_{i}^{2}\left(x_{i}-x_{i}^{\prime}\right)^{2}\right)\label{eq:Gauss-Kernel}
\end{equation}
is suitable when the target function is expected to be highly smooth,
whereas the Matérn kernel allows finite smoothness. If the target
function is known to be periodic with period $T_{i}$ along its $i$-th
input coordinate, this information can be incorporated using the periodic
kernel
\begin{equation}
k_{\boldsymbol{\theta}}\left(\boldsymbol{x},\boldsymbol{x}^{\prime}\right)=\exp\left(-\sum_{i=1}^{d}\theta_{i}^{2}\sin^{2}\left(\frac{\pi\left(x_{i}-x_{i}^{\prime}\right)}{T_{i}}\right)\right).\label{eq:Periodic-Kernel}
\end{equation}
When no additional structural information about the target function
is available, this study uses the Gaussian kernel as a simple default. 

\subsection{Parameter Estimation\label{subsec:Parameter-Estimation}}

The PDE-residual kernel derived in Section \ref{subsec:PDE-Residual_Kernel}
allows the parameters of PRIK to be estimated through the Gaussian
likelihood. Under the PRIK model, $\boldsymbol{y}_{l}\sim\mathcal{N}\left(\boldsymbol{0},\sigma^{2}\boldsymbol{K}_{\boldsymbol{\theta}}\right)$,
and hence the log-likelihood, up to an additive constant, is
\begin{equation}
\ell\left(\boldsymbol{\theta},\sigma^{2}\right)=-\frac{1}{2}\left(N_{l}\log\sigma^{2}+\log\left|\boldsymbol{K}_{\boldsymbol{\theta}}\right|+\frac{\boldsymbol{y}_{l}^{\top}\boldsymbol{K}_{\boldsymbol{\theta}}^{-1}\boldsymbol{y}_{l}}{\sigma^{2}}\right),\label{eq:lik0}
\end{equation}
where $\boldsymbol{K}_{\boldsymbol{\theta}}=\tilde{k}_{\boldsymbol{\theta}}\left(\boldsymbol{X}_{l},\boldsymbol{X}_{l}\right)$.
We assume that the observation kernel matrix $\boldsymbol{K}_{\boldsymbol{\theta}}$
is positive definite for the parameter values under consideration.
Because $\tilde{k}_{\boldsymbol{\theta}}\left(\cdot,\cdot\right)$
is a finite-rank kernel, positive definiteness of $\boldsymbol{A}$
alone does not guarantee that of $\boldsymbol{K}_{\boldsymbol{\theta}}$.
Under the kernel-section construction, positive definiteness of $\boldsymbol{K}_{\boldsymbol{\theta}}$
is equivalent to

\[
\mathrm{rank}\left(k_{\boldsymbol{\theta}}\left(\boldsymbol{X}_{l},\boldsymbol{Z}\right)\right)=N_{l},
\]
which necessarily requires $p\geqslant N_{l}$. Sufficient conditions
based on the locations of the labeled observations $\boldsymbol{X}_{l}$
and interior kernel centers $\boldsymbol{Z}$ are provided in Appendix
\ref{subsec:Kernel-Matrix-PD}.

For fixed $\boldsymbol{\theta}$, differentiating $\ell\left(\boldsymbol{\theta},\sigma^{2}\right)$
in Eq. \eqref{eq:lik0} with respect to $\sigma^{2}$ yields the maximum-likelihood
estimator

\begin{equation}
\hat{\sigma}^{2}\left(\boldsymbol{\theta}\right)=\frac{1}{N_{l}}\boldsymbol{y}_{l}^{\top}\boldsymbol{K}_{\boldsymbol{\theta}}^{-1}\boldsymbol{y}_{l}.\label{eq:Sigma2}
\end{equation}
If $\boldsymbol{y}_{l}=\boldsymbol{0}$, the expression in Eq. \eqref{eq:Sigma2}
equals zero and parameter estimation is unnecessary. In this case,
substituting $\boldsymbol{y}_{l}=\boldsymbol{0}$ into the posterior
mean in Eq. \eqref{eq:Predictor} gives $\hat{f}\left(\boldsymbol{x}\right)=0$
for every $\boldsymbol{x}$.

Finally, for $\boldsymbol{y}_{l}\ne\boldsymbol{0}$, $\boldsymbol{\theta}$
can be numerically optimized by minimizing the negative profile log-likelihood:
\begin{equation}
\hat{\boldsymbol{\theta}}=\mathrm{arg}\underset{\boldsymbol{\theta}}{\min}\left\{ N_{l}\log\hat{\sigma}^{2}\left(\boldsymbol{\theta}\right)+\log\left|\boldsymbol{K}_{\boldsymbol{\theta}}\right|\right\} .\label{eq:MLE-Opt}
\end{equation}
The final variance-scale estimate is given by $\hat{\sigma}^{2}=\hat{\sigma}^{2}\left(\hat{\boldsymbol{\theta}}\right)$.
Thus, PDE information enters parameter estimation through the covariance
matrix itself, without an additional residual penalty in the likelihood.
The PDE-residual interpretations of the variance-scale estimate $\hat{\sigma}^{2}$
in Eq. \eqref{eq:Sigma2} and the profile-likelihood criterion in
Eq. \eqref{eq:MLE-Opt} are developed in Section \ref{subsec:Interpretability-of-PRIK}.

The objective function in Eq. \eqref{eq:MLE-Opt} can be nonconvex.
In our implementation, it is minimized using the MATLAB routine fmincon
from multiple initial values to reduce sensitivity to initialization.
A small diagonal nugget may be introduced for numerical stabilization
\citep{santnerDesignAnalysisComputer2018}.

\subsection{Prediction Distribution\label{subsec:Prediction-Distribution}}

Prediction is obtained by substituting the parameter estimates $\hat{\boldsymbol{\theta}}$
and $\hat{\sigma}^{2}$ from Section \ref{subsec:Parameter-Estimation}
into the PRIK model and treating them as fixed. The posterior mean
of PRIK at a new input $\boldsymbol{x}\in\Omega$ is
\begin{align}
\hat{f}\left(\boldsymbol{x}\right) & =\tilde{k}_{\hat{\boldsymbol{\theta}}}\left(\boldsymbol{x},\boldsymbol{X}_{l}\right)\boldsymbol{K}_{\hat{\boldsymbol{\theta}}}^{-1}\boldsymbol{y}_{l}.\label{eq:Predictor}
\end{align}
For $\boldsymbol{x},\boldsymbol{x}^{\prime}\in\bar{\Omega}$, the
normalized posterior kernel evaluated at $\hat{\boldsymbol{\theta}}$
is defined as
\begin{equation}
c_{\hat{\boldsymbol{\theta}}}\left(\boldsymbol{x},\boldsymbol{x}^{\prime}\right)=\tilde{k}_{\hat{\boldsymbol{\theta}}}\left(\boldsymbol{x},\boldsymbol{x}^{\prime}\right)-\tilde{k}_{\hat{\boldsymbol{\theta}}}\left(\boldsymbol{x},\boldsymbol{X}_{l}\right)\boldsymbol{K}_{\hat{\boldsymbol{\theta}}}^{-1}\tilde{k}_{\hat{\boldsymbol{\theta}}}\left(\boldsymbol{X}_{l},\boldsymbol{x}^{\prime}\right),\label{eq:Normalized-Var}
\end{equation}
and the posterior covariance kernel is defined as $\hat{\sigma}^{2}c_{\hat{\boldsymbol{\theta}}}\left(\cdot,\cdot\right)$.
Then, the posterior variance of PRIK at $\boldsymbol{x}$ is given
by
\begin{align}
\hat{\sigma}_{f}^{2}\left(\boldsymbol{x}\right) & =\hat{\sigma}^{2}c_{\hat{\boldsymbol{\theta}}}\left(\boldsymbol{x},\boldsymbol{x}\right).\label{eq:Var}
\end{align}
Because PRIK retains the GP formulation, Eq. \eqref{eq:Var} provides
a model-based measure of predictive uncertainty in addition to the
posterior mean. This further distinguishes PRIK from the deterministic
PIKL estimator reviewed in Section \ref{subsec:PIKL}. It is worth
noting that
\[
\hat{f}\left(\boldsymbol{X}_{l}\right)=\boldsymbol{y}_{l}\text{ and }\hat{\sigma}_{f}^{2}\left(\boldsymbol{X}_{l}\right)=\boldsymbol{0}.
\]
Thus, the labeled observations are interpolated exactly, and the posterior
variance vanishes at the labeled locations.

For the kernel-section construction with a fixed Sobolev-compatible
base kernel, Theorem \ref{thm:Posterior-Mean-Convergence-Rate} establishes
an explicit $H^{2}$-convergence rate for the PRIK posterior mean.
Corollary \ref{cor:Decay-Predictive-Sigma2} further establishes a
uniform decay bound for the posterior variance in Eq. \eqref{eq:Var}
as the function space is refined.

\subsection{Statistical and Optimization Interpretations \label{subsec:Interpretability-of-PRIK}}

All statements in this subsection hold for each admissible $\boldsymbol{\theta}$.
For each such $\boldsymbol{\theta}$, let $\hat{f}_{\boldsymbol{\theta}}\left(\cdot\right)$
denote the posterior mean computed at $\boldsymbol{\theta}$:
\begin{equation}
\hat{f}_{\boldsymbol{\theta}}\left(\boldsymbol{x}\right)=\tilde{k}_{\boldsymbol{\theta}}\left(\boldsymbol{x},\boldsymbol{X}_{l}\right)\boldsymbol{K}_{\boldsymbol{\theta}}^{-1}\boldsymbol{y}_{l}.\label{eq:Predictor-Theta}
\end{equation}

The PDE-residual inner product introduced in Section \ref{subsec:Model-Formulation}
connects the posterior mean, variance-scale estimate, and profile
likelihood through the same quantity: the PDE-residual energy. We
first characterize the posterior mean as the interpolation function
with minimum PDE-residual energy and then use this characterization
to interpret the likelihood-based parameter estimation in Section
\ref{subsec:Parameter-Estimation}. The results below concern the
noiseless model with the PDE-residual kernel in Eq. \eqref{eq:PR-Kernel},
without numerical nugget modification.

After conditioning on the labeled observations, the PRIK posterior
mean is not merely a GP posterior mean induced by $\tilde{k}_{\boldsymbol{\theta}}\left(\cdot,\cdot\right)$;
it is exactly the interpolation function that minimizes the PDE-residual
energy within the chosen function space. That is, the identity 
\[
\left\Vert u\right\Vert _{\mathcal{H}_{\tilde{k}}}^{2}=\left\Vert \mathscr{D}u\right\Vert _{L^{2}\left(\Omega\right)}^{2},
\]
derived from Eq. \eqref{eq:Desired-Properties}, turns the minimum-RKHS-norm
property of noiseless GP interpolation into a minimum-PDE-residual
property, as shown in the following theorem.
\begin{thm}
\label{thm:Interpolation+minPDE} (Optimization Characterization of
the PRIK Posterior Mean) Assume that the PDE-residual Gram matrix
$\boldsymbol{A}$ and the kernel matrix $\boldsymbol{K}_{\boldsymbol{\theta}}$
are positive definite. The PRIK posterior mean $\hat{f}_{\boldsymbol{\theta}}\left(\cdot\right)$
in Eq. \eqref{eq:Predictor-Theta} is the unique solution to
\begin{equation}
\min_{u\in\mathbb{V}_{p}}\left\Vert \mathscr{D}u\right\Vert _{L^{2}\left(\Omega\right)}^{2},\text{ s.t. }u\left(\boldsymbol{X}_{l}\right)=\boldsymbol{y}_{l}.\label{eq:Constrained-Opt-Vp}
\end{equation}
\end{thm}
Theorem \ref{thm:Interpolation+minPDE} shows that, among all functions
in $\mathbb{V}_{p}$ that interpolate the labeled observations, the
PRIK posterior mean has the smallest PDE-residual energy. Thus, PRIK
incorporates the PDE through the covariance construction while retaining
exact interpolation of the observed data. This minimum-residual property
also yields an orthogonality relation, i.e.,
\[
\left<\mathscr{D}\hat{f}_{\boldsymbol{\theta}},\mathscr{D}w\right>_{L^{2}\left(\Omega\right)}=0,\quad\forall w\in\mathbb{W}_{p},
\]
where $\mathbb{W}_{p}=\left\{ w\in\mathbb{V}_{p}:w\left(\boldsymbol{X}_{l}\right)=\boldsymbol{0}\right\} $
denotes the label-homogeneous subspace of $\mathbb{V}_{p}$. Consequently,
every interpolation function $u\in\mathbb{V}_{p}$ can be written
as $u=\hat{f}_{\boldsymbol{\theta}}+w$ for some $w\in\mathbb{W}_{p}$,
and its PDE-residual energy satisfies
\begin{equation}
\left\Vert \mathscr{D}u\right\Vert _{L^{2}\left(\Omega\right)}^{2}=\left\Vert \mathscr{D}\hat{f}_{\boldsymbol{\theta}}\right\Vert _{L^{2}\left(\Omega\right)}^{2}+\left\Vert \mathscr{D}\left(u-\hat{f}_{\boldsymbol{\theta}}\right)\right\Vert _{L^{2}\left(\Omega\right)}^{2}.\label{eq:PDE-Energy-Decomposition}
\end{equation}
Hence, every nonzero admissible perturbation produces a strictly larger
PDE-residual energy. This identity makes explicit how the PDE-residual
inner product introduced in Section \ref{subsec:Model-Formulation}
selects the unique posterior mean after conditioning on the observations.

The variance-scale estimate also has a direct PDE-residual interpretation.
Combining the identity in Eq. \eqref{eq:Desired-Properties} and the
variance-scale estimate in Eq. \eqref{eq:Sigma2} yields
\begin{equation}
\left\Vert \hat{f}_{\boldsymbol{\theta}}\right\Vert _{\mathcal{H}_{\tilde{k}}}^{2}=\left\Vert \mathscr{D}\hat{f}_{\boldsymbol{\theta}}\right\Vert _{L^{2}\left(\Omega\right)}^{2}=\boldsymbol{y}_{l}^{\top}\boldsymbol{K}_{\boldsymbol{\theta}}^{-1}\boldsymbol{y}_{l}=N_{l}\hat{\sigma}^{2}\left(\boldsymbol{\theta}\right).\label{eq:Sigma2-Interpretation}
\end{equation}
Hence, for each admissible $\boldsymbol{\theta}$,  the variance-scale
estimate $\hat{\sigma}^{2}\left(\boldsymbol{\theta}\right)$ equals
the PDE-residual energy of the posterior mean normalized by $N_{l}$.
In view of Theorem \ref{thm:Interpolation+minPDE}, $\hat{\sigma}^{2}\left(\boldsymbol{\theta}\right)$
is also the minimum PDE-residual energy attainable under the interpolation
constraints in the corresponding function space, divided by $N_{l}$.
In addition, combining Eq. \eqref{eq:Var} with Eq. \eqref{eq:Sigma2-Interpretation}
gives
\[
\hat{\sigma}_{f}^{2}\left(\boldsymbol{x};\boldsymbol{\theta}\right)=\frac{1}{N_{l}}\left\Vert \mathscr{D}\hat{f}_{\boldsymbol{\theta}}\right\Vert _{L^{2}\left(\Omega\right)}^{2}c_{\boldsymbol{\theta}}\left(\boldsymbol{x},\boldsymbol{x}\right).
\]
Here, $c_{\boldsymbol{\theta}}\left(\cdot,\cdot\right)$ denotes the
normalized posterior kernel from Section \ref{subsec:Prediction-Distribution}
evaluated at $\boldsymbol{\theta}$, and $\hat{\sigma}_{f}^{2}\left(\boldsymbol{x};\boldsymbol{\theta}\right)$
denotes the posterior variance evaluated using $\boldsymbol{\theta}$
and $\hat{\sigma}^{2}\left(\boldsymbol{\theta}\right)$. Thus, the
PDE-residual energy determines the overall scale of the posterior
variance, while the normalized posterior kernel specifies its spatial
variation. 

Finally, the same identity provides a residual--complexity interpretation
of the profile likelihood. Using Eq. \eqref{eq:Sigma2-Interpretation},
the negative profile log-likelihood in Eq. \eqref{eq:MLE-Opt} can
be written, up to a constant, as
\[
N_{l}\log\left\Vert \mathscr{D}\hat{f}_{\boldsymbol{\theta}}\right\Vert _{L^{2}\left(\Omega\right)}^{2}+\log\left|\boldsymbol{K}_{\boldsymbol{\theta}}\right|,
\]
which holds for each admissible $\boldsymbol{\theta}$. The first
term reflects the fitted PDE-residual energy of the PRIK posterior
mean, whereas the second term is the usual GP log-determinant term
associated with covariance complexity. Hence, maximum-likelihood estimation
jointly evaluates the fitted PDE-residual energy and the corresponding
GP complexity. Throughout the parameter selection of $\boldsymbol{\theta}$,
the labeled observations remain exact interpolation constraints; no
separate residual-weight parameter is introduced to balance PDE consistency
against data fitting. These interpretations distinguish two complementary
optimization steps: the posterior mean minimizes the PDE-residual
energy within a fixed function space, whereas the profile likelihood
selects the parameters that determine that space and its associated
covariance structure.

\section{Theoretical Properties\label{sec:Theoretical-Properties}}

Although Theorem \ref{thm:Interpolation+minPDE} in Section \ref{subsec:Interpretability-of-PRIK}
characterizes the posterior mean as the minimum-PDE-residual interpolation
function, residual minimization alone does not guarantee an exact
solution. This section establishes error bounds and convergence guarantees
for the PRIK posterior mean under the second-order elliptic setting
of Eq. \eqref{eq:Linear-PDE}, supplemented by the Dirichlet boundary
condition. Complementary results characterize the limit of the normalized
posterior kernel and establish uniform decay of the posterior variance.

The analysis begins with an $H^{2}$-quasi-optimality bound for the
posterior mean in Eq. \eqref{eq:Predictor-Theta}. We then specialize
to kernel-section function spaces and derive an $H^{2}$-convergence
rate and a PDE-residual norm bound. The limit of the normalized posterior
kernel is also characterized as the variational Green-type kernel
associated with the PDE-residual inner product. Finally, we combine
a uniform diagonal bound with the variance-scale identity in Eq. \eqref{eq:Sigma2-Interpretation}
to establish uniform decay of the posterior variance in Eq. \eqref{eq:Var}.
Appendix \ref{sec:Numerical-Illustration-of-Theory} illustrates numerical
results for a screened Poisson equation on $\left[0,\pi\right]^{2}$,
using exact boundary observations and interior Matérn kernel sections.
For notational simplicity, we fix the kernel parameter $\boldsymbol{\theta}$
throughout this section, and suppress its dependence. For example,
we write $\hat{f}\left(\cdot\right)$ instead of $\hat{f}_{\boldsymbol{\theta}}\left(\cdot\right)$.

\subsection{Dirichlet Boundary Condition}

For the analysis in this section, we specialize the PDE in Eq. \eqref{eq:Linear-PDE}
to a second-order elliptic equation and impose the additional Dirichlet
boundary condition
\begin{equation}
\left.f\right|_{\mathbb{E}}=g.\label{eq:Dirichlet-Boundary}
\end{equation}
Unlike the general observation setting in Sections \ref{sec:Review}
and \ref{sec:PRIK}, we assume $\boldsymbol{X}_{l}\subset\mathbb{E}$
in this section.

For any finite set $\mathcal{S}\subseteq\mathcal{O}$ with at least
two points, where $\mathcal{O}\subseteq\mathbb{R}^{d}$ is a fixed
bounded set, the fill distance of $\mathcal{S}$ with respect to $\mathcal{O}$
is defined as 
\[
h_{\mathcal{S},\mathcal{O}}=\underset{\boldsymbol{x}\in\mathcal{O}}{\mathrm{sup}}\mathrm{dist}\left(\boldsymbol{x},\mathcal{S}\right),
\]
where $\mathrm{dist}\left(\boldsymbol{x},\mathcal{S}\right)=\min_{\boldsymbol{x}^{\prime}\in\mathcal{S}}\left\Vert \boldsymbol{x}-\boldsymbol{x}^{\prime}\right\Vert _{2}$,
and the separation distance of $\mathcal{S}$ is defined as
\[
q_{\mathcal{S}}=\frac{1}{2}\underset{\boldsymbol{x}\neq\boldsymbol{x}'}{\min}\left\Vert \boldsymbol{x}-\boldsymbol{x}^{\prime}\right\Vert _{2}\text{ for any }\boldsymbol{x},\boldsymbol{x}^{\prime}\in\mathcal{S}.
\]
For any function $u$ defined on $\mathcal{S}$, define the discrete
norm on $\mathcal{S}$ by $\left\Vert u\right\Vert _{\mathcal{S}}=\left\Vert u\right\Vert _{\ell^{2}\left(\mathcal{S}\right)}=\left(\sum_{\boldsymbol{x}\in\mathcal{S}}\left|u\left(\boldsymbol{x}\right)\right|^{2}\right)^{1/2}$.
Throughout, the notation $a\lesssim b$ means that there exists a
constant $C>0$, independent of the quantities of interest, such that
$a\leqslant Cb$. The notation $a\asymp b$ means that $a\lesssim b$
and $b\lesssim a$.

\subsection{Accuracy of the Posterior Mean\label{subsec:Results-Mean}}

In this subsection, we establish three results for the PRIK posterior
mean in Eq. \eqref{eq:Predictor-Theta}. First, we derive an $H^{2}$-quasi-optimality
bound on a general finite-dimensional function space. Second, we specialize
the analysis to kernel-section spaces and obtain an explicit $H^{2}$-convergence
rate. Third, we derive a corresponding bound for the PDE-residual
norm.

We first study the posterior mean on a general finite-dimensional
function space $\mathbb{V}_{p}\subset H^{2}\left(\Omega\right)$.
By Theorem \ref{thm:Interpolation+minPDE}, the PRIK posterior mean
in Eq. \eqref{eq:Predictor-Theta} minimizes the PDE-residual energy
among all functions in $\mathbb{V}_{p}$ that interpolate the observed
values. The following two assumptions allow us to translate this minimum-PDE-residual
property into an $H^{2}\left(\Omega\right)$ error bound.
\begin{assumption}
\label{assu:Bounded-Differential-Operator} (Boundedness of the Differential
Operator) The differential operator $\mathscr{D}:H^{2}\left(\Omega\right)\to L^{2}\left(\Omega\right)$
is bounded:
\[
\left\Vert \mathscr{D}u\right\Vert _{L^{2}\left(\Omega\right)}\lesssim\left\Vert u\right\Vert _{H^{2}\left(\Omega\right)},\quad u\in H^{2}\left(\Omega\right).
\]
\end{assumption}
\begin{assumption}
\label{assu:Stability-on-Wp} (Stability on $\mathbb{W}_{p}$) For
the fixed label-homogeneous subspace $\mathbb{W}_{p}$,
\[
\left\Vert w\right\Vert _{H^{2}\left(\Omega\right)}\lesssim\left\Vert \mathscr{D}w\right\Vert _{L^{2}\left(\Omega\right)},\quad w\in\mathbb{W}_{p}.
\]
\end{assumption}
Assumption \ref{assu:Bounded-Differential-Operator} holds, for example,
when the PDE coefficients $\left\{ a_{\boldsymbol{\alpha}}\left(\boldsymbol{x}\right)\right\} _{\left|\boldsymbol{\alpha}\right|\leqslant s}$
of $\mathscr{D}$ belong to $L^{\infty}\left(\Omega\right)$. Assumption
\ref{assu:Stability-on-Wp} can be satisfied by constructing $\mathbb{V}_{p}$
so that $\mathbb{W}_{p}$ contains no nonzero solution of $\mathscr{D}w=0$,
i.e., $\mathbb{W}_{p}\cap\mathrm{ker}\left(\mathscr{D}\right)=\left\{ 0\right\} $.
\begin{lem}
\label{lem:H2-Quasi-Optimality}($H^{2}$-Quasi-Optimality of the
Posterior Mean) Suppose that Assumptions \ref{assu:Bounded-Differential-Operator}
and \ref{assu:Stability-on-Wp} hold. Then the PRIK posterior mean
satisfies
\begin{equation}
\left\Vert \hat{f}-f\right\Vert _{H^{2}\left(\Omega\right)}\lesssim\underset{\substack{v\in\mathbb{V}_{p}\\
v\left(\boldsymbol{X}_{l}\right)=\boldsymbol{y}_{l}
}
}{\inf}\left\Vert v-f\right\Vert _{H^{2}\left(\Omega\right)}.\label{eq:H2-quasi-optimality}
\end{equation}
\end{lem}
Lemma \ref{lem:H2-Quasi-Optimality} shows that the PRIK posterior
mean is as accurate as the best function in $\mathbb{V}_{p}$ that
interpolates the observed values. In other words, the PRIK posterior
mean achieves the same $H^{2}$-convergence order as the best interpolation
function in $\mathbb{V}_{p}$. 

To obtain an explicit convergence rate from the $H^{2}$-quasi-optimality
bound in Eq. \eqref{eq:H2-quasi-optimality}, we impose additional
assumptions on the domain, kernel, refinement design, and boundary
regularity.
\begin{assumption}
\label{assu:Smoothness-of-Domain}(Smoothness of Domain) Assume that
the bounded domain $\Omega\subset\mathbb{R}^{d}$ has a piecewise
$C^{\lceil m\rceil}$ boundary $\mathbb{E}=\partial\Omega$ so that
$\Omega$ is a Lipschitz domain and satisfies an interior cone condition,
where $d\leqslant3$ and $m>2+d/2$.
\end{assumption}
Here, the $C^{\lceil m\rceil}$ boundary \citep[Appendix C.1]{evansPartialDifferentialEquations2010}
means that each piece of the boundary $\mathbb{E}$ is sufficiently
smooth to support the Sobolev estimates used in our analysis. Geometrically,
this smoothness means that the normal vector varies smoothly along
each piece. The term piecewise allows these pieces to meet at corners
and edges, where the normal vector may change abruptly. For example,
circles and spheres both have smooth boundaries, whereas rectangles
and boxes have piecewise smooth boundaries.
\begin{assumption}
\label{assu:Sobolev-Compatible-Kernel}(Sobolev-Compatible Kernel)
Assume that $k\left(\boldsymbol{x},\boldsymbol{x}^{\prime}\right)$
is a symmetric, translation-invariant positive-definite kernel of
the form $k\left(\boldsymbol{x},\boldsymbol{x}^{\prime}\right)=\Phi_{m}\left(\boldsymbol{x}-\boldsymbol{x}^{\prime}\right)$,
whose RKHS on $\Omega$ is $H^{m}\left(\Omega\right)$ with an equivalent
norm. The Fourier transform of $\Phi_{m}$ satisfies
\begin{equation}
c_{\Phi_{m}}\left(1+\left\Vert \boldsymbol{\omega}\right\Vert _{2}^{2}\right)^{-m}\leqslant\mathscr{F}\left[\Phi_{m}\right]\left(\boldsymbol{\omega}\right)\leqslant C_{\Phi_{m}}\left(1+\left\Vert \boldsymbol{\omega}\right\Vert _{2}^{2}\right)^{-m},\quad\boldsymbol{\omega}\in\mathbb{R}^{d},\label{eq:Matern-Fourier}
\end{equation}
for two constants $0<c_{\Phi_{m}}\leqslant C_{\Phi_{m}}<\infty$.
\end{assumption}
A standard example is the Matérn kernel with
\[
\Phi_{m}\left(\boldsymbol{x}-\boldsymbol{x}^{\prime}\right)=\frac{2^{1-\nu}}{\Gamma\left(\nu\right)}\left\Vert \boldsymbol{x}-\boldsymbol{x}^{\prime}\right\Vert _{2}^{\nu}K_{\nu}\left(\left\Vert \boldsymbol{x}-\boldsymbol{x}^{\prime}\right\Vert _{2}\right),\quad\nu=m-\frac{d}{2},
\]
where $\Gamma\left(\cdot\right)$ is the gamma function and $K_{\nu}\left(\cdot\right)$
is the modified Bessel function of the second kind. 
\begin{assumption}
\label{assu:Refinement-Design}(Separated and Balanced Refinement)
Consider a refinement family $\left\{ \left(\boldsymbol{X}_{l,h},\boldsymbol{X}_{\mathrm{int},h}\right)\right\} _{h}$,
where $\boldsymbol{X}_{\mathrm{int},h}\subset\Omega$ denotes the
interior points at the refinement level $h=h_{\boldsymbol{X}_{\mathrm{int},h},\Omega}$,
and $\boldsymbol{X}_{l,h}\subset\mathbb{E}$ denotes the labeled points
at the refinement level $h$. Assume that $h_{\boldsymbol{X}_{\mathrm{int},h},\Omega}\asymp q_{\boldsymbol{X}_{\mathrm{int},h}}\rightarrow0$.
When $d\geqslant2$, we further assume that $\boldsymbol{X}_{l,h}$
satisfy $h_{\boldsymbol{X}_{l,h},\mathbb{E}}\asymp q_{\boldsymbol{X}_{l,h}}\asymp h_{\boldsymbol{X}_{\mathrm{int},h},\Omega}$;
while we take $\boldsymbol{X}_{l,h}=\mathbb{E}$ when $d=1$.
\end{assumption}
For each refinement level, we set $\mathbb{V}_{p_{h}}=\mathcal{H}_{h}$,
where $p_{h}=\left|\boldsymbol{X}_{{\rm int},h}\right|$, and 
\[
\mathcal{H}_{h}=\mathrm{span}\left\{ k\left(\cdot,\boldsymbol{z}\right):\boldsymbol{z}\in\boldsymbol{X}_{\mathrm{int},h}\right\} .
\]
Under Assumptions \ref{assu:Smoothness-of-Domain}--\ref{assu:Refinement-Design},
standard kernel approximation results give
\[
\left\Vert \mathcal{I}_{\boldsymbol{X}_{\mathrm{int},h}}u-u\right\Vert _{H^{2}\left(\Omega\right)}\lesssim h^{m-2}\left\Vert u\right\Vert _{H^{m}\left(\Omega\right)},\quad u\in H^{m}\left(\Omega\right),
\]
where $\mathcal{I}_{\boldsymbol{X}_{\mathrm{int},h}}u\in\mathcal{H}_{h}$
denotes the kernel interpolation function  of $u$ at $\boldsymbol{X}_{\mathrm{int},h}$. 
\begin{assumption}
\label{assu:classical-solution-existence}(Existence of a Classical
Solution) Assume that the PDE in Eq. \eqref{eq:Linear-PDE}, supplemented
by the Dirichlet boundary condition in Eq. \eqref{eq:Dirichlet-Boundary},
admits a solution $f\in H^{m}\left(\Omega\right)$.
\end{assumption}
\begin{assumption}
\label{assu:Boundary-Regularity-Estimate} (Boundary Regularity Estimate)
There exists a constant $C_{B}>0$, depending only on $\Omega$ and
$\mathscr{D}$, such that 
\begin{equation}
\left\Vert u\right\Vert _{H^{2}\left(\Omega\right)}\leqslant C_{B}\left(\left\Vert \mathscr{D}u\right\Vert _{L^{2}\left(\Omega\right)}+\left\Vert u\right\Vert _{H^{3/2}\left(\mathbb{E}\right)}\right),\quad u\in H^{2}\left(\Omega\right).\label{eq:Boundary-Regularity}
\end{equation}
\end{assumption}
\begin{assumption}
\label{assu:Separated-Stability-Smallness} (Separated-Design Stability
Scale) When $d\geqslant2$, for sufficiently small $h$, 
\[
C_{B}C_{S}C_{I}\left(\frac{h_{\boldsymbol{X}_{l,h},\mathbb{E}}}{q_{\boldsymbol{X}_{\mathrm{int},h}}}\right)^{m-2}<\frac{1}{2},
\]
where $C_{S}>0$ and $C_{I}>0$ are the constants in the boundary
sampling inequality and the inverse inequality given in Appendix \ref{subsec:Proof-Posterior-Mean}.
\end{assumption}
It is worth noting that Assumption \ref{assu:Separated-Stability-Smallness}
imposes a prescribed smallness threshold on the design ratio, rather
than merely an asymptotic bound \citep[Section 1.5]{buffa2022mathematical}
of the form $h_{\boldsymbol{X}_{l,h},\mathbb{E}}\lesssim q_{\boldsymbol{X}_{\mathrm{int},h}}$.
\begin{assumption}
\label{assu:Stable-Discrete-Boundary-Correction} (Stable Discrete
Boundary Correction) For every $\boldsymbol{b}\in\mathbb{R}^{N_{l,h}}$,
\[
\inf_{r_{h}\left(\cdot\right)\in\mathcal{H}_{h}\text{ and }r_{h}\left(\boldsymbol{X}_{l,h}\right)=\boldsymbol{b}}\left\Vert r_{h}\right\Vert _{H^{2}\left(\Omega\right)}\lesssim h^{d/2-2}\left\Vert \boldsymbol{b}\right\Vert _{2},
\]
where $N_{l,h}=\left|\boldsymbol{X}_{l,h}\right|$. 
\end{assumption}
Assumption \ref{assu:Stable-Discrete-Boundary-Correction} allows
this approximation to be modified so that it also interpolates the
observed boundary values. It is worth noting that the scaling $h^{d/2-2}$
is not imposed arbitrarily: it can be verified using the interior
Lagrange functions and is attained under an explicit Matérn interior
design; see Appendix \ref{sec:Stable-Discrete-Boundary-Correction-Justification}.

We are now ready to establish the convergence rate of the posterior
mean.
\begin{thm}
\label{thm:Posterior-Mean-Convergence-Rate}(Posterior-Mean Convergence
Rate) Suppose that Assumption \ref{assu:Bounded-Differential-Operator}
and Assumptions \ref{assu:Smoothness-of-Domain}--\ref{assu:Stable-Discrete-Boundary-Correction}
hold. Then the PRIK posterior mean $\hat{f}_{h}$ on $\mathcal{H}_{h}$
satisfies
\begin{equation}
\left\Vert \hat{f}_{h}-f\right\Vert _{H^{2}\left(\Omega\right)}\lesssim h^{m-2}\left\Vert f\right\Vert _{H^{m}\left(\Omega\right)}.\label{eq:Predictive-Mean-Convergence-Rate}
\end{equation}
\end{thm}
Theorem \ref{thm:Posterior-Mean-Convergence-Rate} shows that the
PRIK posterior mean inherits the $H^{2}$-approximation rate of the
underlying kernel interpolation function. 

To illustrate this theoretical convergence rate, we consider the screened
Poisson example in Appendix \ref{sec:Numerical-Illustration-of-Theory},
which uses Matérn smoothness parameters $\nu=9/2$ and $\nu=11/2$.
The corresponding theoretical $H^{2}\left(\Omega\right)$-error upper
bounds are of orders $h^{7/2}$ and $h^{9/2}$, respectively. Figure
\ref{fig:H2-error-convergence} shows the $H^{2}\left(\Omega\right)$
error as the interior and boundary designs are refined. For certain
kernel parameter, the observed error decay is close to the corresponding
reference rate.

The $H^{2}\left(\Omega\right)$-error bound in Theorem \ref{thm:Posterior-Mean-Convergence-Rate}
also provides a convergence guarantee for the PDE-residual norm. Since
$\mathscr{D}f=0$ and $\mathscr{D}$ is bounded from $H^{2}\left(\Omega\right)$
to $L^{2}\left(\Omega\right)$, we obtain the following corollary.
\begin{cor}
\label{cor:PDE-Residual-Convergence-Rate}(Convergence Rate of the
PDE-residual Norm) Under the assumptions of Theorem \ref{thm:Posterior-Mean-Convergence-Rate},
the PRIK posterior mean satisfies
\begin{equation}
\left\Vert \mathscr{D}\hat{f}_{h}\right\Vert _{L^{2}\left(\Omega\right)}\lesssim h^{m-2}\left\Vert f\right\Vert _{H^{m}\left(\Omega\right)}.\label{eq:PDE-Residual-Convergence-Rate}
\end{equation}
\end{cor}
Corollary \ref{cor:PDE-Residual-Convergence-Rate} shows that the
PDE-residual norm admits an upper bound of the same order as the $H^{2}$-approximation
error. For the screened Poisson example of Appendix \ref{sec:Numerical-Illustration-of-Theory},
Figure \ref{fig:PDE-residual-norm-convergence} shows that the PDE-residual
norm exhibits a decay patterns similar to that of the $H^{2}\left(\Omega\right)$
error, consistent with the theoretical $h^{m-2}$ upper bounds in
Eqs. \eqref{eq:Predictive-Mean-Convergence-Rate} and \eqref{eq:PDE-Residual-Convergence-Rate}.

\subsection{Normalized Posterior Kernel and Posterior Variance\label{subsec:Results-Var}}

We next study the normalized posterior kernel and the posterior variance
in Section \ref{subsec:Prediction-Distribution} under refinement
of the kernel-section space. For each $h$, let $\tilde{k}_{h}\left(\cdot,\cdot\right)$
denote the PDE-residual kernel on $\mathcal{H}_{h}$. At refinement
level $h$, the normalized posterior kernel in Eq. \eqref{eq:Normalized-Var}
is denoted by 

\begin{equation}
c_{h}\left(\boldsymbol{x},\boldsymbol{x}^{\prime}\right)=\tilde{k}_{h}\left(\boldsymbol{x},\boldsymbol{x}^{\prime}\right)-\tilde{k}_{h}\left(\boldsymbol{x},\boldsymbol{X}_{l,h}\right)\tilde{k}_{h}\left(\boldsymbol{X}_{l,h},\boldsymbol{X}_{l,h}\right)^{-1}\tilde{k}_{h}\left(\boldsymbol{X}_{l,h},\boldsymbol{x}^{\prime}\right).\label{eq:Normalized-Posterior-Kernel}
\end{equation}
with the kernel parameter $\boldsymbol{\theta}$ suppressed.

To characterize its limiting behavior, define the homogeneous-boundary
function space $\mathbb{W}=\left\{ w\in H^{2}\left(\Omega\right):\left.w\right|_{\mathbb{E}}=0\right\} $.
Assumptions \ref{assu:Bounded-Differential-Operator} and \ref{assu:Boundary-Regularity-Estimate}
imply
\begin{equation}
\left\Vert \mathscr{D}w\right\Vert _{L^{2}\left(\Omega\right)}\asymp\left\Vert w\right\Vert _{H^{2}\left(\Omega\right)},\quad w\in\mathbb{W}.\label{eq:Continuous-Energy-Norm-Equivalence}
\end{equation}
Hence, $\left<\cdot,\cdot\right>_{\mathscr{D}}=\left<\mathscr{D}\cdot,\mathscr{D}\cdot\right>_{L^{2}\left(\Omega\right)}$
defines an inner product on $\mathbb{W}$, and its induced norm is
equivalent to the $H^{2}\left(\Omega\right)$-norm. Since $d\leqslant3$,
the embedding $H^{2}\left(\Omega\right)\hookrightarrow C\left(\bar{\Omega}\right)$
is compact. In particular, point evaluation is continuous on $\mathbb{W}$.
Since $\mathbb{W}$ is a closed subspace of $H^{2}\left(\Omega\right)$,
$\left(\mathbb{W},\left<\cdot,\cdot\right>_{\mathscr{D}}\right)$
is an RKHS. We denote its reproducing kernel by $G\left(\cdot,\cdot\right)$,
which satisfies 
\begin{equation}
\left<G\left(\cdot,\boldsymbol{\xi}\right),w\right>_{\mathscr{D}}=w\left(\boldsymbol{\xi}\right),\quad w\in\mathbb{W},\quad\boldsymbol{\xi}\in\Omega.\label{eq:Green-Reproducing-Identity}
\end{equation}
This reproducing identity characterizes $G\left(\cdot,\cdot\right)$
as the variational Green-type kernel associated with the PDE-residual
inner product.

To establish the convergence of the normalized posterior kernel \eqref{eq:Normalized-Posterior-Kernel},
we impose the following additional assumption.
\begin{assumption}
\label{assu:Dense-Regular-Core}(Dense Regular Core) There exists
a linear subspace
\[
\mathbb{W}_{\mathrm{reg}}\subseteq\mathbb{W}\cap H^{m}\left(\Omega\right),
\]
which is dense in $\mathbb{W}$ under the $H^{2}\left(\Omega\right)$-norm.
\end{assumption}
Assumption \ref{assu:Dense-Regular-Core} allows functions in $\mathbb{W}$
to be approximated by sufficiently smooth functions. The following
theorem shows that $c_{h}\left(\cdot,\boldsymbol{\xi}\right)$ converges
to $G\left(\cdot,\boldsymbol{\xi}\right)$ for every fixed $\boldsymbol{\xi}\in\Omega$.
\begin{thm}
\label{thm:posterior-kernel-convergence} (Convergence of the Normalized
Posterior Kernel) Suppose that Assumptions \ref{assu:Bounded-Differential-Operator},
\ref{assu:Smoothness-of-Domain}--\ref{assu:Refinement-Design},
and \ref{assu:Boundary-Regularity-Estimate}--\ref{assu:Dense-Regular-Core}
hold. Then, for every fixed $\boldsymbol{\xi}\in\Omega$, 
\begin{equation}
\left\Vert c_{h}\left(\cdot,\boldsymbol{\xi}\right)-G\left(\cdot,\boldsymbol{\xi}\right)\right\Vert _{C\left(\bar{\Omega}\right)}\rightarrow0\quad\text{as }h\to0.\label{eq:N.Posterior-Kernel-Convergence}
\end{equation}
\end{thm}
For the screened Poisson example in Appendix \ref{sec:Numerical-Illustration-of-Theory},
Figure \ref{fig:posterior-kernel-heatmaps} compares $c_{h}\left(\cdot,\boldsymbol{\xi}\right)$
with the variational Green-type kernel $G\left(\cdot,\boldsymbol{\xi}\right)$,
where $\boldsymbol{\xi}=\left[\pi/3,\pi/2\right]^{\top}$. The results
show that both Matérn kernels accurately reproduce the spatial profile
of the variational Green-type kernel.

The preceding Therorem \ref{thm:posterior-kernel-convergence} characterizes
the limiting behavior of the normalized posterior kernel, but this
convergence alone does not imply decay of the posterior variance.
The latter also depends on the estimated variance scale. Let $\hat{\sigma}_{h}^{2}$
denote the variance-scale estimate obtained from Eq. \eqref{eq:Sigma2}
at refinement level $h$, with $\boldsymbol{\theta}$ held fixed.
The posterior variance at refinement level $h$ can be written as
\[
\hat{\sigma}_{f,h}^{2}\left(\boldsymbol{x}\right)=\hat{\sigma}_{h}^{2}c_{h}\left(\boldsymbol{x},\boldsymbol{x}\right).
\]
It can be shown that the diagonal $c_{h}\left(\boldsymbol{x},\boldsymbol{x}\right)$
is uniformly bounded over $\bar{\Omega}$. Combining this with Eq.
\eqref{eq:Sigma2-Interpretation} and Corollary \ref{cor:PDE-Residual-Convergence-Rate}
gives the following result.
\begin{cor}
\label{cor:Decay-Predictive-Sigma2}(Uniform Decay of the Posterior
Variance) Under the assumptions of Theorem \ref{thm:Posterior-Mean-Convergence-Rate},
\[
\underset{\boldsymbol{x}\in\bar{\Omega}}{\mathrm{sup}}\hat{\sigma}_{f,h}^{2}\left(\boldsymbol{x}\right)\lesssim\frac{h^{2m-4}}{N_{l,h}}\left\Vert f\right\Vert _{H^{m}\left(\Omega\right)}^{2}.
\]
\end{cor}
Corollary \ref{cor:Decay-Predictive-Sigma2} provides a uniform upper
bound for the posterior variance.

\section{Simulation Studies for PDE Solving \label{sec:Numerical-PDE-Solving} }

This section evaluates PRIK on PDE solving problems beyond the elliptic
settings considered above. The purpose is twofold. First, we examine
whether the PDE-residual-kernel construction remains effective for
non-elliptic PDEs. Second, we compare PRIK with physics-informed neural
networks (PINNs) and PIKL on benchmark problems for which reference
results have been reported in the literature \citep{krishnapriyanCharacterizingPossibleFailure2021,wang2022and,doumechePhysicsinformedKernelLearning2025}.

We consider two classes of hyperbolic PDEs: a one-dimensional wave
equation with mixed initial-boundary conditions and a periodic convection
equation. These examples have been used in prior studies of PINNs
and PIKL, and therefore provide a useful basis for comparison \citep{krishnapriyanCharacterizingPossibleFailure2021,wang2022and,doumechePhysicsinformedKernelLearning2025}.
For PRIK, we generate $p=1000$ interior points $\boldsymbol{X}_{\mathrm{int}}\in\mathbb{R}^{p\times d}$
using a maximin design and construct the function space using the
kernel sections $\left\{ k_{\boldsymbol{\theta}}\left(\cdot,\boldsymbol{x}\right):\boldsymbol{x}\in\boldsymbol{X}_{\mathrm{int}}\right\} $.
The number of labeled observations $N_{l}$ varies across experiments
and is reported separately for each case. The performance is assessed
by the $L^{2}$ relative error, $\left\Vert \hat{f}-f\right\Vert _{L^{2}\left(\Omega\right)}/\left\Vert f\right\Vert _{L^{2}\left(\Omega\right)}$.

\subsection{Wave Equation with Mixed Initial-Boundary Conditions\label{Wave-Equation}}
\begin{example}
(Wave Equation with Mixed Initial-Boundary Conditions) We consider
the one-dimensional wave equation on $\Omega=\left[0,1\right]\times\left[0,1\right]$:
\[
\begin{cases}
\mathscr{D}\left(f\right)=\partial_{t,t}^{2}f-4\partial_{x,x}^{2}f=0, & \left(t,x\right)\in\Omega,\\
f\left(0,x\right)=\sin\left(\pi x\right)+\sin\left(4\pi x\right)/2, & \forall x\in\left[0,1\right],\\
\partial_{t}f\left(0,x\right)=0, & \forall x\in\left[0,1\right],\\
f\left(t,0\right)=f\left(t,1\right)=0, & \forall t\in\left[0,1\right].
\end{cases}
\]
The exact solution is 
\[
f\left(t,x\right)=\sin\left(\pi x\right)\cos\left(2\pi t\right)+\sin\left(4\pi x\right)\cos\left(8\pi t\right)/2.
\]
\end{example}
This example is a hyperbolic PDE with mixed initial-boundary conditions.
The solution contains multiple oscillatory components, and therefore
provides a test case for evaluating whether the PDE-residual kernel
can be effective beyond the elliptic settings considered earlier.
Let $U_{i}$ be sampled from the uniform distribution on $\left[0,1\right]$.
The labeled data $\left\{ \left(\boldsymbol{x}_{i},y_{i}\right)\right\} _{i=1}^{N_{l}}$
are constructed such that
\begin{itemize}
\item if $1\leqslant i\leqslant\lfloor N_{l}/4\rfloor$, then $\boldsymbol{x}_{i}=\left[0,U_{i}\right]^{\top}$
and $y_{i}=\sin\left(\pi U_{i}\right)+\sin\left(4\pi U_{i}\right)/2$,
\item if $\lfloor N_{l}/4\rfloor+1\leqslant i\leqslant2\lfloor N_{l}/4\rfloor$,
then $\boldsymbol{x}_{i}=\left[U_{i},0\right]^{\top}$ and $y_{i}=0$,
\item if $2\lfloor N_{l}/4\rfloor+1\leqslant i\leqslant3\lfloor N_{l}/4\rfloor$,
then $\boldsymbol{x}_{i}=\left[U_{i},1\right]^{\top}$ and $y_{i}=0$,
\item if $3\lfloor N_{l}/4\rfloor+1\leqslant i\leqslant N_{l}$, then $\boldsymbol{x}_{i}=\left[1/N_{l},U_{i}\right]^{\top}$,
and $y_{i}$ is assigned by a second-order Taylor approximation at
$t=0$:
\begin{align*}
y_{i} & \approx f\left(\frac{1}{N_{l}},U_{i}\right)\approx f\left(0,U_{i}\right)+\frac{1}{2N_{l}^{2}}\partial_{t,t}^{2}f\left(0,U_{i}\right)=f\left(0,U_{i}\right)+\frac{2}{N_{l}^{2}}\partial_{x,x}^{2}f\left(0,U_{i}\right)\\
 & =\left(1-\frac{2\pi^{2}}{N_{l}^{2}}\right)\sin\left(\pi U_{i}\right)+\left(\frac{1}{2}-\frac{16\pi^{2}}{N_{l}^{2}}\right)\sin\left(4\pi U_{i}\right).
\end{align*}
\end{itemize}
For PRIK in this experiment, we set $N_{l}=200$ and use the Gaussian
kernel in Eq. \eqref{eq:Gauss-Kernel} as the base kernel. Table \ref{tab:Results-Wave}
reports the relative $L^{2}$ errors and model sizes. The results
for vanilla PINNs, neural tangent kernel (NTK)-optimized PINNs, and
PIKL are taken from Table 2 of \citet{doumechePhysicsinformedKernelLearning2025},
where the two PINN results in that table were originally reported
by \citet{wang2022and}. In this table, model size refers to the number
of trainable weights for PINN-based methods and to the number of basis
functions used in the kernel representation for PIKL and PRIK. PRIK
attains a smaller relative error than the reported PIKL result and
the two PINN-based reference results, while using substantially fewer
labeled observations. Compared with PIKL, PRIK reduces the $L^{2}$
relative error by roughly one order of magnitude while using fewer
basis functions in the kernel representation.
\begin{table}
\centering{}\caption{Performance of PINN/PIKL/PRIK methods for solving the wave equation.
The form $a\pm b$ denotes the mean and standard deviation over 10
repetitions.\label{tab:Results-Wave}}
\begin{tabular}{ccccc}
\toprule 
 & \shortstack{Vanilla\\PINNs} & \shortstack{NTK-optimized\\PINNs} & PIKL & PRIK\tabularnewline
\midrule 
$L^{2}$ relative error & $4.52\times10^{-1}$ & $1.73\times10^{-3}$ & $\left(8.70\pm0.08\right)\times10^{-4}$ & $\left(7.07\pm2.55\right)\times10^{-5}$\tabularnewline
Training data ($N_{l}$) & $2.4\times10^{6}$ & $2.4\times10^{6}$ & $10^{5}$ & $200$\tabularnewline
Model size & $5.03\times10^{5}$ & $5.03\times10^{5}$ & $1.68\times10^{3}$ & $1.0\times10^{3}$\tabularnewline
\bottomrule
\end{tabular}
\end{table}

\subsection{Periodic Convection Equation with a Sinusoidal Initial Profile\label{subsec:Periodic-Convection-Sin}}
\begin{example}
(Periodic Convection Equation with a Sinusoidal Initial Profile) We
consider the periodic convection equation on $\Omega=\left[0,1\right]\times\left[0,2\pi\right]$:
\[
\begin{cases}
\mathscr{D}\left(f\right)=\partial_{t}f+\beta\partial_{x}f=0, & \left(t,x\right)\in\Omega,\\
f\left(0,x\right)=\sin x, & \forall x\in\left[0,2\pi\right],\\
f\left(t,0\right)=f\left(t,2\pi\right), & \forall t\in\left[0,1\right],
\end{cases}
\]
where $\beta>0$ is the constant convection speed. The exact solution
is 
\[
f\left(t,x\right)=\sin\left(x-\beta t\right).
\]
\end{example}
This example is a transport-dominated hyperbolic PDE. As $\beta$
increases, the solution becomes increasingly oscillatory in the time
direction, making the problem more challenging for physics-informed
learning methods. For PRIK, we use the periodic kernel in Eq. \eqref{eq:Periodic-Kernel}.
In this example, $x_{1}=t$ and $x_{2}=x$. The spatial period is
set to $T_{2}=2\pi$ according to the periodic boundary condition,
and the temporal period is set to $T_{1}=2\pi/\beta$, following the
standard characteristic interpretation of first-order transport equations
\citep{evansPartialDifferentialEquations2010}. The PDE-residual Gram
matrix $\boldsymbol{A}$ is computed by separable Gauss--Legendre
quadrature \citep{davis2007methods}.

For PRIK in this experiment, we set $N_{l}=100$, following the PIKL
setting in \citet{doumechePhysicsinformedKernelLearning2025}. Table
\ref{tab:Results-Convection-Sin} reports the $L^{2}$ relative errors
for $\beta=20,30,40$. The results for vanilla PINNs, curriculum-trained
PINNs, and PIKL are taken from Table 1 of \citet{doumechePhysicsinformedKernelLearning2025},
where the two PINN results in that table were originally reported
by \citet{krishnapriyanCharacterizingPossibleFailure2021}. PRIK achieves
consistently smaller $L^{2}$ relative errors than the reported PINN-based
reference results across all three values of $\beta$. Compared with
PIKL, PRIK gives smaller errors for $\beta=20$ and $\beta=30$, and
remains of the same order while attaining a smaller error for $\beta=40$.

\begin{table}
\centering{}\caption{$L^{2}$ relative error of PINN/PIKL/PRIK methods for solving the
periodic convection equation with a sinusoidal initial profile. The
form $a\pm b$ denotes the mean and standard deviation over 10 repetitions.\label{tab:Results-Convection-Sin}}
\begin{tabular}{ccccc}
\toprule 
 & \shortstack{Vanilla\\PINNs} & \shortstack{Curriculum-trained\\PINNs} & PIKL & \shortstack{PRIK\\(Periodic)}\tabularnewline
\midrule 
$\beta=20$ & $7.50\times10^{-1}$ & $9.84\times10^{-3}$ & $\left(1.56\pm3.46\right)\times10^{-8}$ & $\left(1.11\pm0.38\right)\times10^{-9}$\tabularnewline
$\beta=30$ & $8.97\times10^{-1}$ & $2.02\times10^{-2}$ & $\left(0.91\pm2.20\right)\times10^{-7}$ & $\left(1.53\pm0.37\right)\times10^{-9}$\tabularnewline
$\beta=40$ & $9.61\times10^{-1}$ & $5.33\times10^{-2}$ & $\left(7.31\pm6.44\right)\times10^{-9}$ & $\left(1.69\pm0.48\right)\times10^{-9}$\tabularnewline
\bottomrule
\end{tabular}
\end{table}

\subsection{Periodic Convection Equation with an Exponential Periodic Initial
Profile\label{subsec:Periodic-Convection-ExpSin}}
\begin{example}
(Periodic Convection Equation with an Exponential Periodic Initial
Profile) We consider the periodic convection equation on $\Omega=\left[0,1\right]\times\left[0,2\pi\right]$:
\[
\begin{cases}
\mathscr{D}\left(f\right)=\partial_{t}f+\beta\partial_{x}f=0, & \left(t,x\right)\in\Omega\\
f\left(0,x\right)=\mathrm{e}^{-\sin x}, & \forall x\in\left[0,2\pi\right],\\
f\left(t,0\right)=f\left(t,2\pi\right), & \forall t\in\left[0,1\right].
\end{cases}
\]
The exact solution is 
\[
f\left(t,x\right)=\mathrm{e}^{-\sin\left(x-\beta t\right)}.
\]
\end{example}
This example uses the same periodic convection equation as in Section
\ref{subsec:Periodic-Convection-Sin}, but replaces the sinusoidal
initial profile by the more complex periodic profile $\mathrm{e}^{-\sin x}$.
Unlike the single-mode sinusoidal profile, $\mathrm{e}^{-\sin x}$
has a non-bandlimited Fourier expansion and contains infinitely many
periodic components. This setting is therefore used to examine a regime
in which a finite Fourier-mode representation may become restrictive,
while a kernel-based construction can incorporate the periodic structure
through the appropriate choice of kernel function.

We use the same periodic kernel in Eq. \eqref{eq:Periodic-Kernel},
period specification, PDE-residual Gram-matrix construction and number
of labeled observations $N_{l}$ as in Section \ref{subsec:Periodic-Convection-Sin}.
For PIKL, the results are obtained from our implementation based on
the public code of \citet{doumechePhysicsinformedKernelLearning2025}\footnote{\url{https://github.com/NathanDoumeche/numerical_PIML_kernel}}.
We keep the original PDE-solving settings used in that code, except
the initial-boundary conditions and exact solution.

Table \ref{tab:Results-Convection-ExpSin} reports the $L^{2}$ relative
errors for $\beta=20,30,40$. In the sinusoidal case in Section \ref{subsec:Periodic-Convection-Sin},
PIKL performs extremely well, with $L^{2}$ relative errors around
$10^{-8}$, which is consistent with the fact that the solution is
essentially single-mode under an appropriate Fourier representation.
For the exponential periodic profile, however, the PIKL error increases
rapidly as $\beta$ grows, from $2.66\times10^{-3}$ at $\beta=20$
to $1.31\times10^{-1}$ at $\beta=40$. This behavior reflects the
limitation of the Fourier truncation used by PIKL in this comparison:
for the considered values of $\beta$, the solution exhibits high
temporal oscillations, and the non-bandlimited initial profile cannot
be captured accurately by a small truncated Fourier basis.

By contrast, PRIK retains errors at the level of $10^{-9}$--$10^{-8}$
across all three values of $\beta$, which are substantially smaller
than those of PIKL. The comparison indicates that, for this multi-component
periodic profile, an appropriate kernel choice is critical for representing
the solution structure, rather than merely relying on a fixed physics-informed
representation. PRIK can utilize a periodic base kernel tailored to
the solution characteristic while retaining the PDE-residual kernel
construction, which provides a more flexible representation than the
fixed Fourier truncation used by PIKL in this experiment.

\begin{table}
\centering{}\caption{$L^{2}$ relative error of PIKL/PRIK methods for solving the periodic
convection equation with an exponential periodic initial profile.
The form $a\pm b$ denotes the mean and standard deviation over 10
repetitions.\label{tab:Results-Convection-ExpSin}}
\begin{tabular}{ccc}
\toprule 
 & PIKL & PRIK (Periodic)\tabularnewline
\midrule 
$\beta=20$ & $\left(2.66\pm0.05\right)\times10^{-3}$ & $\left(7.31\pm3.02\right)\times10^{-9}$\tabularnewline
$\beta=30$ & $\left(2.15\pm0.05\right)\times10^{-2}$ & $\left(9.63\pm6.20\right)\times10^{-9}$\tabularnewline
$\beta=40$ & $\left(1.31\pm0.02\right)\times10^{-1}$ & $\left(1.12\pm0.30\right)\times10^{-8}$\tabularnewline
\bottomrule
\end{tabular}
\end{table}

\section{Conclusion \label{sec:Conclusion}}

This paper develops PRIK to make PDE-residual energy control an intrinsic
part of GP modeling for linear PDE solving. Rather than enforcing
the governing equation only through finite pointwise PDE constraints
or by adding the PDE-residual energy as an external penalty, PRIK
directly uses the PDE-residual energy to define a PDE-residual kernel
from a user-specified base kernel. Among all functions in $\mathbb{V}_{p}$
interpolating the labeled data, the resulting posterior mean  minimizes
the PDE-residual energy within the function space. In addition, the
variance-scale estimate and the profile likelihood admit interpretations
in terms of PDE-residual energy and model complexity. Under the stated
assumptions, we show that the posterior mean is $H^{2}$-quasi-optimal
and converges at the corresponding kernel approximation rate, while
the normalized posterior kernel converges to a variational Green-type
kernel associated with the PDE-residual inner product. The numerical
examples indicate that PRIK can achieve smaller errors than PINN-based
methods and competitive or smaller errors than PIKL, especially when
a problem-adapted base kernel is used to represent solution structures
such as periodicity.

Several directions remain for future work. First, the present paper
assumes that the linear differential operator is known. When the PDE
contains unknown parameters, the likelihood formulation in Eq. \eqref{eq:MLE-Opt}
can in principle be extended to jointly optimize the kernel hyperparameters
and the PDE parameters. Such an extension, however, would require
a separate treatment of identifiability, numerical stability, and
statistical uncertainty for the estimated PDE parameters. Second,
the interior points are currently selected independently by a maximin
design, which is problem-independent and may not be uniformly efficient
across different PDEs. A PDE-residual-aware design rule would therefore
be a meaningful direction, as it could adapt the placement of interior
kernel centers and select centers associated with the largest expected
reduction in PDE-residual energy. Third, extending PRIK to nonlinear
PDEs is a natural direction for future work. A key goal is to develop
a residual-informed kernel construction that can accommodate nonlinear
PDEs. 

\bigskip{}

\begin{spacing}{1}

\bibliographystyle{asa}
\bibliography{PRIK}

\end{spacing}

\appendix
\setcounter{thm}{0}
\setcounter{lem}{0}
\setcounter{prop}{0}
\setcounter{assumption}{0}
\setcounter{cor}{0}

\renewcommand{\thethm}{A\arabic{thm}}
\renewcommand{\thelem}{A\arabic{lem}}
\renewcommand{\theprop}{A\arabic{prop}}
\renewcommand{\theassumption}{A\arabic{assumption}}
\renewcommand{\thecor}{A\arabic{cor}}

\section{Computation and Properties of the PDE-Residual Gram Matrix\label{sec:PDE-Gram}}

For simplicity, the dependence on $\boldsymbol{\theta}$ is omitted
throughout the Appendix when there is no confusion. In particular,
$k\left(\cdot,\cdot\right)$ and $\tilde{k}\left(\cdot,\cdot\right)$
denote $k_{\boldsymbol{\theta}}\left(\cdot,\cdot\right)$ and $\tilde{k}_{\boldsymbol{\theta}}\left(\cdot,\cdot\right)$,
respectively.

Recall the kernel-section construction of the function space used
in Sections \ref{subsec:Model-Formulation} and \ref{subsec:PDE-Residual_Kernel}
of the main paper. Let $\boldsymbol{Z}=\left[\boldsymbol{z}_{1},\cdots,\boldsymbol{z}_{p}\right]^{\top}\in\mathbb{R}^{p\times d}$
contain pairwise distinct interior kernel centers $\boldsymbol{z}_{j}\in\Omega$.
Assume that the kernel sections $\left\{ k\left(\cdot,\boldsymbol{z}_{j}\right)\right\} _{j=1,\cdots,p}$
are linearly independent, so that they form a basis of the $p$-dimensional
function space $\mathbb{V}_{p}$. 

The PDE-residual Gram matrix is
\begin{equation}
\boldsymbol{A}=\left<\mathscr{D}k\left(\cdot,\boldsymbol{Z}\right),\mathscr{D}k\left(\cdot,\boldsymbol{Z}\right)\right>_{L^{2}\left(\Omega\right)}=\left[\left<\mathscr{D}k\left(\cdot,\boldsymbol{z}_{i}\right),\mathscr{D}k\left(\cdot,\boldsymbol{z}_{j}\right)\right>_{L^{2}\left(\Omega\right)}\right]_{i,j\in\left\{ 1,2,\cdots,p\right\} },\label{eq:PDE-Gram-Definition-Appen}
\end{equation}
where $k\left(\cdot,\boldsymbol{Z}\right)=\left[k\left(\cdot,\boldsymbol{z}_{1}\right),\cdots,k\left(\cdot,\boldsymbol{z}_{p}\right)\right]$.

In the main text, we assume that the PDE-residual Gram matrix $\boldsymbol{A}$
is invertible. This section provides several conditions under which
the invertibility assumption on $\boldsymbol{A}$ holds and then discusses
the computation of $\boldsymbol{A}$. For the Gaussian and Matérn
kernels considered here, positive definiteness is equivalent to the
non-vanishing of the derivative kernel sections $\left\{ \mathscr{D}k\left(\cdot,\boldsymbol{z}_{j}\right)\right\} _{j=1}^{p}$.
We also derive explicit formulas for $\boldsymbol{A}$ for the Gaussian
kernel with constant-coefficient operators and discuss numerical evaluation
strategies for other kernel--operator settings.

\subsection{Invertibility for Gaussian and Matérn PDE-Residual Matrices}

This section mainly considers two stationary kernels. The first kernel
is the Gaussian kernel 
\begin{equation}
k\left(\boldsymbol{x},\boldsymbol{x}^{\prime}\right)=\exp\left(-\left\Vert \boldsymbol{\varTheta}\left(\boldsymbol{x}-\boldsymbol{x}^{\prime}\right)\right\Vert _{2}^{2}\right)=\exp\left(-\sum_{i=1}^{d}\theta_{i}^{2}\left(x_{i}-x_{i}^{\prime}\right)^{2}\right),\label{eq:Gauss-Kernel-Appen}
\end{equation}
where $\boldsymbol{\varTheta}=\mathrm{diag}\left(\boldsymbol{\theta}\right)=\mathrm{diag}\left(\theta_{1},\cdots,\theta_{d}\right)$
with $\theta_{i}>0$ for $i=1,\cdots,d$. The standard Matérn kernel
is written as $k\left(\boldsymbol{x},\boldsymbol{x}^{\prime}\right)=\Phi_{m}\left(\boldsymbol{x}-\boldsymbol{x}^{\prime}\right)$,
with Fourier transform 
\begin{equation}
\mathscr{F}\left[\Phi_{m}\right]\left(\boldsymbol{\omega}\right)=C_{m,d}\left(1+\left\Vert \boldsymbol{\omega}\right\Vert _{2}^{2}\right)^{-m},\quad C_{m,d}>0,\label{eq:standard-Matern-spectrum}
\end{equation}
where $m>s+d/2$ \citep{wendlandScatteredDataApproximation2004}.
Under this parameterization, the conventional Matérn smoothness is
$\nu=m-d/2$.

The following proposition covers polynomial-coefficient operators
and reduces the positive definiteness of the PDE-residual Gram matrix
to a non-vanishing condition on the PDE residuals. 
\begin{prop}
\label{Prop:PDE-Gram-Gauss+Matern-polycoeff}(Gaussian and Matérn
Kernels with Polynomial-Coefficient Operators) Let $\Omega\subset\mathbb{R}^{d}$
be a bounded connected domain with nonempty interior and let $\boldsymbol{Z}=\left\{ \boldsymbol{z}_{1},\cdots,\boldsymbol{z}_{p}\right\} \subset\Omega$
be pairwise distinct. Let
\begin{equation}
\mathscr{D}=\sum_{\left|\boldsymbol{\alpha}\right|\leqslant s}a_{\boldsymbol{\alpha}}\left(\boldsymbol{x}\right)\partial^{\boldsymbol{\alpha}}\label{eq:PDE-Gram-Polynomial-Operator}
\end{equation}
be a polynomial-coefficient operator of order at most $s$. Assume
that $k\left(\cdot,\cdot\right)$ is either the Gaussian kernel in
Eq. \eqref{eq:Gauss-Kernel-Appen} or the Matérn kernel satisfying
Eq. \eqref{eq:standard-Matern-spectrum} with a real number $m>s+d/2$.
Then,
\[
\mathscr{D}k\left(\cdot,\boldsymbol{z}_{j}\right)\in L^{2}\left(\Omega\right),\quad j=1,\cdots,p.
\]
Moreover, the PDE-residual Gram matrix $\boldsymbol{A}$ is positive
definite if and only if
\begin{equation}
\mathscr{D}k\left(\cdot,\boldsymbol{z}_{j}\right)\ne0\quad\text{for every }j=1,\cdots,p.\label{eq:PDE-Gram-Nonvanishing-Condition}
\end{equation}
\end{prop}
\begin{rem}
Proposition \ref{Prop:PDE-Gram-Gauss+Matern-polycoeff} should be
read as a kernel-specific criterion. For a general collection of $L^{2}\left(\Omega\right)$
functions, non-vanishing of each element does not by itself imply
positive definiteness of the associated Gram matrix; linear independence
is also required. For the Gaussian and Matérn kernels considered here,
the required linear independence follows from the structures of the
derivative kernel sections, so the criterion reduces to the simpler
non-vanishing condition. For a general polynomial-coefficient operator,
however, this condition is not automatic and must be verified for
the operator under consideration.
\end{rem}
The constant-coefficient case is the main setting considered in this
work. Although restrictive, this setting occurs frequently in numerical
PDE studies, for example in problems involving the Laplacian, shifted
Laplacian, and constant-coefficient advection--diffusion operators;
see \citet{harkonenGaussianProcessPriors2023,dalton2024boundary,liParameterInferenceBased2024,doumechePhysicsinformedKernelLearning2025,daltonFiniteelementGaussianProcesses2026}.
In this case, the non-vanishing condition in Proposition \ref{Prop:PDE-Gram-Gauss+Matern-polycoeff}
can be verified directly for every nonzero constant-coefficient operator,
leading to the following corollary.
\begin{cor}
\label{Cor:PDE-Gram-Gauss+Matern-conscoeff}(Gaussian and Matérn Kernels
with Constant-Coefficient Operators) Under the assumptions of Proposition
\ref{Prop:PDE-Gram-Gauss+Matern-polycoeff}, suppose that $\mathscr{D}$
is a nonzero constant-coefficient differential operator. Then $\boldsymbol{A}\succ0$.
\end{cor}
Here, $\boldsymbol{A}\succ0$ denotes that $\boldsymbol{A}$ is positive
definite. Complete proofs of Proposition \ref{Prop:PDE-Gram-Gauss+Matern-polycoeff}
and Corollary \ref{Cor:PDE-Gram-Gauss+Matern-conscoeff} are given
in Section \ref{subsec:Proofs-PDE-Gram}.

\subsection{Computation of the PDE-Residual Gram Matrix}

For the Gaussian kernel and a constant-coefficient differential operator,
the PDE-residual Gram matrix admits closed-form expressions in two
representative domain settings. These expressions reduce the PDE-residual
inner product defining $\boldsymbol{A}$ to Gaussian--polynomial
moments and provide reference values for numerical evaluation.
\begin{prop}
\label{Prop:PDE-Gram-Formula-Gauss}(Gaussian PDE-Residual Gram Formulas)
Let $\Omega$ be either the whole space $\mathbb{R}^{d}$ or a hyper-rectangular
domain $\prod_{r=1}^{d}\left[a_{r},b_{r}\right]$, and let $\mathscr{D}=\sum_{\left|\boldsymbol{\alpha}\right|\leqslant s}a_{\boldsymbol{\alpha}}\partial^{\boldsymbol{\alpha}}$
be a constant-coefficient linear differential operator. For the Gaussian
kernel defined in Eq. \eqref{eq:Gauss-Kernel-Appen}, the entries
of the PDE-residual Gram matrix are given by

\begin{equation}
A_{ij}=\begin{cases}
C_{\boldsymbol{\theta}}\Pi_{\mathscr{D}}\left(\boldsymbol{z}_{i}-\boldsymbol{z}_{j}\right)\exp\left(-\frac{1}{2}\left\Vert \boldsymbol{\varTheta}\left(\boldsymbol{z}_{i}-\boldsymbol{z}_{j}\right)\right\Vert _{2}^{2}\right), & \Omega=\mathbb{R}^{d},\\
\sum_{\left|\boldsymbol{\alpha}\right|\leqslant s}\sum_{\left|\boldsymbol{\beta}\right|\leqslant s}a_{\boldsymbol{\alpha}}a_{\boldsymbol{\beta}}\left(-1\right)^{\left|\boldsymbol{\alpha}\right|+\left|\boldsymbol{\beta}\right|}\boldsymbol{\theta}^{\boldsymbol{\alpha}+\boldsymbol{\beta}}\prod_{r=1}^{d}J_{\alpha_{r},\beta_{r}}^{\left(r\right)}\left(z_{ir},z_{jr}\right), & \Omega=\prod_{r=1}^{d}\left[a_{r},b_{r}\right].
\end{cases}\label{eq:PDE-Gram-Formula-Gauss}
\end{equation}
Here $C_{\boldsymbol{\theta}}=\left(\pi/2\right)^{d/2}\prod_{i=1}^{d}\theta_{i}^{-1}$
depends only on the Gaussian scale parameters $\boldsymbol{\theta}$.
The factor $\Pi_{\mathscr{D}}\left(\cdot\right)$ is a multivariate
polynomial of degree at most $2s$, with coefficients determined by
$\mathscr{D}$ and $\boldsymbol{\theta}$, and has a finite representation
in terms of multivariate Hermite polynomials. For the hyper-rectangular
case, $J_{\alpha_{r},\beta_{r}}^{\left(r\right)}\left(z_{ir},z_{jr}\right)$
is a one-dimensional truncated Gaussian--polynomial moment over $\left[a_{r},b_{r}\right]$;
its coefficients are obtained by expanding the corresponding product
of one-dimensional Hermite polynomials.
\end{prop}
Proposition \ref{Prop:PDE-Gram-Formula-Gauss} follows from the fact
that each derivative of a Gaussian kernel is a polynomial multiplied
by a Gaussian factor. Consequently, the PDE-residual inner product
defining $\boldsymbol{A}$ reduces to Gaussian--polynomial moments.
Complete derivations and explicit evaluation formulas of Eq. \eqref{eq:PDE-Gram-Formula-Gauss}
are provided in Appendix \ref{subsec:Proof-of-A-Formulas}. 

For hyper-rectangular domains, the second formula in Eq. \eqref{eq:PDE-Gram-Formula-Gauss}
further separates the original $d$-dimensional integrals into products
of one-dimensional moments. In our computations, for the Gaussian
kernel, the required one-dimensional moments are evaluated analytically,
with the kernel derivatives computed through Hermite-polynomial recursions.
For the Matérn kernels, the derivative kernel sections are evaluated
analytically, and the resulting integrals are computed using tensor-product
Gauss--Legendre quadrature. For the periodic kernel, the required
one-dimensional derivatives are evaluated analytically, the corresponding
derivative-product integrals are computed by Gauss--Legendre quadrature,
and the multidimensional entries are assembled from these one-dimensional
quantities.

For more complicated geometries, the same definition of the PDE-residual
Gram matrix can be evaluated by numerical integration over $\Omega$,
including Monte Carlo or quasi-Monte Carlo sampling when deterministic
tensor-product quadrature is not convenient. This option is consistent
with the mesh-free character of PRIK.

\subsection{Positive Definiteness of the Kernel Matrix\label{subsec:Kernel-Matrix-PD}}

Positive definiteness of the PDE-residual Gram matrix $\boldsymbol{A}$
ensures that the PDE-residual kernel in Eq. \eqref{eq:PR-Kernel}
is well defined. However, because the PDE-residual kernel $\tilde{k}\left(\cdot,\cdot\right)$
has finite rank, $\boldsymbol{A}\succ0$ alone does not guarantee
that  $\tilde{k}\left(\boldsymbol{X}_{l},\boldsymbol{X}_{l}\right)$
is positive definite. The following proposition characterizes this
requirement and gives a sufficient condition based on the relative
locations of $\boldsymbol{X}_{l}$ and $\boldsymbol{Z}$.
\begin{prop}
\label{prop:Kernel-Matrix-PD}(Rank Characterization and Nearby Interior
Centers) Let $\boldsymbol{A}\succ0$. Under the kernel-section construction,
\[
\tilde{k}\left(\boldsymbol{X}_{l},\boldsymbol{X}_{l}\right)\succ0\iff\mathrm{rank}\left\{ k\left(\boldsymbol{X}_{l},\boldsymbol{Z}\right)\right\} =N_{l}.
\]
Suppose further that the base kernel $k\left(\cdot,\cdot\right)$
is continuous and strictly positive definite on $\bar{\Omega}$, and
that the labeled locations $\boldsymbol{X}_{l}=\left\{ \boldsymbol{x}_{1},\cdots,\boldsymbol{x}_{N_{l}}\right\} \subset\mathbb{E}$
are pairwise distinct. Then there exists $\delta>0$ such that $\tilde{k}\left(\boldsymbol{X}_{l},\boldsymbol{X}_{l}\right)\succ0$
whenever $\boldsymbol{Z}$ contains $N_{l}$ distinct interior centers
$\boldsymbol{z}_{j_{1}},\cdots,\boldsymbol{z}_{j_{N_{l}}}$ satisfying
\[
\max_{1\leqslant i\leqslant N_{l}}\left\Vert \boldsymbol{z}_{j_{i}}-\boldsymbol{x}_{i}\right\Vert _{2}<\delta.
\]
\end{prop}
\begin{proof}
Under the kernel-section construction, Eq. \eqref{eq:PR-Kernel} gives
\[
\tilde{k}\left(\boldsymbol{X}_{l},\boldsymbol{X}_{l}\right)=k\left(\boldsymbol{X}_{l},\boldsymbol{Z}\right)\boldsymbol{A}^{-1}k\left(\boldsymbol{Z},\boldsymbol{X}_{l}\right).
\]
Hence, for any $\boldsymbol{a}\in\mathbb{R}^{N_{l}}$,
\[
\boldsymbol{a}^{\top}\tilde{k}\left(\boldsymbol{X}_{l},\boldsymbol{X}_{l}\right)\boldsymbol{a}=\left\Vert \boldsymbol{A}^{-1/2}k\left(\boldsymbol{Z},\boldsymbol{X}_{l}\right)\boldsymbol{a}\right\Vert _{2}^{2}.
\]
Since $\boldsymbol{A}^{-1/2}$ is invertible, the quadratic form is
strictly positive for every nonzero $\boldsymbol{a}$ if and only
if
\[
k\left(\boldsymbol{Z},\boldsymbol{X}_{l}\right)\boldsymbol{a}\neq\boldsymbol{0}\qquad\text{for every }\boldsymbol{a}\neq\boldsymbol{0}.
\]
This is equivalent to
\[
\mathrm{rank}\left\{ k\left(\boldsymbol{X}_{l},\boldsymbol{Z}\right)\right\} =N_{l}.
\]

For the sufficient condition, define $\boldsymbol{K}_{l}=k\left(\boldsymbol{X}_{l},\boldsymbol{X}_{l}\right)$.
Since $k\left(\cdot,\cdot\right)$ is strictly positive definite and
the labeled locations are pairwise distinct, the smallest eigenvalue
of $\boldsymbol{K}_{l}$ is positive, 
\[
\lambda_{\min}\left(\boldsymbol{K}_{l}\right)>0.
\]
Consider the $N_{l}\times N_{l}$ submatrix $\boldsymbol{B}=\left[k\left(\boldsymbol{x}_{i},\boldsymbol{z}_{j_{r}}\right)\right]_{i,r=1}^{N_{l}}$
of $k\left(\boldsymbol{X}_{l},\boldsymbol{Z}\right)$. By continuity
of $k\left(\cdot,\cdot\right)$, $\boldsymbol{B}$ converges to $\boldsymbol{K}_{l}$
as
\[
\max_{1\leqslant i\leqslant N_{l}}\left\Vert \boldsymbol{z}_{j_{i}}-\boldsymbol{x}_{i}\right\Vert _{2}\rightarrow0.
\]
Therefore, there exists $\delta>0$ such that the stated proximity
condition implies
\[
\left\Vert \boldsymbol{B}-\boldsymbol{K}_{l}\right\Vert _{2}<\lambda_{\min}\left(\boldsymbol{K}_{l}\right).
\]
It follows that
\[
\sigma_{\min}\left(\boldsymbol{B}\right)\geqslant\lambda_{\min}\left(\boldsymbol{K}_{l}\right)-\left\Vert \boldsymbol{B}-\boldsymbol{K}_{l}\right\Vert _{2}>0.
\]
Thus, $\boldsymbol{B}$ is invertible, so $k\left(\boldsymbol{X}_{l},\boldsymbol{Z}\right)$
has full row rank. The first part of the proposition then gives
\[
\tilde{k}\left(\boldsymbol{X}_{l},\boldsymbol{X}_{l}\right)\succ0,
\]
which finishes the proof.
\end{proof}
The preceding result gives a deterministic design condition. For the
Gaussian and Matérn kernels used in this work, the same rank condition
also holds almost surely under continuous random interior designs.
\begin{prop}
Let $p\geqslant N_{l}$, and let the labeled locations $\boldsymbol{X}_{l}\subset\mathbb{E}$
be fixed and pairwise distinct. Suppose that $k\left(\cdot,\cdot\right)$
is either the Gaussian kernel in Eq. \eqref{eq:Gauss-Kernel-Appen}
or the Matérn kernel in Eq. \eqref{eq:standard-Matern-spectrum}.
Suppose further that the joint distribution of the interior centers
$\left(\boldsymbol{z}_{1},\cdots,\boldsymbol{z}_{p}\right)\in\Omega^{p}$
is absolutely continuous with respect to $dp$-dimensional Lebesgue
measure. Then 
\[
\mathbb{P}\left\{ \mathrm{rank}\left\{ k\left(\boldsymbol{X}_{l},\boldsymbol{Z}\right)\right\} =N_{l}\right\} =1.
\]
Consequently, if the corresponding PDE-residual Gram matrix satisfies
$\boldsymbol{A}\succ0$ almost surely, then
\[
\mathbb{P}\left\{ \tilde{k}\left(\boldsymbol{X}_{l},\boldsymbol{X}_{l}\right)\succ0\right\} =1.
\]
\end{prop}
\begin{proof}
Consider the first $N_{l}$ centers in $\boldsymbol{Z}$ and define
\[
F\left(\boldsymbol{z}_{1},\cdots,\boldsymbol{z}_{N_{l}}\right)=\det\left[k\left(\boldsymbol{x}_{i},\boldsymbol{z}_{j}\right)\right]_{i,j=1}^{N_{l}}.
\]
For the Gaussian kernel, $F$ is real analytic on $\Omega^{N_{l}}$.
For the Matérn kernel, every labeled location lies on $\mathbb{E}$
whereas every center lies in $\Omega$, so
\[
\left\Vert \boldsymbol{x}_{i}-\boldsymbol{z}_{j}\right\Vert _{2}>0,\quad\forall i,j\in\left\{ 1,\cdots,N_{l}\right\} .
\]
The Matérn radial function is real analytic at every positive distance;
hence $F$ is again real analytic on the connected open set $\Omega^{N_{l}}$. 

Choose distinct interior points
\[
\boldsymbol{z}_{j}^{\left(q\right)}\rightarrow\boldsymbol{x}_{j},\quad q\to\infty,j=1,\ldots,N_{l}.
\]
By continuity of the kernel,
\[
F\left(\boldsymbol{z}_{1}^{\left(q\right)},\cdots,\boldsymbol{z}_{N_{l}}^{\left(q\right)}\right)\rightarrow\det k\left(\boldsymbol{X}_{l},\boldsymbol{X}_{l}\right)>0.
\]
Thus, $F$ is not identically zero. Its zero set therefore has Lebesgue
measure zero. Since the joint distribution of the first $N_{l}$ centers
is absolutely continuous, $F\neq0$ almost surely. Hence the corresponding
$N_{l}\times N_{l}$ submatrix of $k\left(\boldsymbol{X}_{l},\boldsymbol{Z}\right)$
is invertible almost surely, which proves the full-row-rank result.
The conclusion for the positive definiteness of $\tilde{k}\left(\boldsymbol{X}_{l},\boldsymbol{X}_{l}\right)$
follows from Proposition \ref{prop:Kernel-Matrix-PD}.
\end{proof}

\section{Numerical Illustration of the Convergence Results\label{sec:Numerical-Illustration-of-Theory}}
\begin{example}
(Screened Poisson Equation) Let $\Omega=\left[0,\pi\right]^{2}$ and
consider the homogeneous screened Poisson equation
\[
\mathscr{D}\left(f\right)=-\partial_{x_{1},x_{1}}^{2}f-\partial_{x_{2},x_{2}}^{2}f+f=0,\quad\left(x_{1},x_{2}\right)\in\Omega.
\]
The labeled data are obtained by observing the exact solution
\[
f\left(\boldsymbol{x}\right)=\sinh\left(\sqrt{2}x_{1}\right)\sin\left(x_{2}\right),
\]
on the boundary $\mathbb{E}$.
\end{example}
We now use the above screened Poisson equation to examine the convergence
results in Sections \ref{subsec:Results-Mean} and \ref{subsec:Results-Var}.
We focus on three quantities: the $H^{2}\left(\Omega\right)$ error
of the posterior mean, the PDE-residual norm $\left\Vert \mathscr{D}\hat{f}\right\Vert _{L^{2}\left(\Omega\right)}$,
and the convergence of the normalized posterior kernel $c_{h}\left(\cdot,\cdot\right)$
to the variational Green-type kernel $G\left(\cdot,\cdot\right)$.
The function space contains kernel sections centered at the interior
points,
\[
\mathbb{V}_{p}=\mathrm{span}\left\{ k\left(\cdot,\boldsymbol{z}\right):\boldsymbol{z}\in\boldsymbol{Z}\right\} ,
\]
while the labeled points are used only as interpolation constraints.
For each $N_{\mathrm{int}}\in\left\{ 49,64,81,100,121,144,169,196\right\} $,
we generate 10 maximin designs with all interior points strictly inside
$\Omega$. The labeled points are placed uniformly along the four
edges, with $N_{\mathrm{side}}=\sqrt{N_{\mathrm{int}}}+1$ points
on each edge. The same point sets are used for all kernels and values
of $\theta$. In this experiment, we consider the Matérn kernels

\begin{equation}
k\left(\boldsymbol{x},\boldsymbol{x}^{\prime}\right)=\frac{2^{1-\nu}}{\Gamma\left(\nu\right)}\left(\sqrt{2\nu}\theta\left\Vert \boldsymbol{x}-\boldsymbol{x}^{\prime}\right\Vert \right)^{\nu}K_{\nu}\left(\sqrt{2\nu}\theta\left\Vert \boldsymbol{x}-\boldsymbol{x}^{\prime}\right\Vert \right),\label{eq:Matern}
\end{equation}
with smoothness parameters $\nu=9/2$ and $\nu=11/2$. Here, $\Gamma\left(\cdot\right)$
is the gamma function and $K_{\nu}\left(\cdot\right)$ is the modified
Bessel function of the second kind. Since $d=2$, these correspond
to the Sobolev smoothness indices $m=\nu+d/2=11/2$ and $m=\nu+d/2=13/2$,
respectively. Therefore, Theorem \ref{thm:Posterior-Mean-Convergence-Rate}
and Corollary \ref{cor:PDE-Residual-Convergence-Rate} suggest the
reference powers $m-2=7/2$ and $m-2=9/2$ for the $H^{2}\left(\Omega\right)$
error and the PDE-residual norm. These powers are used as reference
slopes in the log--log plots. The fill distance $h$ is approximated
on a fine tensor grid over $\Omega$, and the $H^{2}\left(\Omega\right)$
error and the PDE-residual norm are evaluated by tensor-product Gauss--Legendre
quadrature \citep{davis2007methods}.

\begin{figure}
\begin{centering}
\subfloat[Matérn 9/2]{\centering{}\includegraphics[width=7.9cm]{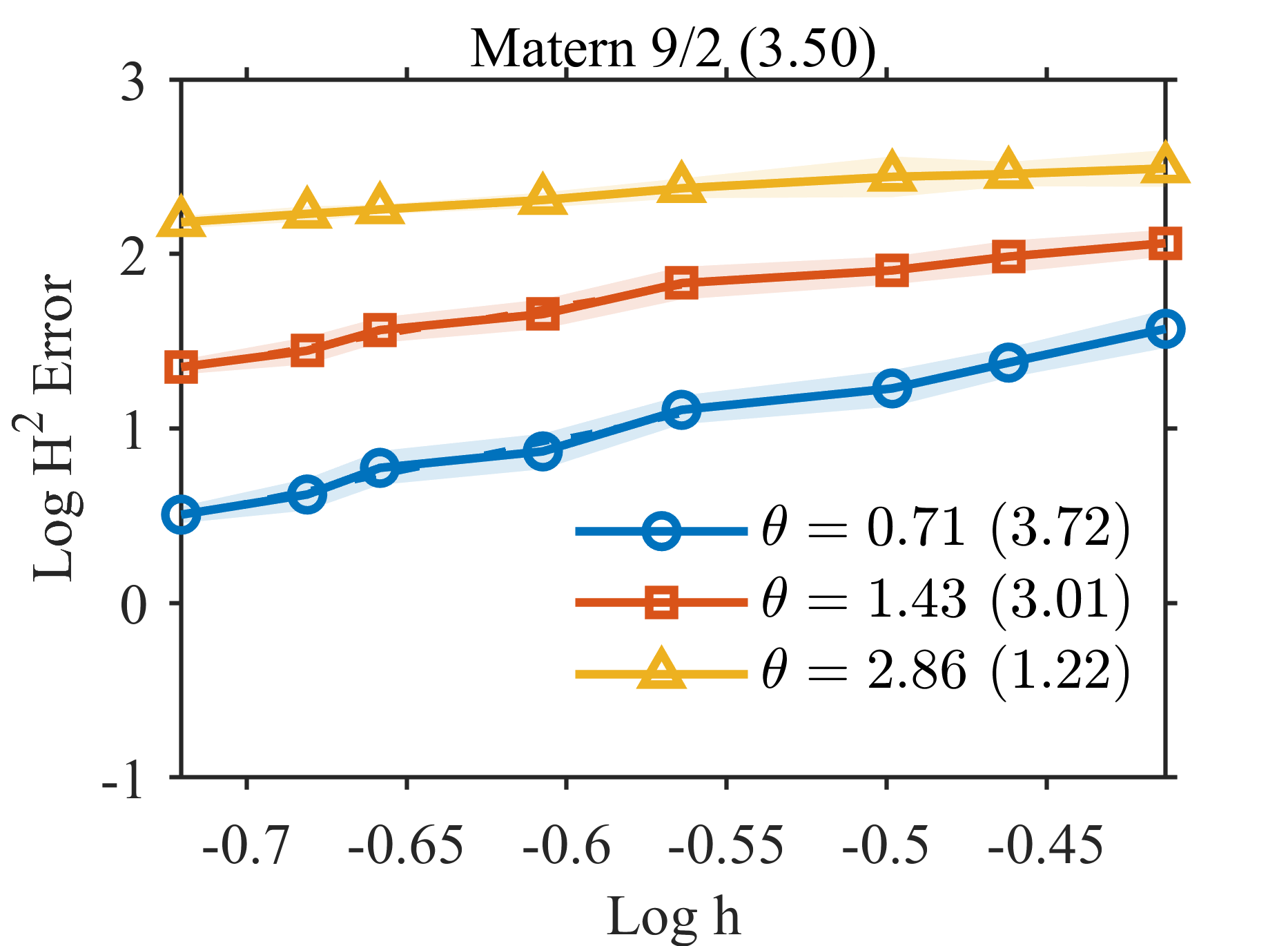}}\subfloat[Matérn 11/2]{\begin{centering}
\includegraphics[width=7.9cm]{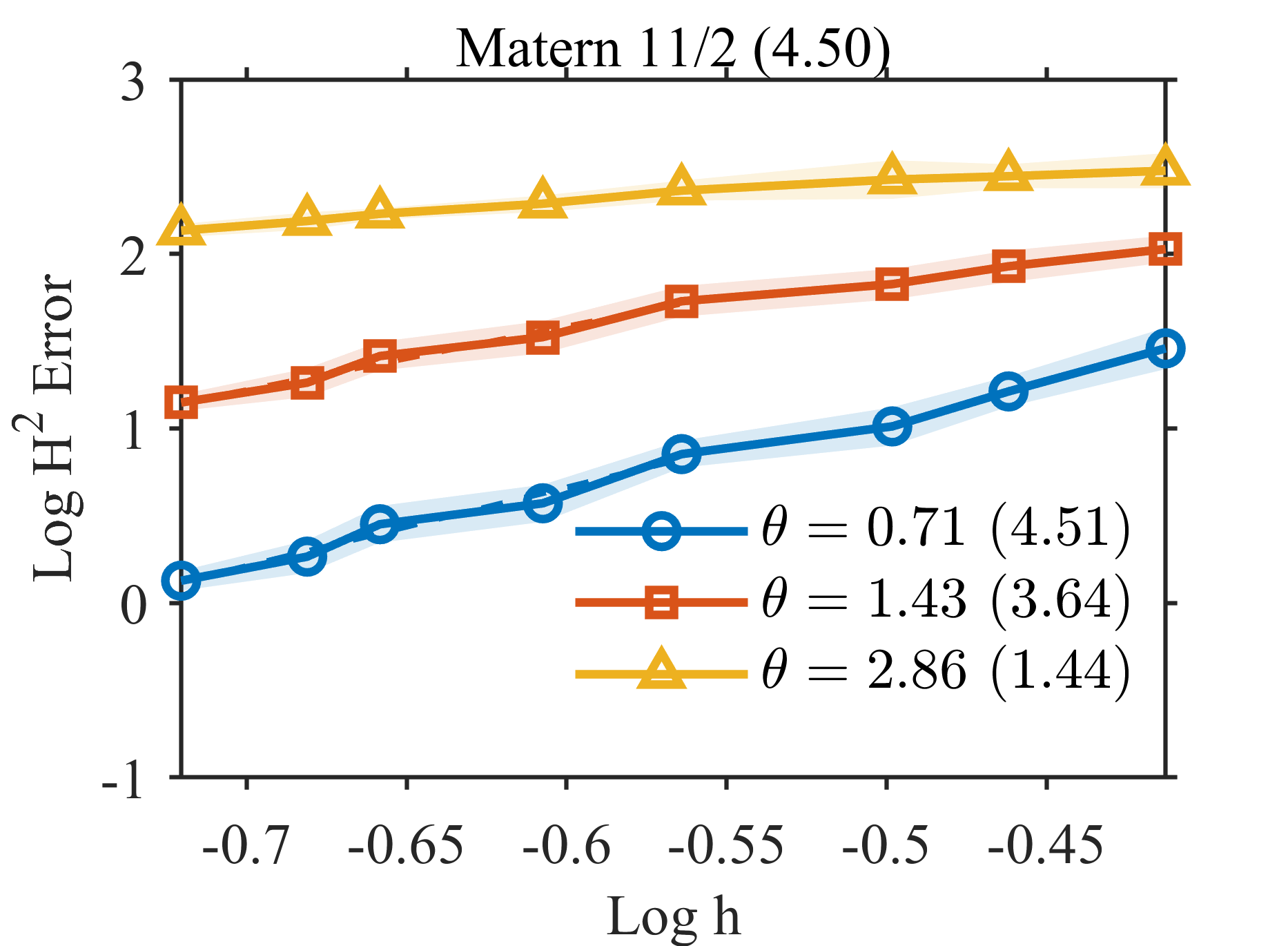}
\par\end{centering}
}
\par\end{centering}
\caption{$H^{2}\left(\Omega\right)$ error of the PRIK posterior mean for the
screened Poisson equation. Points show the mean over 10 maximin interior
designs, and shaded bands show one standard deviation on the $\log_{10}$
scale. The empirical slopes, shown in parentheses in the legends,
are fitted over $N_{\mathrm{int}}\geqslant100$. The numbers in the
panel titles indicate the reference powers $m-2$.\label{fig:H2-error-convergence}}
\end{figure}

\begin{figure}
\begin{centering}
\subfloat[Matérn 9/2]{\centering{}\includegraphics[width=7.9cm]{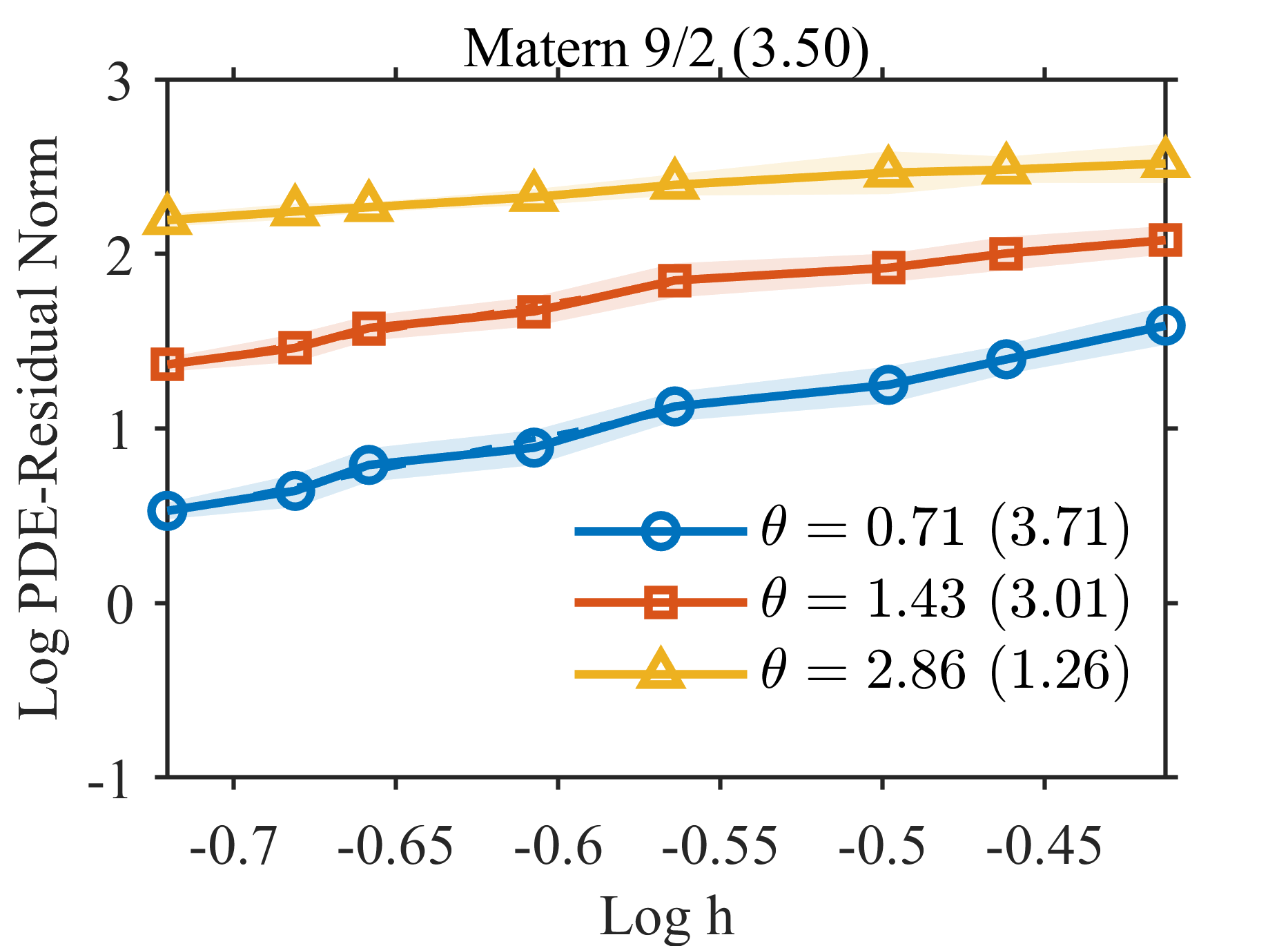}}\subfloat[Matérn 11/2]{\centering{}\includegraphics[width=7.9cm]{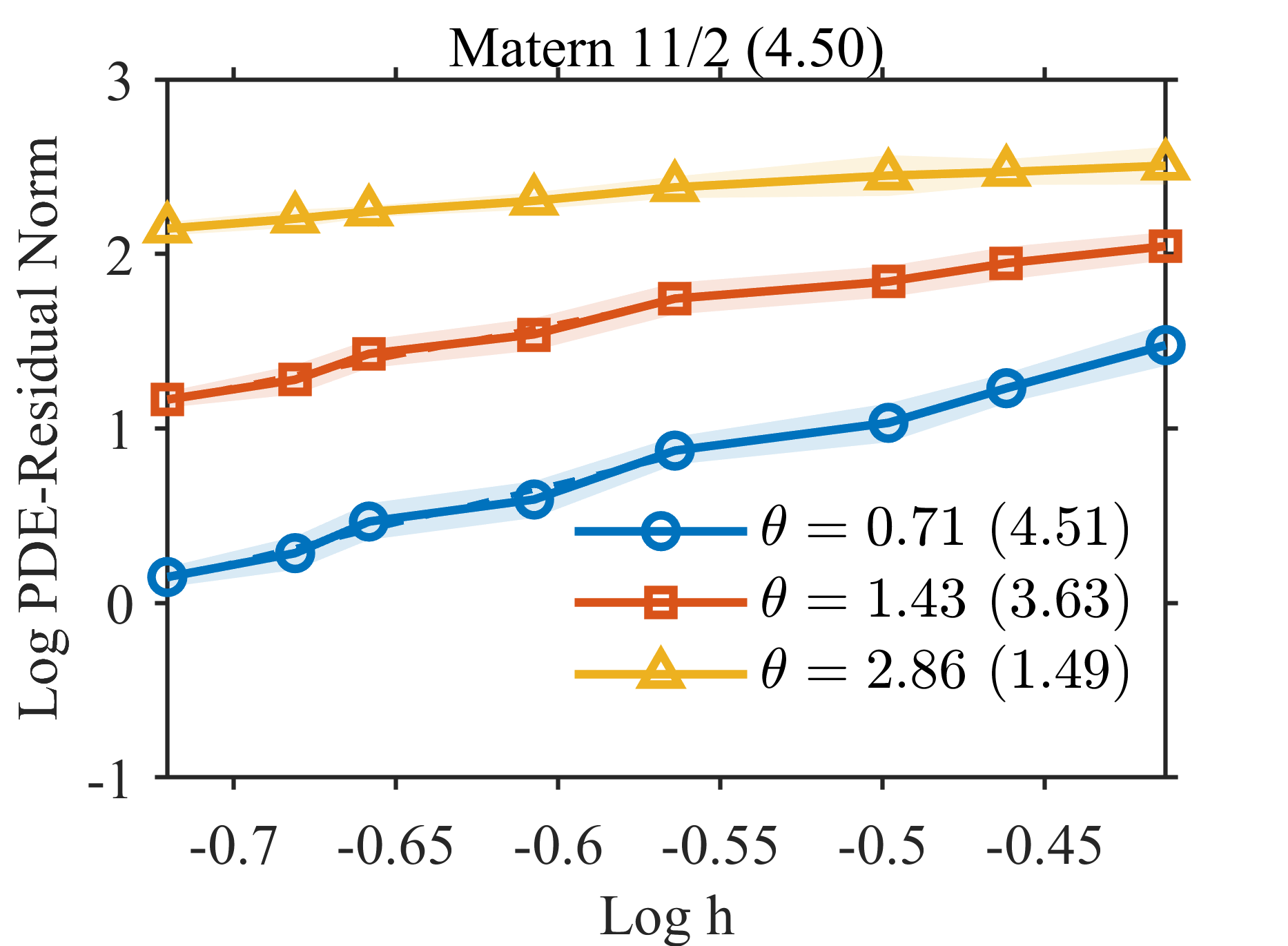}}
\par\end{centering}
\caption{PDE-residual norm $\left\Vert \mathscr{D}\hat{f}\right\Vert _{L^{2}\left(\Omega\right)}$
of PRIK for the screened Poisson equation. Points show the mean over
10 maximin interior designs, and shaded bands show one standard deviation
on the $\log_{10}$ scale. The empirical slopes are fitted over $N_{\mathrm{int}}\geqslant100$.
The numbers in the panel titles indicate the reference powers $m-2$.\label{fig:PDE-residual-norm-convergence}}
\end{figure}

Figures \ref{fig:H2-error-convergence} and \ref{fig:PDE-residual-norm-convergence}
report the $H^{2}\left(\Omega\right)$ error and the PDE-residual
norm against the approximate fill distance $h$. The empirical slopes
are fitted over the range $N_{\mathrm{int}}\geqslant100$ and are
shown in parentheses in the legends. The two quantities exhibit closely
aligned decay patterns. This agreement is consistent with the elliptic
stability estimate for the screened Poisson equation \citep{gilbarg1998elliptic,evansPartialDifferentialEquations2010}
and with the fact that Theorem \ref{thm:Posterior-Mean-Convergence-Rate}
and Corollary \ref{cor:PDE-Residual-Convergence-Rate} both suggest
the same upper-order behavior $h^{m-2}$ for the $H^{2}\left(\Omega\right)$
error and the PDE-residual norm. 

The fitted slopes need not coincide exactly with the reference powers,
since the theoretical rates are upper-bound rates under their stated
assumptions. Finite-design slopes also depend on the design points,
the accuracy of the Gauss--Legendre quadrature, and the appropriate
choice of the kernel parameter $\theta$. The dependence on $\theta$
is particularly visible in these plots. For both Matérn kernels, the
fitted slopes are close to the reference values when $\theta=0.71$.
The slopes are weaker when $\theta=1.43$, and the errors show little
further improvement when $\theta=2.86$. This sensitivity is also
reasonable for the present solution. The exact solution is smooth
and contains the factor $\sinh\left(\sqrt{2}x_{1}\right)$, which
induces a monotone exponential-type growth in the $x_{1}$-direction.
A stationary isotropic Matérn kernel is not specifically adapted to
this structure, and its performance can deteriorate when the length-scale
is too small or too large. Hence, the Matérn kernels here should be
viewed mainly as theory-aligned kernels for illustrating the Sobolev-type
convergence rates. 

We next examine the convergence of the normalized posterior kernel
described in Theorem \ref{thm:posterior-kernel-convergence}. In the
present example, $\mathscr{D}=-\Delta+1$ is self-adjoint under homogeneous
Dirichlet boundary conditions. Hence, the variational Green-type kernel
associated with $\mathscr{D}^{\ast}\mathscr{D}=\left(-\Delta+1\right)^{2}$
is given by 
\[
G\left(\boldsymbol{x},\boldsymbol{x}^{\prime}\right)=\frac{4}{\pi^{2}}\sum_{i=1}^{\infty}\sum_{j=1}^{\infty}\frac{\sin\left(ix_{1}\right)\sin\left(jx_{2}\right)\sin\left(ix_{1}^{\prime}\right)\sin\left(jx_{2}^{\prime}\right)}{\left(i^{2}+j^{2}+1\right)^{2}},
\]
where $\mathscr{D}^{\ast}$ denotes the $L^{2}\left(\Omega\right)$-adjoint
of $\mathscr{D}$. In the numerical implementation, the series is
truncated at 120 modes in each coordinate and evaluated at the reference
source point $\boldsymbol{x}^{\prime}=\left[\pi/3,\pi/2\right].$ 

\begin{figure}
\begin{centering}
\includegraphics[width=16cm]{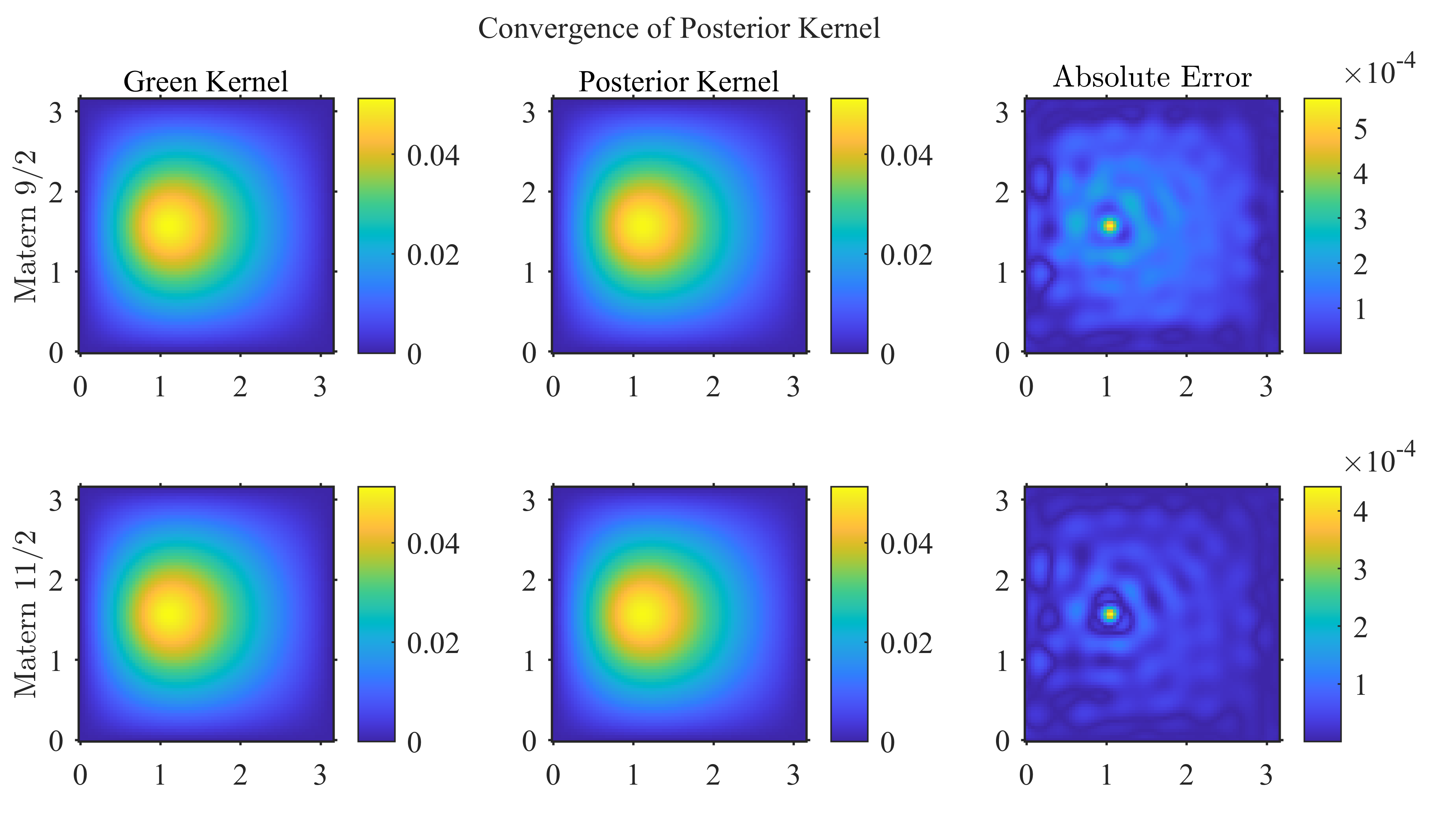}
\par\end{centering}
\caption{Approximation by the normalized posterior kernel at $\boldsymbol{x}^{\prime}=\left[\pi/3,\pi/2\right]$
for the screened Poisson equation. The first column shows the variational
Green-type kernel $G\left(\cdot,\boldsymbol{x}^{\prime}\right)$,
the second column shows the normalized PRIK posterior kernel $c_{h}\left(\cdot,\boldsymbol{x}^{\prime}\right)$,
and the third column shows the absolute error. The rows correspond
to the Matérn 9/2 and Matérn 11/2 kernels at the design with $N_{\mathrm{int}}=196$.\label{fig:posterior-kernel-heatmaps}}
\end{figure}

Figure \ref{fig:posterior-kernel-heatmaps} compares the variational
Green-type kernel $G\left(\cdot,\boldsymbol{x}^{\prime}\right)$ with
the PRIK normalized posterior kernel $c_{h}\left(\cdot,\boldsymbol{x}^{\prime}\right)$
for $\theta=1.43$ and the design with $N_{\mathrm{int}}=196$. The
two kernels are visually almost indistinguishable, and the absolute
error is of order $10^{-4}$. Both Matérn kernels capture the same
global structure of the variational Green-type kernel, with the Matérn
11/2 kernel producing a slightly smoother approximation. This comparison
is consistent with the limiting behavior described in Theorem \ref{thm:posterior-kernel-convergence},
namely that the PRIK normalized posterior kernel provides a good approximation
to the variational Green-type kernel. 

The preceding results also illustrate the practical role of kernel
choice and kernel-parameter selection. The Matérn kernels are used
in this section primarily because their Sobolev-type native spaces
match the assumptions of Theorem \ref{thm:Posterior-Mean-Convergence-Rate}
and Corollary \ref{cor:PDE-Residual-Convergence-Rate}, making them
suitable for illustrating the predicted $h^{m-2}$ behavior. This
theoretical alignment does not imply that Matérn kernels are always
the most effective empirical choice for PDE solving. In addition,
as shown by the sensitivity to $\theta$, the approximation quality
can depend noticeably on the kernel scale. Similar sensitivity has
been widely discussed in kernel methods, where the shape or scale
parameter affects both approximation accuracy and numerical stability
\citep{fasshauer2007choosing,fasshauer2012stable}. Accordingly, in
the PDE-solving examples in Section \ref{sec:Numerical-PDE-Solving},
smoother kernels and problem-adapted kernels can be preferable when
the target solution is very smooth or has a periodic structure. The
corresponding kernel parameters are then selected by the likelihood-based
procedure in Section \ref{subsec:Parameter-Estimation}, rather than
being fixed solely for theoretical illustration.

\section{Stable Discrete Boundary Correction Justification\label{sec:Stable-Discrete-Boundary-Correction-Justification}}

This section discusses the justification for Assumption \ref{assu:Stable-Discrete-Boundary-Correction}
and clarifies how it can be assessed for the kernel-section function
space $\mathcal{H}_{h}$ considered in Section \ref{sec:Theoretical-Properties}.
We first derive a basis-invariant characterization and a sufficient
condition within the kernel-section function space. We then give an
explicit interior design for Matérn kernels on a square domain under
which the assumption holds uniformly.

\subsection{Characterization and a Sufficient Condition}

This subsection translates Assumption \ref{assu:Stable-Discrete-Boundary-Correction}
into finite-dimensional conditions on the kernel-section function
space $\mathcal{H}_{h}$. We first derive an exact characterization
that is independent of the choice of basis. We then use the Lagrange
basis associated with the interior centers to obtain a sufficient
condition in terms of boundary cross-evaluations.

We begin with the following lemma, which gives an exact characterization
of Assumption \ref{assu:Stable-Discrete-Boundary-Correction}.
\begin{lem}
\label{lem:Boundary-Correction-Matrix-Characterization} (Basis-Invariant
Matrix Characterization) Let $\left\{ \psi_{j,h}\right\} {}_{j=1}^{p_{h}}$
be any basis of $\mathcal{H}_{h}$, and define the boundary basis-evaluation
matrix
\[
\boldsymbol{\varPsi}_{l,h}=\left[\psi_{1,h}\left(\boldsymbol{X}_{l,h}\right),\cdots,\psi_{p_{h},h}\left(\boldsymbol{X}_{l,h}\right)\right]=\left[\psi_{j,h}\left(\boldsymbol{x}_{i,h}\right)\right]_{i\in\left\{ 1,\cdots,N_{l,h}\right\} ,j\in\left\{ 1,\cdots,p_{h}\right\} },
\]
where $\boldsymbol{X}_{l,h}=\left\{ \boldsymbol{x}_{1,h},\cdots,\boldsymbol{x}_{N_{l,h},h}\right\} $.
Define also the $H^{2}$-Gram matrix $\boldsymbol{M}_{2,h}=\left[\left<\psi_{i,h},\psi_{j,h}\right>_{H^{2}\left(\Omega\right)}\right]_{i,j=1}^{p_{h}}$
and the generalized boundary Gram matrix $\boldsymbol{S}_{2,h}=\boldsymbol{\varPsi}_{l,h}\boldsymbol{M}_{2,h}^{-1}\boldsymbol{\varPsi}_{l,h}^{\top}$.
Then Assumption \ref{assu:Stable-Discrete-Boundary-Correction} holds
if and only if
\begin{equation}
\lambda_{\min}\left(\boldsymbol{S}_{2,h}\right)\gtrsim h^{4-d},\label{eq:Boundary-Correction-Frame-Eigenvalue}
\end{equation}
where $\lambda_{\min}\left(\boldsymbol{S}_{2,h}\right)$ denotes the
smallest eigenvalue of $\boldsymbol{S}_{2,h}$. Moreover, $\boldsymbol{S}_{2,h}$
and the eigenvalue condition in Eq. \eqref{eq:Boundary-Correction-Frame-Eigenvalue}
are invariant under any invertible change of basis in $\mathcal{H}_{h}$.
\end{lem}
\begin{proofof}{lem:Boundary-Correction-Matrix-Characterization}

Write $r_{h}=\sum_{j=1}^{p_{h}}c_{j}\psi_{j,h}$. For prescribed labels
$\boldsymbol{b}\in\mathbb{R}^{N_{l,h}}$, the minimum-norm correction
is obtained from
\[
\underset{\boldsymbol{c}\in\mathbb{R}^{p_{h}}}{\min}\boldsymbol{c}^{\top}\boldsymbol{M}_{2,h}\boldsymbol{c}\quad\text{subject to}\quad\boldsymbol{\varPsi}_{l,h}\boldsymbol{c}=\boldsymbol{b}.
\]
If $\boldsymbol{\varPsi}_{l,h}$ does not have full row rank, Assumption
\ref{assu:Stable-Discrete-Boundary-Correction} fails because not
every vector $\boldsymbol{b}$ is interpolated. We therefore consider
the case where $\boldsymbol{\varPsi}_{l,h}$ has full row rank. The
Lagrange multiplier equations give $\boldsymbol{c}_{\boldsymbol{b}}=\boldsymbol{M}_{2,h}^{-1}\boldsymbol{\varPsi}_{l,h}^{\top}\boldsymbol{S}_{2,h}^{-1}\boldsymbol{b}$,
and hence
\[
\inf_{\substack{r_{h}\in\mathcal{H}_{h}\\
r_{h}\left(\boldsymbol{X}_{l,h}\right)=\boldsymbol{b}
}
}\left\Vert r_{h}\right\Vert _{H^{2}\left(\Omega\right)}^{2}=\boldsymbol{b}^{\top}\boldsymbol{S}_{2,h}^{-1}\boldsymbol{b}.
\]
After squaring the bound in Assumption \ref{assu:Stable-Discrete-Boundary-Correction}
and using the equality above, the assumption is therefore equivalent
to
\[
\left\Vert \boldsymbol{S}_{2,h}^{-1}\right\Vert _{2}=\sup_{\boldsymbol{b}\neq\boldsymbol{0}}\frac{\left\Vert \boldsymbol{S}_{2,h}^{-1}\boldsymbol{b}\right\Vert _{2}}{\left\Vert \boldsymbol{b}\right\Vert _{2}}\lesssim h^{d-4}.
\]
This is exactly Eq. \eqref{eq:Boundary-Correction-Frame-Eigenvalue},
since $\lambda_{\min}\left(\boldsymbol{S}_{2,h}\right)\left\Vert \boldsymbol{S}_{2,h}^{-1}\right\Vert _{2}=1$.

It remains to verify that this characterization does not depend on
the particular basis used for $\mathcal{H}_{h}$. Let $\boldsymbol{R}\in\mathbb{R}^{p_{h}\times p_{h}}$
be an invertible matrix and define a new basis by $\tilde{\psi}_{j,h}=\sum_{k=1}^{p_{h}}\psi_{k,h}R_{kj}$.
Then
\[
\tilde{\boldsymbol{\varPsi}}_{l,h}=\boldsymbol{\varPsi}_{l,h}\boldsymbol{R},\quad\tilde{\boldsymbol{M}}_{2,h}=\boldsymbol{R}^{\top}\boldsymbol{M}_{2,h}\boldsymbol{R}.
\]
Direct substitution shows that
\[
\tilde{\boldsymbol{\varPsi}}_{l,h}\tilde{\boldsymbol{M}}_{2,h}^{-1}\tilde{\boldsymbol{\varPsi}}_{l,h}^{\top}=\boldsymbol{S}_{2,h},
\]
which proves the basis invariance.\end{proofof}

Lemma \ref{lem:Boundary-Correction-Matrix-Characterization} provides
a direct finite-dimensional diagnostic of Assumption \ref{assu:Stable-Discrete-Boundary-Correction}.
In particular, the normalized quantity $h^{d-4}\lambda_{\min}\left(\boldsymbol{S}_{2,h}\right)$
should remain bounded away from zero along $h\to0$. Moreover, since
this characterization is invariant under a change of basis, we are
free to utilize a convenient basis of $\mathcal{H}_{h}$ when deriving
sufficient conditions. We therefore use the Lagrange basis associated
with the interior centers in the remainder of this section.

Let $\left\{ \ell_{\boldsymbol{z},h}:\boldsymbol{z}\in\boldsymbol{X}_{\mathrm{int},h}\right\} $
denote the Lagrange basis associated with the interior centers, characterized
by
\[
\ell_{\boldsymbol{z},h}\left(\boldsymbol{\zeta}\right)=\begin{cases}
1, & \boldsymbol{z}=\boldsymbol{\zeta},\\
0, & \boldsymbol{z}\neq\boldsymbol{\zeta},
\end{cases}\quad\boldsymbol{z},\boldsymbol{\zeta}\in\boldsymbol{X}_{\mathrm{int},h}.
\]
Writing $\boldsymbol{X}_{\mathrm{int},h}=\left\{ \boldsymbol{z}_{1,h},\cdots,\boldsymbol{z}_{p_{h},h}\right\} $,
the Lagrange basis can equivalently be expressed as
\[
\left[\begin{array}{c}
\ell_{\boldsymbol{z}_{1,h},h}\left(\boldsymbol{x}\right)\\
\vdots\\
\ell_{\boldsymbol{z}_{p_{h},h},h}\left(\boldsymbol{x}\right)
\end{array}\right]=k\left(\boldsymbol{X}_{\mathrm{int},h},\boldsymbol{X}_{\mathrm{int},h}\right)^{-1}k\left(\boldsymbol{X}_{\mathrm{int},h},\boldsymbol{x}\right).
\]
These functions form an alternative basis of the same kernel-section
function space $\mathcal{H}_{h}$. Thus, the change of basis does
not alter the correction problem characterized in Lemma \ref{lem:Boundary-Correction-Matrix-Characterization}.

The next lemma shows that, in this basis, the $\ell^{2}$-norm of
the coefficient vector controls the $H^{2}$-norm of the resulting
kernel-space function with the same $h^{d/2-2}$ scaling required
by Assumption \ref{assu:Stable-Discrete-Boundary-Correction}.
\begin{lem}
\label{lem:Interior-Lagrange-H2-Coefficient-Stability} (Coefficient
Stability of the Interior Lagrange Basis) Under Assumptions \ref{assu:Smoothness-of-Domain},
\ref{assu:Sobolev-Compatible-Kernel}, and \ref{assu:Refinement-Design},
\begin{equation}
\left\Vert \sum_{\boldsymbol{z}\in\boldsymbol{X}_{\mathrm{int},h}}a_{\boldsymbol{z}}\ell_{\boldsymbol{z},h}\right\Vert _{H^{2}\left(\Omega\right)}\lesssim h^{d/2-2}\left(\sum_{\boldsymbol{z}\in\boldsymbol{X}_{\mathrm{int},h}}\left|a_{\boldsymbol{z}}\right|^{2}\right)^{1/2}.\label{eq:Interior-Lagrange-H2-Coefficient-Stability}
\end{equation}
\end{lem}
\begin{proofof}{lem:Interior-Lagrange-H2-Coefficient-Stability}

For $s_{h}=\sum_{\boldsymbol{z}}a_{\boldsymbol{z}}\ell_{\boldsymbol{z},h}$,
the function-space interpolation stability estimate of \citet[Corollary 2]{deMarchiSchabackStability2010}
gives
\[
\left\Vert s_{h}\right\Vert _{L^{2}\left(\Omega\right)}\lesssim h^{d/2}\left\Vert \boldsymbol{a}\right\Vert _{2}
\]
for sufficiently dense quasi-uniform centers and a finite-smoothness
kernel whose native space is norm equivalent to $H^{m}\left(\Omega\right)$.
Moreover, \citet[Theorem 4.4]{sunInverseInequalitiesKernel2025},
applied with Sobolev orders $\alpha=2$ and $t=0$, gives
\[
\left\Vert s_{h}\right\Vert _{H^{2}\left(\Omega\right)}\lesssim q_{\boldsymbol{X}_{\mathrm{int},h}}^{-2}\left\Vert s_{h}\right\Vert _{L^{2}\left(\Omega\right)}.
\]
Since $q_{\boldsymbol{X}_{\mathrm{int},h}}\asymp h$ under Assumption
\ref{assu:Refinement-Design}, combining the two estimates proves
Eq. \eqref{eq:Interior-Lagrange-H2-Coefficient-Stability}.\end{proofof}

Assumption \ref{assu:Stable-Discrete-Boundary-Correction} only requires
the existence of one admissible correction with the prescribed $H^{2}$-norm
bound for each boundary vector. Therefore, it is sufficient to construct
such a correction explicitly. The next proposition does so using only
$N_{l,h}$ suitably selected functions from the interior Lagrange
basis.
\begin{prop}
\label{prop:Boundary-Proxy-Sufficient-Correction} (Boundary-Proxy
Sufficient Condition) Select $N_{l,h}$ distinct centers from $\boldsymbol{X}_{\mathrm{int},h}$,
denoted by 
\[
\boldsymbol{Z}_{\mathbb{E},h}=\left\{ \boldsymbol{z}_{1,h},\cdots,\boldsymbol{z}_{N_{l,h},h}\right\} \subseteq\boldsymbol{X}_{\mathrm{int},h},
\]
and refer to them as the interior proxy centers. Define the cross-evaluation
matrix
\begin{equation}
\boldsymbol{B}_{\mathbb{E},h}=\left[\ell_{\boldsymbol{z}_{j,h},h}\left(\boldsymbol{x}_{i,h}\right)\right]_{i,j=1}^{N_{l,h}}.\label{eq:Boundary-Proxy-Cross-Evaluation}
\end{equation}
If there is a constant $c_{\mathbb{E}}>0$, independent of $h$, such
that
\begin{equation}
\sigma_{\min}\left(\boldsymbol{B}_{\mathbb{E},h}\right)\geqslant c_{\mathbb{E}},\label{eq:Boundary-Proxy-Singular-Lower-Bound}
\end{equation}
where $\sigma_{\min}\left(\boldsymbol{B}_{\mathbb{E},h}\right)$ denotes
the smallest singular value of $\boldsymbol{B}_{\mathbb{E},h}$, then
Assumption \ref{assu:Stable-Discrete-Boundary-Correction} holds.
\end{prop}
\begin{proofof}{prop:Boundary-Proxy-Sufficient-Correction}

For prescribed labels $\boldsymbol{b}$, set
\[
\boldsymbol{a}=\boldsymbol{B}_{\mathbb{E},h}^{-1}\boldsymbol{b},\quad r_{h}=\sum_{j=1}^{N_{l,h}}a_{j}\ell_{\boldsymbol{z}_{j,h},h}.
\]
Then $r_{h}\left(\boldsymbol{X}_{l,h}\right)=\boldsymbol{b}$. By
Lemma \ref{lem:Interior-Lagrange-H2-Coefficient-Stability} and Eq.
\eqref{eq:Boundary-Proxy-Singular-Lower-Bound},
\[
\left\Vert r_{h}\right\Vert _{H^{2}\left(\Omega\right)}\lesssim h^{d/2-2}\left\Vert \boldsymbol{a}\right\Vert _{2}\leqslant c_{\mathbb{E}}^{-1}h^{d/2-2}\left\Vert \boldsymbol{b}\right\Vert _{2}.
\]
Since $r_{h}$ is an admissible correction, 
\[
\inf_{\substack{v_{h}\in\mathcal{H}_{h}\\
v_{h}\left(\boldsymbol{X}_{l,h}\right)=\boldsymbol{b}
}
}\left\Vert v_{h}\right\Vert _{H^{2}\left(\Omega\right)}\leqslant\left\Vert r_{h}\right\Vert _{H^{2}\left(\Omega\right)}\lesssim h^{d/2-2}\left\Vert \boldsymbol{b}\right\Vert _{2},
\]
which is precisely Assumption \ref{assu:Stable-Discrete-Boundary-Correction}
for $\mathbb{V}_{p_{h}}=\mathcal{H}_{h}$.\end{proofof}

Proposition \ref{prop:Boundary-Proxy-Sufficient-Correction} reduces
Assumption \ref{assu:Stable-Discrete-Boundary-Correction} to two
stability requirements. The first, established in Lemma \ref{lem:Interior-Lagrange-H2-Coefficient-Stability},
controls how the coefficients of a Lagrange expansion in the kernel-section
function space $\mathcal{H}_{h}$ affect its $H^{2}$-norm. The second
requires that the selected interior Lagrange functions can stably
reproduce boundary values, which is exactly the role of the lower
bound on $\sigma_{\min}\left(\boldsymbol{B}_{\mathbb{E},h}\right)$.
The next subsection gives an explicit design for which this condition
can be verified.

\subsection{An Explicit Sufficient Design for Matérn Kernels}

Proposition \ref{prop:Boundary-Proxy-Sufficient-Correction} shows
that Assumption \ref{assu:Stable-Discrete-Boundary-Correction} holds
if the boundary cross-evaluation matrix remains uniformly nonsingular.
We now give an explicit interior design for which this property can
be verified, within the Matérn smoothness setting considered in Section
\ref{sec:Numerical-Illustration-of-Theory}. The main idea is to add
one interior center close to each labeled point. Since the Lagrange
basis is interpolatory at the interior centers, its cross-evaluation
matrix approaches the identity when these paired interior centers
approach the corresponding labeled points. The following lemma controls
this change.
\begin{lem}
\label{lem:Discrete-Normal-Trace-Inverse-Square} (Discrete Gradient
Bound Near the Boundary) Let $\Omega=\left(0,1\right)^{2}$, let $m>3$,
and let $k\left(\cdot,\cdot\right)=\Phi_{m}\left(\cdot-\cdot\right)$
satisfy Assumption \ref{assu:Sobolev-Compatible-Kernel}. For $h=1/n$,
$n\in\mathbb{N}$, define the midpoint boundary grid
\[
\boldsymbol{X}_{l,h}=\bigcup_{j=0}^{n-1}\left\{ \left(\left(j+\frac{1}{2}\right)h,0\right),\left(\left(j+\frac{1}{2}\right)h,1\right),\left(0,\left(j+\frac{1}{2}\right)h\right),\left(1,\left(j+\frac{1}{2}\right)h\right)\right\} .
\]
Let $\boldsymbol{X}_{\mathrm{int},h}\subset\Omega$ satisfy the interior-design
condition in Assumption \ref{assu:Refinement-Design}. For $0\leqslant r\leqslant h/4$,
let $\boldsymbol{X}_{l,h}\left(r\right)$ be obtained by moving each
labeled point in $\boldsymbol{X}_{l,h}$ a distance $r$ in the inward
normal direction. Then there exists a constant $C_{{\rm nt}}>0$,
independent of $h$, $r$, and $v_{h}$, such that
\begin{equation}
\left(\sum_{\boldsymbol{x}\in\boldsymbol{X}_{l,h}(r)}\left\Vert \nabla v_{h}\left(\boldsymbol{x}\right)\right\Vert _{2}^{2}\right)^{1/2}\leqslant C_{{\rm nt}}h^{-1}\left\Vert v_{h}\left(\boldsymbol{X}_{\mathrm{int},h}\right)\right\Vert _{2},\quad v_{h}\in\mathcal{H}_{h}.\label{eq:Discrete-Normal-Trace-Inverse-Square}
\end{equation}
\end{lem}
\begin{proofof}{lem:Discrete-Normal-Trace-Inverse-Square}

Let $\boldsymbol{c}_{h}=v_{h}\left(\boldsymbol{X}_{\mathrm{int},h}\right)$.
By the stability estimates for quasi-uniform kernel-section function
spaces, \citet[Theorem 4.6]{sunInverseInequalitiesKernel2025}, applied
with $d=2$ and Sobolev orders $1$ and $2$, gives
\[
\left\Vert v_{h}\right\Vert _{H^{1}\left(\Omega\right)}\lesssim\left\Vert \boldsymbol{c}_{h}\right\Vert _{2},\quad\left\Vert v_{h}\right\Vert _{H^{2}\left(\Omega\right)}\lesssim h^{-1}\left\Vert \boldsymbol{c}_{h}\right\Vert _{2}.
\]
Moreover, \citet[Lemma 2.2]{sunInverseInequalitiesKernel2025} gives
\[
\left\Vert v_{h}\right\Vert _{H^{m}\left(\Omega\right)}\lesssim h^{1-m}\left\Vert \boldsymbol{c}_{h}\right\Vert _{2}.
\]
Consider one of the four boundary faces and its parallel translate
at distance $r$, denoted by $F_{r}$. Applying the multiplicative
trace inequality on a fixed-width rectangular neighborhood of $F_{r}$
gives, uniformly for $0\leqslant r\leqslant h/4$, 
\[
\left\Vert \nabla v_{h}\right\Vert _{L^{2}\left(F_{r}\right)}\lesssim\left\Vert v_{h}\right\Vert _{H^{1}\left(\Omega\right)}^{1/2}\left\Vert v_{h}\right\Vert _{H^{2}\left(\Omega\right)}^{1/2}\lesssim h^{-1/2}\left\Vert \boldsymbol{c}_{h}\right\Vert _{2}.
\]
The Sobolev trace theorem also yields
\[
\left\Vert \nabla v_{h}\right\Vert _{H^{m-3/2}\left(F_{r}\right)}\lesssim\left\Vert v_{h}\right\Vert _{H^{m}\left(\Omega\right)}\lesssim h^{1-m}\left\Vert \boldsymbol{c}_{h}\right\Vert _{2}.
\]
The shifted points on $F_{r}$ are separated at scale $h$. Applying
the one-dimensional upper sampling inequality to $\nabla v_{h}|_{F_{r}}$
therefore gives
\[
\begin{aligned}\left(\sum_{\boldsymbol{x}\in\boldsymbol{X}_{l,h}(r)\cap F_{r}}\left\Vert \nabla v_{h}\left(\boldsymbol{x}\right)\right\Vert _{2}^{2}\right)^{1/2} & \lesssim h^{-1/2}\left\Vert \nabla v_{h}\right\Vert _{L^{2}\left(F_{r}\right)}+h^{m-2}\left\Vert \nabla v_{h}\right\Vert _{H^{m-3/2}\left(F_{r}\right)}\\
 & \lesssim h^{-1}\left\Vert \boldsymbol{c}_{h}\right\Vert _{2}.
\end{aligned}
\]
Summing over the four faces proves Eq. \eqref{eq:Discrete-Normal-Trace-Inverse-Square}.\end{proofof}

The lemma allows us to construct an interior design for which the
boundary cross-evaluation matrix is asymptotically close to the identity,
thereby verifying Assumption \ref{assu:Stable-Discrete-Boundary-Correction}.
\begin{cor}
\label{cor:Matern-Normal-Proxy-Stable-Boundary-Correction} Let $\Omega=\left(0,1\right){}^{2}$,
let $h=1/n$, and let $\boldsymbol{X}_{l,h}$ be the midpoint boundary
grid in Lemma \ref{lem:Discrete-Normal-Trace-Inverse-Square}. Fix
$\alpha>0$ and set $\delta_{h}=h^{1+\alpha}$. For each $\boldsymbol{x}_{i,h}\in\boldsymbol{X}_{l,h}$,
let $\boldsymbol{n}_{i,h}$ be the outward unit normal to the corresponding
boundary face and define the paired interior center $\boldsymbol{z}_{i,h}=\boldsymbol{x}_{i,h}-\delta_{h}\boldsymbol{n}_{i,h}$.
Let $\boldsymbol{Z}_{\mathbb{E},h}=\left\{ \boldsymbol{z}_{i,h}:\boldsymbol{x}_{i,h}\in\boldsymbol{X}_{l,h}\right\} $,
and define the cell-center grid
\[
\boldsymbol{G}_{h}=\left\{ \left(\left(i+\frac{1}{2}\right)h,\left(j+\frac{1}{2}\right)h\right):i,j=0,\ldots,n-1\right\} .
\]
Take $\boldsymbol{X}_{\mathrm{int},h}=\boldsymbol{G}_{h}\cup\boldsymbol{Z}_{\mathbb{E},h}$.
Let $k\left(\cdot,\cdot\right)$ be the Matérn kernel considered in
Eq. \eqref{eq:standard-Matern-spectrum}, with Sobolev smoothness
$m>3$.

Then, for all sufficiently small $h$, Assumption \ref{assu:Stable-Discrete-Boundary-Correction}
holds uniformly. In particular, there exists a constant $C>0$, independent
of $h$ and $\boldsymbol{b}$, such that
\begin{equation}
\inf_{\substack{r_{h}\in\mathcal{H}_{h}\\
r_{h}\left(\boldsymbol{X}_{l,h}\right)=\boldsymbol{b}
}
}\left\Vert r_{h}\right\Vert _{H^{2}\left(\Omega\right)}\leqslant Ch^{-1}\left\Vert \boldsymbol{b}\right\Vert _{2},\quad\boldsymbol{b}\in\mathbb{R}^{N_{l,h}}.\label{eq:Matern-Normal-Proxy-Stable-Boundary-Correction}
\end{equation}
In particular, the smoothness condition includes the Matérn 9/2 and
Matérn 11/2 kernels considered in Section \ref{sec:Numerical-Illustration-of-Theory}.
\end{cor}
\begin{proofof}{cor:Matern-Normal-Proxy-Stable-Boundary-Correction}

For all sufficiently small $h$,
\[
\delta_{h}=h^{1+\alpha}\leqslant\frac{h}{4}.
\]
The cell-center grid provides fill distance of order $h$, and, for
all sufficiently small $h$, the additional paired interior centers
remain separated from each other and from the cell centers at the
same order. Hence
\[
h_{\boldsymbol{X}_{\mathrm{int},h},\Omega}\asymp q_{\boldsymbol{X}_{\mathrm{int},h}}\asymp h.
\]
Let $\left\{ \ell_{\boldsymbol{z},h}:\boldsymbol{z}\in\boldsymbol{X}_{\mathrm{int},h}\right\} $
be the Lagrange basis of $\mathcal{H}_{h}$, and define the cross-evaluation
matrix 
\[
\boldsymbol{B}_{\mathbb{E},h}=\left[\ell_{\boldsymbol{z}_{j,h},h}\left(\boldsymbol{x}_{i,h}\right)\right]_{i,j=1}^{N_{l,h}},
\]
where $\boldsymbol{z}_{i,h}$ is paired with $\boldsymbol{x}_{i,h}$.

For $\boldsymbol{a}\in\mathbb{R}^{N_{l,h}}$, define $s_{\boldsymbol{a},h}=\sum_{j=1}^{N_{l,h}}a_{j}\ell_{\boldsymbol{z}_{j,h},h}$.
By the Lagrange property, $s_{\boldsymbol{a},h}\left(\boldsymbol{z}_{i,h}\right)=a_{i}$,
and 
\[
\left\Vert s_{\boldsymbol{a},h}\left(\boldsymbol{X}_{\mathrm{int},h}\right)\right\Vert _{2}=\left\Vert \boldsymbol{a}\right\Vert _{2}.
\]
For each paired labeled and interior point, the fundamental theorem
of calculus along the normal segment gives
\[
s_{\boldsymbol{a},h}\left(\boldsymbol{x}_{i,h}\right)-s_{\boldsymbol{a},h}\left(\boldsymbol{z}_{i,h}\right)=\int_{0}^{\delta_{h}}\nabla s_{\boldsymbol{a},h}\left(\boldsymbol{x}_{i,h}-r\boldsymbol{n}_{i,h}\right)^{\top}\boldsymbol{n}_{i,h}\mathrm{d}r.
\]
Therefore, by Minkowski's inequality and Lemma \ref{lem:Discrete-Normal-Trace-Inverse-Square}
\[
\begin{aligned}\left\Vert \left(\boldsymbol{B}_{\mathbb{E},h}-\boldsymbol{I}\right)\boldsymbol{a}\right\Vert _{2} & \leqslant\int_{0}^{\delta_{h}}\left(\sum_{\boldsymbol{x}\in\boldsymbol{X}_{l,h}(r)}\left\Vert \nabla s_{\boldsymbol{a},h}\left(\boldsymbol{x}\right)\right\Vert _{2}^{2}\right)^{1/2}\mathrm{d}r\\
 & \leqslant C_{{\rm nt}}\delta_{h}h^{-1}\left\Vert \boldsymbol{a}\right\Vert _{2}\\
 & =C_{{\rm nt}}h^{\alpha}\left\Vert \boldsymbol{a}\right\Vert _{2}.
\end{aligned}
\]
It follows that
\begin{equation}
\left\Vert \boldsymbol{B}_{\mathbb{E},h}-\boldsymbol{I}\right\Vert _{2}\leqslant C_{{\rm nt}}h^{\alpha}.\label{eq:Matern-Normal-Proxy-Cross-Matrix-Convergence}
\end{equation}
Hence, for all sufficiently small $h$,
\[
\sigma_{\min}\left(\boldsymbol{B}_{\mathbb{E},h}\right)\geqslant1-\left\Vert \boldsymbol{B}_{\mathbb{E},h}-\boldsymbol{I}\right\Vert _{2}\geqslant\frac{1}{2}.
\]
The condition of Proposition \ref{prop:Boundary-Proxy-Sufficient-Correction}
therefore holds with a constant independent of $h$. Since $d=2$,
Proposition \ref{prop:Boundary-Proxy-Sufficient-Correction} gives
\[
\inf_{\substack{r_{h}\in\mathcal{H}_{h}\\
r_{h}\left(\boldsymbol{X}_{l,h}\right)=\boldsymbol{b}
}
}\left\Vert r_{h}\right\Vert _{H^{2}\left(\Omega\right)}\lesssim h^{-1}\left\Vert \boldsymbol{b}\right\Vert _{2},
\]
which proves the result.\end{proofof}

The choice $\delta_{h}=h^{1+\alpha}$ makes the cross-evaluation matrix
converge directly to the identity $\left\Vert \boldsymbol{B}_{\mathbb{E},h}-\boldsymbol{I}\right\Vert _{2}=O\left(h^{\alpha}\right)$.
A fixed offset of the form $\delta_{h}=\varepsilon h$ also works
whenever $0<\varepsilon\leqslant1/4$ and $C_{{\rm nt}}\varepsilon<1$.
The choice $h^{1+\alpha}$ avoids the need to determine this threshold
explicitly.

\section{Proofs}

In the proofs below, $\boldsymbol{\theta}$ is fixed at an arbitrary
admissible value, and $\hat{f}\left(\cdot\right)$ denotes the corresponding
posterior mean $\hat{f}_{\boldsymbol{\theta}}\left(\cdot\right)$
defined in Section \ref{subsec:Interpretability-of-PRIK}. Under refinement,
the same parameter value is used at every level.

\subsection{Proof of Theorem \ref{thm:Interpolation+minPDE}}

\begin{proofof}{thm:Interpolation+minPDE}

Since $\tilde{k}\left(\boldsymbol{X}_{l},\boldsymbol{X}_{l}\right)$
is positive definite, $\hat{f}\left(\cdot\right)$ is well defined.
Consequently, 
\[
\hat{f}\left(\boldsymbol{X}_{l}\right)=\tilde{k}\left(\boldsymbol{X}_{l},\boldsymbol{X}_{l}\right)\tilde{k}\left(\boldsymbol{X}_{l},\boldsymbol{X}_{l}\right)^{-1}\boldsymbol{y}_{l}=\boldsymbol{y}_{l},
\]
so $\hat{f}\left(\cdot\right)$ is feasible for the constrained optimization
problem in Eq. \eqref{eq:Constrained-Opt-Vp}. Define 
\[
\mathbb{W}_{p}=\left\{ w\in\mathbb{V}_{p}:w\left(\boldsymbol{X}_{l}\right)=\boldsymbol{0}\right\} .
\]
For any $w\in\mathbb{W}_{p}$, the reproducing property of $\tilde{k}\left(\cdot,\cdot\right)$
and Eq. \eqref{eq:Desired-Properties} give 
\begin{align*}
\left<\hat{f}\left(\cdot\right),w\left(\cdot\right)\right>_{\mathscr{D}} & =\left<\tilde{k}\left(\cdot,\boldsymbol{X}_{l}\right)\tilde{k}\left(\boldsymbol{X}_{l},\boldsymbol{X}_{l}\right)^{-1}\boldsymbol{y}_{l},w\left(\cdot\right)\right>_{\mathscr{D}}\\
 & =w\left(\boldsymbol{X}_{l}\right)^{\top}\tilde{k}\left(\boldsymbol{X}_{l},\boldsymbol{X}_{l}\right)^{-1}\boldsymbol{y}_{l}=0.
\end{align*}
Equivalently, 
\[
\left<\mathscr{D}\hat{f},\mathscr{D}w\right>_{L^{2}\left(\Omega\right)}=0.
\]
Now let $u\in\mathbb{V}_{p}$ be any other feasible function. Since
\[
u\left(\boldsymbol{X}_{l}\right)=\hat{f}\left(\boldsymbol{X}_{l}\right)=\boldsymbol{y}_{l},
\]
the difference $w=u-\hat{f}$ belongs to $\mathbb{W}_{p}$. Therefore,
\begin{align*}
\left\Vert \mathscr{D}u\right\Vert _{L^{2}\left(\Omega\right)}^{2} & =\left\Vert \mathscr{D}\hat{f}+\mathscr{D}w\right\Vert _{L^{2}\left(\Omega\right)}^{2}\\
 & =\left\Vert \mathscr{D}\hat{f}\right\Vert _{L^{2}\left(\Omega\right)}^{2}+2\left<\mathscr{D}\hat{f},\mathscr{D}w\right>_{L^{2}\left(\Omega\right)}+\left\Vert \mathscr{D}w\right\Vert _{L^{2}\left(\Omega\right)}^{2}\\
 & =\left\Vert \mathscr{D}\hat{f}\right\Vert _{L^{2}\left(\Omega\right)}^{2}+\left\Vert \mathscr{D}w\right\Vert _{L^{2}\left(\Omega\right)}^{2}\geqslant\left\Vert \mathscr{D}\hat{f}\right\Vert _{L^{2}\left(\Omega\right)}^{2}.
\end{align*}
Moreover, because $\boldsymbol{A}$ is positive definite, $\left<\cdot,\cdot\right>_{\mathscr{D}}$
is an inner product on $\mathbb{V}_{p}$. Thus, the definiteness of
$\left<\cdot,\cdot\right>_{\mathscr{D}}$ on $\mathbb{V}_{p}$ implies
that equality holds only when $w=0$, or equivalently, $u=\hat{f}$.
Therefore, $\hat{f}$ is the unique solution to the problem in Eq.
\eqref{eq:Constrained-Opt-Vp}.\end{proofof}

\subsection{Proofs of Lemma \ref{lem:H2-Quasi-Optimality}, Theorem \ref{thm:Posterior-Mean-Convergence-Rate}
and Corollary \ref{cor:PDE-Residual-Convergence-Rate} \label{subsec:Proof-Posterior-Mean}}

\begin{proofof}{lem:H2-Quasi-Optimality}

Define $\mathbb{V}_{p,l}=\left\{ v\in\mathbb{V}_{p}:v\left(\boldsymbol{X}_{l}\right)=\boldsymbol{y}_{l}\right\} $.
Let $v\in\mathbb{V}_{p,l}$ be arbitrary and define 
\[
w=v-\hat{f}.
\]
Since both $v$ and $\hat{f}$ belong to $\mathbb{V}_{p,l}$,
\[
w\in\mathbb{V}_{p},\qquad w\left(\boldsymbol{X}_{l}\right)=v\left(\boldsymbol{X}_{l}\right)-\hat{f}\left(\boldsymbol{X}_{l}\right)=\boldsymbol{0}.
\]
Hence, $w\in\mathbb{W}_{p}$. Assumption \ref{assu:Stability-on-Wp}
therefore gives
\begin{equation}
\left\Vert v-\hat{f}\right\Vert _{H^{2}\left(\Omega\right)}=\left\Vert w\right\Vert _{H^{2}\left(\Omega\right)}\lesssim\left\Vert \mathscr{D}w\right\Vert _{L^{2}\left(\Omega\right)}.\label{eq:H2-Stability-Comparison}
\end{equation}
By the PDE-residual energy decomposition in Eq. \eqref{eq:PDE-Energy-Decomposition},
\[
\left\Vert \mathscr{D}v\right\Vert _{L^{2}\left(\Omega\right)}^{2}=\left\Vert \mathscr{D}\hat{f}\right\Vert _{L^{2}\left(\Omega\right)}^{2}+\left\Vert \mathscr{D}w\right\Vert _{L^{2}\left(\Omega\right)}^{2}.
\]
Consequently,
\begin{equation}
\left\Vert \mathscr{D}w\right\Vert _{L^{2}\left(\Omega\right)}\leqslant\left\Vert \mathscr{D}v\right\Vert _{L^{2}\left(\Omega\right)}.\label{eq:PDE-Residual-Projection-Bound}
\end{equation}
Since $f$ solves the homogeneous PDE in Eq. \eqref{eq:Linear-PDE},
\[
\mathscr{D}f=0\Rightarrow\mathscr{D}v=\mathscr{D}\left(v-f\right).
\]
Assumption \ref{assu:Bounded-Differential-Operator} now yields
\begin{equation}
\left\Vert \mathscr{D}v\right\Vert _{L^{2}\left(\Omega\right)}=\left\Vert \mathscr{D}\left(v-f\right)\right\Vert _{L^{2}\left(\Omega\right)}\lesssim\left\Vert v-f\right\Vert _{H^{2}\left(\Omega\right)}.\label{eq:Bounded-Operator-Comparison}
\end{equation}
Combining Eqs. \eqref{eq:H2-Stability-Comparison}, \eqref{eq:PDE-Residual-Projection-Bound},
and \eqref{eq:Bounded-Operator-Comparison} gives
\[
\left\Vert v-\hat{f}\right\Vert _{H^{2}\left(\Omega\right)}\lesssim\left\Vert v-f\right\Vert _{H^{2}\left(\Omega\right)}.
\]
Therefore, by the triangle inequality,
\[
\begin{aligned}\left\Vert \hat{f}-f\right\Vert _{H^{2}\left(\Omega\right)} & \leqslant\left\Vert \hat{f}-v\right\Vert _{H^{2}\left(\Omega\right)}+\left\Vert v-f\right\Vert _{H^{2}\left(\Omega\right)}\\
 & \lesssim\left\Vert v-f\right\Vert _{H^{2}\left(\Omega\right)}.
\end{aligned}
\]
Since $v\in\mathbb{V}_{p,l}$ was arbitrary, taking the infimum over
$\mathbb{V}_{p,l}$ yields
\[
\left\Vert \hat{f}-f\right\Vert _{H^{2}\left(\Omega\right)}\lesssim\underset{v\in\mathbb{V}_{p,l}}{\mathrm{inf}}\left\Vert v-f\right\Vert _{H^{2}\left(\Omega\right)}.
\]
This proves Eq. \eqref{eq:H2-quasi-optimality}.\end{proofof}

Before proving Theorem \ref{thm:Posterior-Mean-Convergence-Rate}
and Corollary \ref{cor:PDE-Residual-Convergence-Rate}, we first establish
several auxiliary lemmas.
\begin{lem}
\label{lem:Boundary-Sampling-Inequality}(Boundary Sampling Inequality
of Fractional Order) Suppose that $d\geqslant2$. Under Assumptions
\ref{assu:Smoothness-of-Domain} and \ref{assu:Refinement-Design},
there exists a constant $C_{S}>0$, independent of $u$ and $\boldsymbol{X}_{l,h}$
such that
\begin{equation}
\left\Vert u\right\Vert _{H^{3/2}\left(\mathbb{E}\right)}\leqslant C_{S}\left(h_{\boldsymbol{X}_{l,h},\mathbb{E}}^{m-2}\left\Vert u\right\Vert _{H^{m}\left(\Omega\right)}+h_{\boldsymbol{X}_{l,h},\mathbb{E}}^{d/2-2}\left\Vert u\right\Vert _{\boldsymbol{X}_{l,h}}\right).\label{eq:Boundary-Sampling-Inequality}
\end{equation}
\end{lem}
\begin{proofof}{lem:Boundary-Sampling-Inequality}

There exists $h_{0}>0$ such that, whenever $h_{\boldsymbol{X}_{l,h},\mathbb{E}}\leqslant h_{0}$,
the boundary sampling argument in \citet[Lemma 3.1]{cheung$H^2$ConvergenceLeastSquaresKernel2018}
with $s=2$ yields 
\[
\left\Vert u\right\Vert _{H^{3/2}\left(\mathbb{E}\right)}\leqslant C_{S,0}\left(h_{\boldsymbol{X}_{l,h},\mathbb{E}}^{m-2}\left\Vert u\right\Vert _{H^{m}\left(\Omega\right)}+h_{\boldsymbol{X}_{l,h},\mathbb{E}}^{d/2-2}\left\Vert u\right\Vert _{\boldsymbol{X}_{l,h}}\right).
\]
Although \citet{cheung$H^2$ConvergenceLeastSquaresKernel2018} adopt
an integer-valued smoothness parameter $m$, their proof of the boundary
estimate relies only on the fractional-order sampling inequalities
of \citet{arcangeliExtensionSamplingInequalities2012}. Therefore,
the argument remains valid for the real-valued $m$ considered here.

It remains to consider $h_{\boldsymbol{X}_{l,h},\mathbb{E}}>h_{0}$.
By the boundedness of the boundary restriction from $H^{m}\left(\Omega\right)$
to $H^{3/2}\left(\mathbb{E}\right)$, 
\[
\left\Vert u\right\Vert _{H^{3/2}\left(\mathbb{E}\right)}\leqslant C_{{\rm tr}}\left\Vert u\right\Vert _{H^{m}\left(\Omega\right)}.
\]
Since $m-2>0$ and $h_{\boldsymbol{X}_{l,h},\mathbb{E}}>h_{0}$, 
\[
1\leqslant h_{0}^{-\left(m-2\right)}h_{\boldsymbol{X}_{l,h},\mathbb{E}}^{m-2}.
\]
Consequently,
\[
\begin{aligned}\left\Vert u\right\Vert _{H^{3/2}\left(\mathbb{E}\right)} & \leqslant C_{{\rm tr}}h_{0}^{-\left(m-2\right)}h_{\boldsymbol{X}_{l,h},\mathbb{E}}^{m-2}\left\Vert u\right\Vert _{H^{m}\left(\Omega\right)}\\
 & \leqslant C_{{\rm tr}}h_{0}^{-\left(m-2\right)}\left(h_{\boldsymbol{X}_{l,h},\mathbb{E}}^{m-2}\left\Vert u\right\Vert _{H^{m}\left(\Omega\right)}+h_{\boldsymbol{X}_{l,h},\mathbb{E}}^{d/2-2}\left\Vert u\right\Vert _{\boldsymbol{X}_{l,h}}\right).
\end{aligned}
\]
Therefore, Eq. \eqref{eq:Boundary-Sampling-Inequality} holds with
\[
C_{S}=\max\left\{ C_{S,0},C_{{\rm tr}}h_{0}^{-\left(m-2\right)}\right\} .
\]
\end{proofof}
\begin{lem}
\label{lem:Inverse-Inequality}(Inverse Inequality) Under Assumptions
\ref{assu:Smoothness-of-Domain}, \ref{assu:Sobolev-Compatible-Kernel},
and \ref{assu:Refinement-Design}, there exists a constant $C_{I}$,
independent of $h$ and $u_{h}$, such that
\begin{equation}
\left\Vert u_{h}\right\Vert _{H^{m}\left(\Omega\right)}\leqslant C_{I}\frac{1}{q_{\boldsymbol{X}_{\mathrm{int},h}}^{m-2}}\left\Vert u_{h}\right\Vert _{H^{2}\left(\Omega\right)}\label{eq:Inverse-Inequality}
\end{equation}
for all $u_{h}\in\mathcal{H}_{h}$.
\end{lem}
\begin{proofof}{lem:Inverse-Inequality}

This result follows directly from \citet[Theorem 4.4]{sunInverseInequalitiesKernel2025}
by taking $X=\boldsymbol{X}_{\mathrm{int},h}$, $\alpha=m$, and $t=2$.
\end{proofof}

Lemma \ref{lem:Inverse-Inequality} allows the higher-order Sobolev
norm $H^{m}\left(\Omega\right)$ to be reduced to the target norm
$H^{2}\left(\Omega\right)$ within the finite-dimensional function
space $\mathcal{H}_{h}$, while the factor $q_{\boldsymbol{X}_{\mathrm{int},h}}^{-\left(m-2\right)}$
makes the discretization dependence explicit.
\begin{lem}
\label{lem:Uniform-Stability-on-Wph} (Uniform Stability on $\mathbb{W}_{p_{h}}$)
For each refinement, define 
\[
\mathbb{W}_{p_{h}}=\left\{ w_{h}\in\mathcal{H}_{h}:w_{h}\left(\boldsymbol{X}_{l,h}\right)=\boldsymbol{0}\right\} .
\]
Suppose that Assumptions \ref{assu:Smoothness-of-Domain}, \ref{assu:Sobolev-Compatible-Kernel},
\ref{assu:Refinement-Design} and \ref{assu:Boundary-Regularity-Estimate}
hold. When $d\geqslant2$, suppose additionally that Assumption \ref{assu:Separated-Stability-Smallness}
holds. Then, there exists a constant $C_{\mathrm{stab}}>0$, independent
of $h$, such that 
\begin{equation}
\left\Vert w_{h}\right\Vert _{H^{2}\left(\Omega\right)}\leqslant C_{\mathrm{stab}}\left\Vert \mathscr{D}w_{h}\right\Vert _{L^{2}\left(\Omega\right)},\quad w_{h}\in\mathbb{W}_{p_{h}}.\label{eq:Uniform-Stability-on-Wph}
\end{equation}
\end{lem}
\begin{proofof}{lem:Uniform-Stability-on-Wph}

Let $w_{h}\in\mathbb{W}_{p_{h}}$. 

First, suppose that $d\geqslant2$. By Assumption \ref{assu:Boundary-Regularity-Estimate},
\begin{equation}
\left\Vert w_{h}\right\Vert _{H^{2}\left(\Omega\right)}\leqslant C_{B}\left(\left\Vert \mathscr{D}w_{h}\right\Vert _{L^{2}\left(\Omega\right)}+\left\Vert w_{h}\right\Vert _{H^{3/2}\left(\mathbb{E}\right)}\right).\label{eq:Boundary-Regularity-on-Wph}
\end{equation}
Since $w_{h}\left(\boldsymbol{X}_{l,h}\right)=\boldsymbol{0}$, Eq.
\eqref{eq:Boundary-Sampling-Inequality} gives 
\[
\left\Vert w_{h}\right\Vert _{H^{3/2}\left(\mathbb{E}\right)}\leqslant C_{S}h_{\boldsymbol{X}_{l,h},\mathbb{E}}^{m-2}\left\Vert w_{h}\right\Vert _{H^{m}\left(\Omega\right)}.
\]
Applying Eq. \eqref{eq:Inverse-Inequality} to $w_{h}\in\mathcal{H}_{h}$,
we obtain
\begin{equation}
\left\Vert w_{h}\right\Vert _{H^{3/2}\left(\mathbb{E}\right)}\leqslant C_{S}C_{I}\left(\frac{h_{\boldsymbol{X}_{l,h},\mathbb{E}}}{q_{\boldsymbol{X}_{\mathrm{int},h}}}\right)^{m-2}\left\Vert w_{h}\right\Vert _{H^{2}\left(\Omega\right)}.\label{eq:Boundary-Trace-Control-on-Wph}
\end{equation}
Combining Eq. \eqref{eq:Boundary-Regularity-on-Wph} and Eq. \eqref{eq:Boundary-Trace-Control-on-Wph}
yields
\begin{equation}
\left\Vert w_{h}\right\Vert _{H^{2}\left(\Omega\right)}\leqslant C_{B}\left\Vert \mathscr{D}w_{h}\right\Vert _{L^{2}\left(\Omega\right)}+C_{B}C_{S}C_{I}\left(\frac{h_{\boldsymbol{X}_{l,h},\mathbb{E}}}{q_{\boldsymbol{X}_{\mathrm{int},h}}}\right)^{m-2}\left\Vert w_{h}\right\Vert _{H^{2}\left(\Omega\right)}.\label{eq:Pre-Absorption-Stability-on-Wph}
\end{equation}
Assumption \ref{assu:Separated-Stability-Smallness} implies that
\[
C_{B}C_{S}C_{I}\left(\frac{h_{\boldsymbol{X}_{l,h},\mathbb{E}}}{q_{\boldsymbol{X}_{\mathrm{int},h}}}\right)^{m-2}<\frac{1}{2}.
\]
Therefore, the second term on the right-hand side of Eq. \eqref{eq:Pre-Absorption-Stability-on-Wph}
can be absorbed into the left-hand side, which gives
\[
\left\Vert w_{h}\right\Vert _{H^{2}\left(\Omega\right)}\leqslant2C_{B}\left\Vert \mathscr{D}w_{h}\right\Vert _{L^{2}\left(\Omega\right)}.
\]

Now suppose that $d=1$. By Assumption \ref{assu:Refinement-Design},
the set $\boldsymbol{X}_{l,h}$ consists of the two boundary endpoints.
Hence,
\[
\boldsymbol{X}_{l,h}=\mathbb{E}.
\]
Since $w_{h}\left(\boldsymbol{X}_{l,h}\right)=\boldsymbol{0}$, it
follows that
\[
\left.w_{h}\right|_{\mathbb{E}}=0.
\]
Assumption \ref{assu:Boundary-Regularity-Estimate} therefore gives
directly
\[
\left\Vert w_{h}\right\Vert _{H^{2}\left(\Omega\right)}\leqslant C_{B}\left\Vert \mathscr{D}w_{h}\right\Vert _{L^{2}\left(\Omega\right)}\leqslant2C_{B}\left\Vert \mathscr{D}w_{h}\right\Vert _{L^{2}\left(\Omega\right)}.
\]

Thus Eq. \eqref{eq:Uniform-Stability-on-Wph} holds for all $1\leqslant d\leqslant3$
with $C_{{\rm stab}}=2C_{B}$. \end{proofof}

Lemma \ref{lem:Uniform-Stability-on-Wph} verifies Assumption \ref{assu:Stability-on-Wp}
uniformly over the refinement family, with $\mathbb{V}_{p_{h}}=\mathcal{H}_{h}$
and the corresponding label-homogeneous subspace $\mathbb{W}_{p_{h}}$.
So the $H^{2}$-quasi-optimality result also applies to kernel-section
PRIK for the second-order elliptic Dirichlet problem considered here.
Moreover, Lemma \ref{lem:Uniform-Stability-on-Wph} is the key estimate
in the analysis of the posterior mean. It combines the PDE-side $H^{2}$
a priori estimate with two meshfree ingredients: a boundary sampling
inequality and an inverse inequality in the function space.
\begin{lem}
\label{lem:Interior-Kernel-Interpolation-Estimates} (Interior Kernel
Interpolation Estimates) Under Assumptions \ref{assu:Smoothness-of-Domain}
and \ref{assu:Sobolev-Compatible-Kernel}, and the interior fill-distance
condition in Assumption \ref{assu:Refinement-Design}, let $\mathcal{I}_{\boldsymbol{X}_{\mathrm{int},h}}u\in\mathcal{H}_{h}$
denote the kernel interpolation function  of $u$ at $\boldsymbol{X}_{\mathrm{int},h}$,
satisfying 
\[
\mathcal{I}_{\boldsymbol{X}_{\mathrm{int},h}}u\left(\boldsymbol{\xi}\right)=u\left(\boldsymbol{\xi}\right),\quad\forall\boldsymbol{\xi}\in\boldsymbol{X}_{\mathrm{int},h}.
\]
Then, for every $u\in H^{m}\left(\Omega\right)$ and all sufficiently
small $h$,
\begin{equation}
\left\Vert u-\mathcal{I}_{\boldsymbol{X}_{\mathrm{int},h}}u\right\Vert _{L^{2}\left(\Omega\right)}+h\left\Vert u-\mathcal{I}_{\boldsymbol{X}_{\mathrm{int},h}}u\right\Vert _{H^{1}\left(\Omega\right)}+h^{2}\left\Vert u-\mathcal{I}_{\boldsymbol{X}_{\mathrm{int},h}}u\right\Vert _{H^{2}\left(\Omega\right)}\lesssim h^{m}\left\Vert u\right\Vert _{H^{m}\left(\Omega\right)}.\label{eq:Interior-Interpolation-Multinorm-Rate}
\end{equation}
Moreover, 
\begin{equation}
\left\Vert u-\mathcal{I}_{\boldsymbol{X}_{\mathrm{int},h}}u\right\Vert _{H^{m}\left(\Omega\right)}\lesssim\left\Vert u\right\Vert _{H^{m}\left(\Omega\right)},\label{eq:Interior-Interpolation-Hm-Stability}
\end{equation}
and 
\begin{equation}
\left\Vert u-\mathcal{I}_{\boldsymbol{X}_{\mathrm{int},h}}u\right\Vert _{L^{\infty}\left(\Omega\right)}\lesssim h^{m-d/2}\left\Vert u\right\Vert _{H^{m}\left(\Omega\right)}.\label{eq:Interior-Interpolation-Linfty-Rate}
\end{equation}
\end{lem}
\begin{proofof}{lem:Interior-Kernel-Interpolation-Estimates} By \citet[Lemma 4.1]{narcowichSobolevErrorEstimates2006},
applied with 
\[
\tau=m,\quad X=\boldsymbol{X}_{\mathrm{int},h},\quad h_{X,\Omega}=h,
\]
we have 
\begin{equation}
\left\Vert u-\mathcal{I}_{\boldsymbol{X}_{\mathrm{int},h}}u\right\Vert _{H^{\beta}\left(\Omega\right)}\lesssim h^{m-\beta}\left\Vert u\right\Vert _{H^{m}\left(\Omega\right)},\qquad0\leqslant\beta\leqslant m.\label{eq:Interior-Interpolation-General-Sobolev-Rate}
\end{equation}
Taking $\beta=0,1,2$ in Eq. \eqref{eq:Interior-Interpolation-General-Sobolev-Rate},
respectively, gives
\begin{align*}
\left\Vert u-\mathcal{I}_{\boldsymbol{X}_{\mathrm{int},h}}u\right\Vert _{L^{2}\left(\Omega\right)} & \lesssim h^{m}\left\Vert u\right\Vert _{H^{m}\left(\Omega\right)},\\
\left\Vert u-\mathcal{I}_{\boldsymbol{X}_{\mathrm{int},h}}u\right\Vert _{H^{1}\left(\Omega\right)} & \lesssim h^{m-1}\left\Vert u\right\Vert _{H^{m}\left(\Omega\right)},\\
\left\Vert u-\mathcal{I}_{\boldsymbol{X}_{\mathrm{int},h}}u\right\Vert _{H^{2}\left(\Omega\right)} & \lesssim h^{m-2}\left\Vert u\right\Vert _{H^{m}\left(\Omega\right)}.
\end{align*}
Multiplying the second and third inequalities by $h$ and $h^{2}$,
respectively, and adding the resulting estimates proves Eq. \eqref{eq:Interior-Interpolation-Multinorm-Rate}.

Taking $\beta=m$ in Eq. \eqref{eq:Interior-Interpolation-General-Sobolev-Rate}
gives 
\[
\left\Vert u-\mathcal{I}_{\boldsymbol{X}_{\mathrm{int},h}}u\right\Vert _{H^{m}\left(\Omega\right)}\lesssim\left\Vert u\right\Vert _{H^{m}\left(\Omega\right)},
\]
which proves Eq. \eqref{eq:Interior-Interpolation-Hm-Stability}.

Finally, \citet[Proposition 3.2]{narcowichSobolevBoundsScatteredZeros2005},
applied with 
\[
\tau=m,\quad\left|\boldsymbol{\alpha}\right|=0,\quad q=\infty,
\]
gives
\[
\left\Vert u-\mathcal{I}_{\boldsymbol{X}_{\mathrm{int},h}}u\right\Vert _{L^{\infty}\left(\Omega\right)}\lesssim h^{m-d/2}\left\Vert u\right\Vert _{H^{m}\left(\Omega\right)}.
\]
Its smoothness condition is satisfied because
\[
m>2+\frac{d}{2}.
\]
This proves Eq. \eqref{eq:Interior-Interpolation-Linfty-Rate}. \end{proofof}
\begin{lem}
\label{lem:Upper-Discrete-Sampling-on-Boundary} (Upper Discrete Sampling
on $\mathbb{E}$) Suppose that $d\geqslant2$. Under Assumptions \ref{assu:Smoothness-of-Domain}
and \ref{assu:Refinement-Design}, there exists a constant $C_{\mathrm{us}}>0$,
independent of the fill distance $h$ and the boundary function $g$,
such that
\begin{equation}
\left\Vert g\right\Vert _{\boldsymbol{X}_{l,h}}\leqslant C_{\mathrm{us}}\left(q_{\boldsymbol{X}_{l,h}}^{-\left(d-1\right)/2}\left\Vert g\right\Vert _{L^{2}\left(\mathbb{E}\right)}+q_{\boldsymbol{X}_{l,h}}^{m-d/2}\left\Vert g\right\Vert _{H^{m-1/2}\left(\mathbb{E}\right)}\right)\label{eq:Upper-Discrete-Sampling-on-Boundary}
\end{equation}
for every $g\in H^{m-1/2}\left(\mathbb{E}\right)$ and all sufficiently
small $h$.
\end{lem}
\begin{proofof}{lem:Upper-Discrete-Sampling-on-Boundary}This estimate
is the boundary-chart version of the scale-refined one-sided sampling
inequality of \citet[Theorem 2.1]{ArcangeliTorrens2014Sampling}.
Apply the argument underlying the cited theorem to the pullback of
$g$ on each boundary chart, with spatial dimension $d-1$, smoothness
order $m-\frac{1}{2}$, and separation radius $q_{\boldsymbol{X}_{l,h}}$.
Since
\[
m-\frac{1}{2}>\frac{d-1}{2},
\]
the required point evaluations are continuous. The scaled estimate
on each chart gives the factors
\[
q_{\boldsymbol{X}_{l,h}}^{-\left(d-1\right)/2}\quad\text{and}\quad q_{\boldsymbol{X}_{l,h}}^{m-\frac{1}{2}-\frac{d-1}{2}}=q_{\boldsymbol{X}_{l,h}}^{m-d/2}.
\]
The uniform equivalence of the local and boundary Sobolev norms, together
with the finite boundary atlas and the separation of the labeled points,
allows the local estimates to be summed with a constant independent
of $h$. This completes the proof.\end{proofof}
\begin{lem}
\label{lem:Sampled-Boundary-Mismatch}(Sampled Boundary Mismatch)
Under Assumptions \ref{assu:Smoothness-of-Domain}, \ref{assu:Sobolev-Compatible-Kernel},
and \ref{assu:Refinement-Design}, there exists a constant $C_{\mathrm{bm}}>0$,
independent of $h$ and $u$, such that
\begin{equation}
\left\Vert u-\mathcal{I}_{\boldsymbol{X}_{\mathrm{int},h}}u\right\Vert _{\boldsymbol{X}_{l,h}}\leqslant C_{\mathrm{bm}}h^{m-d/2}\left\Vert u\right\Vert _{H^{m}\left(\Omega\right)}\label{eq:Sampled-Boundary-Mismatch}
\end{equation}
for every $u\in H^{m}\left(\Omega\right)$ and all sufficiently small
$h$.
\end{lem}
\begin{proofof}{lem:Sampled-Boundary-Mismatch}First, suppose that
$d\geqslant2$. By the multiplicative trace inequality on $\Omega$,
\[
\left\Vert u-\mathcal{I}_{\boldsymbol{X}_{\mathrm{int},h}}u\right\Vert _{L^{2}\left(\mathbb{E}\right)}\lesssim\left\Vert u-\mathcal{I}_{\boldsymbol{X}_{\mathrm{int},h}}u\right\Vert _{L^{2}\left(\Omega\right)}^{1/2}\left\Vert u-\mathcal{I}_{\boldsymbol{X}_{\mathrm{int},h}}u\right\Vert _{H^{1}\left(\Omega\right)}^{1/2}.
\]
Eq. \eqref{eq:Interior-Interpolation-Multinorm-Rate} implies
\begin{align*}
\left\Vert u-\mathcal{I}_{\boldsymbol{X}_{\mathrm{int},h}}u\right\Vert _{L^{2}\left(\Omega\right)} & \lesssim h^{m}\left\Vert u\right\Vert _{H^{m}\left(\Omega\right)},\\
\left\Vert u-\mathcal{I}_{\boldsymbol{X}_{\mathrm{int},h}}u\right\Vert _{H^{1}\left(\Omega\right)} & \lesssim h^{m-1}\left\Vert u\right\Vert _{H^{m}\left(\Omega\right)}.
\end{align*}
Consequently,
\begin{equation}
\left\Vert u-\mathcal{I}_{\boldsymbol{X}_{\mathrm{int},h}}u\right\Vert _{L^{2}\left(\mathbb{E}\right)}\lesssim h^{m-1/2}\left\Vert u\right\Vert _{H^{m}\left(\Omega\right)}.\label{eq:Boundary-L2-Interpolation-Mismatch}
\end{equation}
On the other hand, the Sobolev trace theorem and Eq. \eqref{eq:Interior-Interpolation-Hm-Stability}
give
\begin{equation}
\left\Vert u-\mathcal{I}_{\boldsymbol{X}_{\mathrm{int},h}}u\right\Vert _{H^{m-1/2}\left(\mathbb{E}\right)}\lesssim\left\Vert u-\mathcal{I}_{\boldsymbol{X}_{\mathrm{int},h}}u\right\Vert _{H^{m}\left(\Omega\right)}\lesssim\left\Vert u\right\Vert _{H^{m}\left(\Omega\right)}.\label{eq:Boundary-Hmhalf-Interpolation-Mismatch}
\end{equation}
By Lemma \ref{lem:Upper-Discrete-Sampling-on-Boundary}, applied to
$g=\left.\left(u-\mathcal{I}_{\boldsymbol{X}_{\mathrm{int},h}}u\right)\right|_{\mathbb{E}}$,
Eqs. \eqref{eq:Boundary-L2-Interpolation-Mismatch} and \eqref{eq:Boundary-Hmhalf-Interpolation-Mismatch}
imply
\[
\left\Vert u-\mathcal{I}_{\boldsymbol{X}_{\mathrm{int},h}}u\right\Vert _{\boldsymbol{X}_{l,h}}\lesssim\left(q_{\boldsymbol{X}_{l,h}}^{-\left(d-1\right)/2}h^{m-1/2}+q_{\boldsymbol{X}_{l,h}}^{m-d/2}\right)\left\Vert u\right\Vert _{H^{m}\left(\Omega\right)}.
\]
By Assumption \ref{assu:Refinement-Design}, $q_{\boldsymbol{X}_{l,h}}\asymp h$.
It follows that
\[
q_{\boldsymbol{X}_{l,h}}^{-\left(d-1\right)/2}h^{m-1/2}\lesssim h^{m-d/2},\quad q_{\boldsymbol{X}_{l,h}}^{m-d/2}\lesssim h^{m-d/2}.
\]
Therefore,
\[
\left\Vert u-\mathcal{I}_{\boldsymbol{X}_{\mathrm{int},h}}u\right\Vert _{\boldsymbol{X}_{l,h}}\lesssim h^{m-d/2}\left\Vert u\right\Vert _{H^{m}\left(\Omega\right)}.
\]

Now suppose that $d=1$. By Assumption \ref{assu:Refinement-Design},
the set $\boldsymbol{X}_{l,h}$ consists of the two boundary endpoints.
Hence,
\[
\left\Vert u-\mathcal{I}_{\boldsymbol{X}_{\mathrm{int},h}}u\right\Vert _{\boldsymbol{X}_{l,h}}\leqslant\sqrt{2}\left\Vert u-\mathcal{I}_{\boldsymbol{X}_{\mathrm{int},h}}u\right\Vert _{L^{\infty}\left(\Omega\right)}.
\]
Eq. \eqref{eq:Interior-Interpolation-Linfty-Rate} therefore gives
\[
\left\Vert u-\mathcal{I}_{\boldsymbol{X}_{\mathrm{int},h}}u\right\Vert _{\boldsymbol{X}_{l,h}}\lesssim h^{m-1/2}\left\Vert u\right\Vert _{H^{m}\left(\Omega\right)}=h^{m-d/2}\left\Vert u\right\Vert _{H^{m}\left(\Omega\right)}.
\]

Thus Eq. \eqref{eq:Sampled-Boundary-Mismatch} holds for all $1\leqslant d\leqslant3$.\end{proofof}
\begin{lem}
\label{lem:Label-Compatible-Approximation}(Label-Compatible Approximation)
Under Assumptions \ref{assu:Smoothness-of-Domain}, \ref{assu:Sobolev-Compatible-Kernel},
\ref{assu:Refinement-Design}, and \ref{assu:Stable-Discrete-Boundary-Correction},
there exists a constant $C_{\mathrm{app}}>0$, independent of $h$
and $u$, such that, for every $u\in H^{m}\left(\Omega\right)$ and
all sufficiently small $h$, there exists $v_{h}\in\mathcal{H}_{h}$
satisfying $v_{h}\left(\boldsymbol{X}_{l,h}\right)=u\left(\boldsymbol{X}_{l,h}\right)$
and 
\begin{equation}
\left\Vert u-v_{h}\right\Vert _{H^{2}\left(\Omega\right)}\leqslant C_{\mathrm{app}}h^{m-2}\left\Vert u\right\Vert _{H^{m}\left(\Omega\right)}.\label{eq:Label-Compatible-Approximation}
\end{equation}
\end{lem}
\begin{proofof}{lem:Label-Compatible-Approximation}Define the sampled
boundary mismatch by 
\[
\boldsymbol{b}_{h}=\left(u-\mathcal{I}_{\boldsymbol{X}_{\mathrm{int},h}}u\right)\left(\boldsymbol{X}_{l,h}\right).
\]
By Lemma \ref{lem:Sampled-Boundary-Mismatch},
\[
\left\Vert \boldsymbol{b}_{h}\right\Vert _{2}\lesssim h^{m-d/2}\left\Vert u\right\Vert _{H^{m}\left(\Omega\right)}.
\]
Assumption \ref{assu:Stable-Discrete-Boundary-Correction} therefore
provides $r_{h}\in\mathcal{H}_{h}$ such that $r_{h}\left(\boldsymbol{X}_{l,h}\right)=\boldsymbol{b}_{h}$
and 
\begin{equation}
\left\Vert r_{h}\right\Vert _{H^{2}\left(\Omega\right)}\lesssim h^{d/2-2}\left\Vert \boldsymbol{b}_{h}\right\Vert _{2}\lesssim h^{d/2-2}h^{m-d/2}\left\Vert u\right\Vert _{H^{m}\left(\Omega\right)}=h^{m-2}\left\Vert u\right\Vert _{H^{m}\left(\Omega\right)}.\label{eq:Boundary-Correction-H2-Bound}
\end{equation}
Now define
\[
v_{h}=\mathcal{I}_{\boldsymbol{X}_{\mathrm{int},h}}u+r_{h}.
\]
Since $\mathcal{I}_{\boldsymbol{X}_{\mathrm{int},h}}u$ and $r_{h}$
belong to $\mathcal{H}_{h}$, we have $v_{h}\in\mathcal{H}_{h}$.
Moreover,
\[
\begin{aligned}v_{h}\left(\boldsymbol{X}_{l,h}\right) & =\mathcal{I}_{\boldsymbol{X}_{\mathrm{int},h}}u\left(\boldsymbol{X}_{l,h}\right)+r_{h}\left(\boldsymbol{X}_{l,h}\right)\\
 & =\mathcal{I}_{\boldsymbol{X}_{\mathrm{int},h}}u\left(\boldsymbol{X}_{l,h}\right)+\left(u-\mathcal{I}_{\boldsymbol{X}_{\mathrm{int},h}}u\right)\left(\boldsymbol{X}_{l,h}\right)\\
 & =u\left(\boldsymbol{X}_{l,h}\right).
\end{aligned}
\]
Thus $v_{h}$ satisfies the required boundary labels. Finally, the
triangle inequality, Eq. \eqref{eq:Interior-Interpolation-Multinorm-Rate},
and Eq. \eqref{eq:Boundary-Correction-H2-Bound} give
\[
\begin{aligned}\left\Vert u-v_{h}\right\Vert _{H^{2}\left(\Omega\right)} & \leqslant\left\Vert u-\mathcal{I}_{\boldsymbol{X}_{\mathrm{int},h}}u\right\Vert _{H^{2}\left(\Omega\right)}+\left\Vert r_{h}\right\Vert _{H^{2}\left(\Omega\right)}\\
 & \lesssim h^{m-2}\left\Vert u\right\Vert _{H^{m}\left(\Omega\right)}.
\end{aligned}
\]
This completes the proof.\end{proofof}

Now, we are ready to prove Theorem \ref{thm:Posterior-Mean-Convergence-Rate}.

\begin{proofof}{thm:Posterior-Mean-Convergence-Rate}By Assumption
\ref{assu:classical-solution-existence}, $f\in H^{m}\left(\Omega\right)$.
By Lemma \ref{lem:Label-Compatible-Approximation}, applied with $u=f$,
there exists $v_{h}\in\mathcal{H}_{h}$ satisfying $v_{h}\left(\boldsymbol{X}_{l,h}\right)=f\left(\boldsymbol{X}_{l,h}\right)=\boldsymbol{y}_{l,h}$
and 
\begin{equation}
\left\Vert f-v_{h}\right\Vert _{H^{2}\left(\Omega\right)}\lesssim h^{m-2}\left\Vert f\right\Vert _{H^{m}\left(\Omega\right)}.\label{eq:Label-Compatible-H2-Approximation}
\end{equation}
Thus $v_{h}$ belongs to the label-compatible affine space
\[
\mathbb{V}_{p_{h},l}=\left\{ v\in\mathcal{H}_{h}:v\left(\boldsymbol{X}_{l,h}\right)=\boldsymbol{y}_{l,h}\right\} .
\]
Lemma \ref{lem:Uniform-Stability-on-Wph} verifies Assumption \ref{assu:Stability-on-Wp}
uniformly for the refinement family $\mathbb{V}_{p_{h}}=\mathcal{H}_{h}$
and the corresponding label-homogeneous subspace $\mathbb{W}_{p_{h}}$.
Therefore, the abstract $H^{2}$-quasi-optimality result in Lemma
\ref{lem:H2-Quasi-Optimality}, applied to $v_{h}$, gives
\[
\left\Vert f-\hat{f}_{h}\right\Vert _{H^{2}\left(\Omega\right)}\lesssim\left\Vert f-v_{h}\right\Vert _{H^{2}\left(\Omega\right)}.
\]
Combining this estimate with Eq. \eqref{eq:Label-Compatible-H2-Approximation}
yields
\[
\left\Vert f-\hat{f}_{h}\right\Vert _{H^{2}\left(\Omega\right)}\lesssim h^{m-2}\left\Vert f\right\Vert _{H^{m}\left(\Omega\right)}.
\]
This completes the proof.\end{proofof}

Finally, we turn to the proof of the convergence rate of the PDE-residual
norm.

\begin{proofof}{cor:PDE-Residual-Convergence-Rate}Since $f$ solves
the homogeneous PDE in Eq. \eqref{eq:Linear-PDE}, 
\[
\mathscr{D}f=0\quad\text{in }\Omega.
\]
Therefore,
\[
\left\Vert \mathscr{D}\hat{f}_{h}\right\Vert _{L^{2}\left(\Omega\right)}=\left\Vert \mathscr{D}\left(\hat{f}_{h}-f\right)\right\Vert _{L^{2}\left(\Omega\right)}.
\]
Assumption \ref{assu:Bounded-Differential-Operator} gives
\[
\left\Vert \mathscr{D}\left(\hat{f}_{h}-f\right)\right\Vert _{L^{2}\left(\Omega\right)}\lesssim\left\Vert \hat{f}_{h}-f\right\Vert _{H^{2}\left(\Omega\right)}.
\]
Applying Theorem \ref{thm:Posterior-Mean-Convergence-Rate}, we obtain
\[
\left\Vert \mathscr{D}\hat{f}_{h}\right\Vert _{L^{2}\left(\Omega\right)}\lesssim h^{m-2}\left\Vert f\right\Vert _{H^{m}\left(\Omega\right)}.
\]
This completes the proof.\end{proofof}

\subsection{Proof of Theorem \ref{thm:posterior-kernel-convergence} and Corollary
\ref{cor:Decay-Predictive-Sigma2}}

Before presenting the proof of Theorem \ref{thm:posterior-kernel-convergence},
we establish several auxiliary lemmas.
\begin{lem}
\label{lem:Posterior-Kernel-Reproducing-on-Wph}(Reproducing Property
of the Normalized Posterior Kernel on $\mathbb{W}_{p_{h}}$) Under
Assumption \ref{assu:Stable-Discrete-Boundary-Correction}, the matrix
$\tilde{k}_{h}\left(\boldsymbol{X}_{l,h},\boldsymbol{X}_{l,h}\right)$
is positive definite for all sufficiently small $h$. Moreover, for
every $\boldsymbol{\xi}\in\bar{\Omega}$, $c_{h}\left(\cdot,\boldsymbol{\xi}\right)\in\mathbb{W}_{p_{h}}$
and 
\begin{equation}
\left<w_{h},c_{h}\left(\cdot,\boldsymbol{\xi}\right)\right>_{\mathscr{D}}=w_{h}\left(\boldsymbol{\xi}\right),\quad w_{h}\in\mathbb{W}_{p_{h}}.\label{eq:Posterior-Kernel-Reproducing-on-Wph}
\end{equation}
\end{lem}
\begin{proofof}{lem:Posterior-Kernel-Reproducing-on-Wph}Assumption
\ref{assu:Stable-Discrete-Boundary-Correction} guarantees that, for
every $\boldsymbol{b}\in\mathbb{R}^{N_{l,h}}$, the constraint set
\[
\left\{ v_{h}\in\mathcal{H}_{h}:v_{h}\left(\boldsymbol{X}_{l,h}\right)=\boldsymbol{b}\right\} 
\]
is nonempty. Hence, the boundary evaluation map

\[
w_{h}\mapsto w_{h}\left(\boldsymbol{X}_{l,h}\right)
\]
from $\mathcal{H}_{h}$ onto $\mathbb{R}^{N_{l,h}}$ is surjective.
We first use this fact to verify the positive definiteness of $\tilde{k}_{h}\left(\boldsymbol{X}_{l,h},\boldsymbol{X}_{l,h}\right)$. 

Let $\boldsymbol{a}\in\mathbb{R}^{N_{l,h}}$ be arbitrary and define
\[
r_{h}=\tilde{k}_{h}\left(\cdot,\boldsymbol{X}_{l,h}\right)\boldsymbol{a}\in\mathcal{H}_{h}.
\]
By the reproducing property of the PDE-residual kernel in Eq. \eqref{eq:Reproducing-Property},
\[
\left\Vert \mathscr{D}r_{h}\right\Vert _{L^{2}\left(\Omega\right)}^{2}=\left<r_{h},r_{h}\right>_{\mathscr{D}}=\boldsymbol{a}^{\top}\tilde{k}_{h}\left(\boldsymbol{X}_{l,h},\boldsymbol{X}_{l,h}\right)\boldsymbol{a}.
\]
Suppose that the left-hand side is zero. Since the PDE-residual inner
product is an inner product on $\mathcal{H}_{h}$, this implies $r_{h}=0$.
Hence, for every $v_{h}\in\mathcal{H}_{h}$,
\[
0=\left<v_{h},r_{h}\right>_{\mathscr{D}}=v_{h}\left(\boldsymbol{X}_{l,h}\right)^{\top}\boldsymbol{a}.
\]
By surjectivity, there exists $v_{h}\in\mathcal{H}_{h}$ satisfying
$v_{h}\left(\boldsymbol{X}_{l,h}\right)=\boldsymbol{a}$. Consequently,
$\left\Vert \boldsymbol{a}\right\Vert _{2}^{2}=0$, and therefore
$\boldsymbol{a}=\boldsymbol{0}$. Thus, $\tilde{k}_{h}\left(\boldsymbol{X}_{l,h},\boldsymbol{X}_{l,h}\right)$
is positive definite.

We next consider the normalized posterior kernel defined in Eq. \eqref{eq:Normalized-Posterior-Kernel}.
Evaluating its first argument at the labeled points gives
\[
c_{h}\left(\boldsymbol{X}_{l,h},\boldsymbol{\xi}\right)=\tilde{k}_{h}\left(\boldsymbol{X}_{l,h},\boldsymbol{\xi}\right)-\tilde{k}_{h}\left(\boldsymbol{X}_{l,h},\boldsymbol{X}_{l,h}\right)\tilde{k}_{h}\left(\boldsymbol{X}_{l,h},\boldsymbol{X}_{l,h}\right)^{-1}\tilde{k}_{h}\left(\boldsymbol{X}_{l,h},\boldsymbol{\xi}\right)=\boldsymbol{0}.
\]
Therefore, 
\[
c_{h}\left(\cdot,\boldsymbol{\xi}\right)\in\mathbb{W}_{p_{h}}.
\]
Finally, for any $w_{h}\in\mathbb{W}_{p_{h}}$, Eq. \eqref{eq:Reproducing-Property}
gives
\[
\left<w_{h},c_{h}\left(\cdot,\boldsymbol{\xi}\right)\right>_{\mathscr{D}}=w_{h}\left(\boldsymbol{\xi}\right)-w_{h}\left(\boldsymbol{X}_{l,h}\right)^{\top}\tilde{k}_{h}\left(\boldsymbol{X}_{l,h},\boldsymbol{X}_{l,h}\right)^{-1}\tilde{k}_{h}\left(\boldsymbol{X}_{l,h},\boldsymbol{\xi}\right)=w_{h}\left(\boldsymbol{\xi}\right),
\]
because $w_{h}\left(\boldsymbol{X}_{l,h}\right)=\boldsymbol{0}$.
This completes the proof.\end{proofof}
\begin{lem}
\label{lem:Homogeneous-Recovery-Sequence}(Homogeneous Recovery Sequence)
Under the assumptions of Theorem \ref{thm:posterior-kernel-convergence},
for every $w\in\mathbb{W}_{\mathrm{reg}}$ and all sufficiently small
$h$, there exists $w_{h}\in\mathbb{W}_{p_{h}}$ such that
\[
\left\Vert w-w_{h}\right\Vert _{H^{2}\left(\Omega\right)}\lesssim h^{m-2}\left\Vert w\right\Vert _{H^{m}\left(\Omega\right)}.
\]
In particular,
\[
w_{h}\to w\quad\text{in }H^{2}\left(\Omega\right).
\]
\end{lem}
\begin{proofof}{lem:Homogeneous-Recovery-Sequence}Let $w\in\mathbb{W}_{\mathrm{reg}}$.
Since $\mathbb{W}_{\mathrm{reg}}\subseteq\mathbb{W}\cap H^{m}\left(\Omega\right)$,
$w\in H^{m}\left(\Omega\right)$ and 
\[
\left.w\right|_{\mathbb{E}}=0.
\]
By Lemma \ref{lem:Label-Compatible-Approximation}, applied with $u=w$,
there exists $w_{h}\in\mathcal{H}_{h}$ satisfying
\[
w_{h}\left(\boldsymbol{X}_{l,h}\right)=w\left(\boldsymbol{X}_{l,h}\right)=\boldsymbol{0}
\]
and 
\[
\left\Vert w-w_{h}\right\Vert _{H^{2}\left(\Omega\right)}\lesssim h^{m-2}\left\Vert w\right\Vert _{H^{m}\left(\Omega\right)}.
\]
Hence $w_{h}\in\mathbb{W}_{p_{h}}$, and the limit result follows.\end{proofof}

We are now in a position to prove Theorem \ref{thm:posterior-kernel-convergence}.

\begin{proofof}{thm:posterior-kernel-convergence}

Fix $\boldsymbol{\xi}\in\Omega$. By Lemma \ref{lem:Posterior-Kernel-Reproducing-on-Wph},
\[
c_{h}\left(\cdot,\boldsymbol{\xi}\right)\in\mathbb{W}_{p_{h}},
\]
and
\[
\left<w_{h},c_{h}\left(\cdot,\boldsymbol{\xi}\right)\right>_{\mathscr{D}}=w_{h}\left(\boldsymbol{\xi}\right),\qquad w_{h}\in\mathbb{W}_{p_{h}}.
\]
Taking $w_{h}=c_{h}\left(\cdot,\boldsymbol{\xi}\right)$ gives 
\[
\left\Vert \mathscr{D}c_{h}\left(\cdot,\boldsymbol{\xi}\right)\right\Vert _{L^{2}\left(\Omega\right)}^{2}=c_{h}\left(\boldsymbol{\xi},\boldsymbol{\xi}\right).
\]
By the continuous embedding $H^{2}\left(\Omega\right)\hookrightarrow C\left(\bar{\Omega}\right)$
and Lemma \ref{lem:Uniform-Stability-on-Wph} 
\[
\begin{aligned}\left\Vert \mathscr{D}c_{h}\left(\cdot,\boldsymbol{\xi}\right)\right\Vert _{L^{2}\left(\Omega\right)}^{2} & =c_{h}\left(\boldsymbol{\xi},\boldsymbol{\xi}\right)\\
 & \leqslant\left\Vert c_{h}\left(\cdot,\boldsymbol{\xi}\right)\right\Vert _{C\left(\bar{\Omega}\right)}\\
 & \leqslant C_{\mathrm{emb}}\left\Vert c_{h}\left(\cdot,\boldsymbol{\xi}\right)\right\Vert _{H^{2}\left(\Omega\right)}\\
 & \leqslant C_{\mathrm{emb}}C_{\mathrm{stab}}\left\Vert \mathscr{D}c_{h}\left(\cdot,\boldsymbol{\xi}\right)\right\Vert _{L^{2}\left(\Omega\right)}.
\end{aligned}
\]
It follows that
\[
\left\Vert \mathscr{D}c_{h}\left(\cdot,\boldsymbol{\xi}\right)\right\Vert _{L^{2}\left(\Omega\right)}\leqslant C_{\mathrm{emb}}C_{\mathrm{stab}},
\]
and hence
\begin{equation}
\left\Vert c_{h}\left(\cdot,\boldsymbol{\xi}\right)\right\Vert _{H^{2}\left(\Omega\right)}\leqslant C_{\mathrm{emb}}C_{\mathrm{stab}}^{2}.\label{eq:Posterior-Kernel-Uniform-H2-Bound}
\end{equation}
Thus the family $\left\{ c_{h}\left(\cdot,\boldsymbol{\xi}\right)\right\} $
is bounded in $H^{2}\left(\Omega\right)$. 

Now let $\left\{ h_{j}\right\} $ be an arbitrary sequence satisfying
$h_{j}\downarrow0$. By weak compactness in $H^{2}\left(\Omega\right)$
and the compact embedding into $C\left(\bar{\Omega}\right)$, there
exist a subsequence, not relabeled, and some $v_{\boldsymbol{\xi}}\in H^{2}\left(\Omega\right)\cap C\left(\bar{\Omega}\right)$
such that
\begin{equation}
c_{h_{j}}\left(\cdot,\boldsymbol{\xi}\right)\rightharpoonup v_{\boldsymbol{\xi}}\qquad\text{in }H^{2}\left(\Omega\right),\label{eq:Posterior-Kernel-Subsequence-Weak-Convergence}
\end{equation}
and
\begin{equation}
c_{h_{j}}\left(\cdot,\boldsymbol{\xi}\right)\to v_{\boldsymbol{\xi}}\quad\text{in }C\left(\bar{\Omega}\right).\label{eq:Posterior-Kernel-Subsequence-Convergence}
\end{equation}
We first show that $v_{\boldsymbol{\xi}}\in\mathbb{W}$. 
\begin{itemize}
\item Suppose that $d\geqslant2$, and fix $\boldsymbol{\eta}\in\mathbb{E}$.
By Assumption \ref{assu:Refinement-Design}, there exists $\boldsymbol{\eta}_{j}\in\boldsymbol{X}_{l,h_{j}}$
such that 
\[
\left\Vert \boldsymbol{\eta}-\boldsymbol{\eta}_{j}\right\Vert _{2}\leqslant h_{\boldsymbol{X}_{l,h_{j}},\mathbb{E}}\longrightarrow0.
\]
Lemma \ref{lem:Posterior-Kernel-Reproducing-on-Wph} gives
\[
c_{h_{j}}\left(\boldsymbol{\eta}_{j},\boldsymbol{\xi}\right)=0.
\]
Therefore,
\[
\left|v_{\boldsymbol{\xi}}\left(\boldsymbol{\eta}\right)\right|\leqslant\left|v_{\boldsymbol{\xi}}\left(\boldsymbol{\eta}\right)-v_{\boldsymbol{\xi}}\left(\boldsymbol{\eta}_{j}\right)\right|+\left|v_{\boldsymbol{\xi}}\left(\boldsymbol{\eta}_{j}\right)-c_{h_{j}}\left(\boldsymbol{\eta}_{j},\boldsymbol{\xi}\right)\right|.
\]
The first term tends to zero because $v_{\boldsymbol{\xi}}\in C\left(\bar{\Omega}\right)$,
while the second tends to zero by uniform convergence in Eq. \eqref{eq:Posterior-Kernel-Subsequence-Convergence}.
Hence, 
\[
v_{\boldsymbol{\xi}}\left(\boldsymbol{\eta}\right)=0,\quad\boldsymbol{\eta}\in\mathbb{E}.
\]
Since $\boldsymbol{\eta}$ was arbitrary, $\left.v_{\boldsymbol{\xi}}\right|_{\mathbb{E}}=0.$
\item When $d=1$, Assumption \ref{assu:Refinement-Design} gives $\boldsymbol{X}_{l,h_{j}}=\mathbb{E}$.
Since 
\[
c_{h_{j}}\left(\boldsymbol{X}_{l,h_{j}},\boldsymbol{\xi}\right)=\boldsymbol{0},
\]
uniform convergence in Eq. \eqref{eq:Posterior-Kernel-Subsequence-Convergence}
gives $\left.v_{\boldsymbol{\xi}}\right|_{\mathbb{E}}=0$.
\end{itemize}
Thus,
\[
v_{\boldsymbol{\xi}}\in\mathbb{W}.
\]

We next identify $v_{\boldsymbol{\xi}}$. Let $w\in\mathbb{W}_{\mathrm{reg}}$.
By Lemma \ref{lem:Homogeneous-Recovery-Sequence}, there exists $w_{h_{j}}\in\mathbb{W}_{p_{h_{j}}}$
such that
\begin{equation}
w_{h_{j}}\to w\quad\text{in }H^{2}\left(\Omega\right).\label{eq:Wj-H2-Convergence}
\end{equation}
The discrete reproducing identity in Eq. \eqref{eq:Posterior-Kernel-Reproducing-on-Wph}
gives
\[
\left<c_{h_{j}}\left(\cdot,\boldsymbol{\xi}\right),w_{h_{j}}\right>_{\mathscr{D}}=w_{h_{j}}\left(\boldsymbol{\xi}\right).
\]
Since $\mathscr{D}:H^{2}\left(\Omega\right)\to L^{2}\left(\Omega\right)$
is bounded and point evaluation is continuous on $H^{2}\left(\Omega\right)$,
the weak convergence in Eq. \eqref{eq:Posterior-Kernel-Subsequence-Weak-Convergence}
and the strong convergence in Eq. \eqref{eq:Wj-H2-Convergence} imply
\[
\left<v_{\boldsymbol{\xi}},w\right>_{\mathscr{D}}=w\left(\boldsymbol{\xi}\right),\quad w\in\mathbb{W}_{\mathrm{reg}}.
\]
By Assumption \ref{assu:Dense-Regular-Core}, $\mathbb{W}_{\mathrm{reg}}$
is dense in $\mathbb{W}$ under the $H^{2}\left(\Omega\right)$-norm.
The continuity of the PDE-residual inner product and point evaluation
therefore extends this identity to every $w\in\mathbb{W}$:
\begin{equation}
\left<v_{\boldsymbol{\xi}},w\right>_{\mathscr{D}}=w\left(\boldsymbol{\xi}\right),\qquad w\in\mathbb{W}.\label{eq:Posterior-Kernel-Limit-Reproducing-Identity}
\end{equation}
By the reproducing identity for $G\left(\cdot,\boldsymbol{\xi}\right)$
in Eq. \eqref{eq:Green-Reproducing-Identity}, both $v_{\boldsymbol{\xi}}$
and $G\left(\cdot,\boldsymbol{\xi}\right)$ belong to $\mathbb{W}$
and represent the same point-evaluation functional. Uniqueness of
the Riesz representer gives
\[
v_{\boldsymbol{\xi}}=G\left(\cdot,\boldsymbol{\xi}\right).
\]
Since the sequence $\left\{ h_{j}\right\} $ was arbitrary, every
sequence $h_{j}\downarrow0$ admits a subsequence such that
\[
c_{h_{j}}\left(\cdot,\boldsymbol{\xi}\right)\longrightarrow G\left(\cdot,\boldsymbol{\xi}\right)\quad\text{in }C\left(\bar{\Omega}\right).
\]
A standard subsequence argument therefore yields
\[
\left\Vert c_{h}\left(\cdot,\boldsymbol{\xi}\right)-G\left(\cdot,\boldsymbol{\xi}\right)\right\Vert _{C\left(\bar{\Omega}\right)}\longrightarrow0.
\]
This completes the proof.\end{proofof}

We next derive a uniform diagonal bound for the normalized posterior
kernel $c_{h}\left(\cdot,\cdot\right)$. This bound controls the kernel
diagonal uniformly over $\bar{\Omega}$ and will be used to prove
uniform decay of the posterior variance. 
\begin{lem}
\label{lem:Uniform-Posterior-Kernel-Diagonal-Bound}(Uniform Diagonal
Bound for the Normalized Posterior Kernel) Under the assumptions of
Theorem \ref{thm:Posterior-Mean-Convergence-Rate}, there exists a
constant $C_{\mathrm{diag}}>0$, independent of $h$, such that
\[
\sup_{\boldsymbol{x}\in\bar{\Omega}}c_{h}\left(\boldsymbol{x},\boldsymbol{x}\right)\leqslant C_{\mathrm{diag}}
\]
for all sufficiently small $h$.
\end{lem}
\begin{proofof}{lem:Uniform-Posterior-Kernel-Diagonal-Bound}Fix $\boldsymbol{x}\in\bar{\Omega}$.
By Lemma \ref{lem:Posterior-Kernel-Reproducing-on-Wph}, $c_{h}\left(\cdot,\boldsymbol{x}\right)\in\mathbb{W}_{p_{h}},$
and
\[
\left<w_{h},c_{h}\left(\cdot,\boldsymbol{x}\right)\right>_{\mathscr{D}}=w_{h}\left(\boldsymbol{x}\right),\quad w_{h}\in\mathbb{W}_{p_{h}}.
\]
Taking $w_{h}=c_{h}\left(\cdot,\boldsymbol{x}\right)$ gives
\[
c_{h}\left(\boldsymbol{x},\boldsymbol{x}\right)=\left\Vert \mathscr{D}c_{h}\left(\cdot,\boldsymbol{x}\right)\right\Vert _{L^{2}\left(\Omega\right)}^{2}.
\]
By the embedding $H^{2}\left(\Omega\right)\hookrightarrow C\left(\bar{\Omega}\right)$
and Lemma \ref{lem:Uniform-Stability-on-Wph},
\[
\begin{aligned}\left\Vert \mathscr{D}c_{h}\left(\cdot,\boldsymbol{x}\right)\right\Vert _{L^{2}\left(\Omega\right)}^{2} & =c_{h}\left(\boldsymbol{x},\boldsymbol{x}\right)\\
 & \leqslant\left\Vert c_{h}\left(\cdot,\boldsymbol{x}\right)\right\Vert _{C\left(\bar{\Omega}\right)}\\
 & \leqslant C_{\mathrm{emb}}\left\Vert c_{h}\left(\cdot,\boldsymbol{x}\right)\right\Vert _{H^{2}\left(\Omega\right)}\\
 & \leqslant C_{\mathrm{emb}}C_{\mathrm{stab}}\left\Vert \mathscr{D}c_{h}\left(\cdot,\boldsymbol{x}\right)\right\Vert _{L^{2}\left(\Omega\right)}.
\end{aligned}
\]
If the last norm is zero, the desired bound is immediate. Otherwise,
division gives
\[
\left\Vert \mathscr{D}c_{h}\left(\cdot,\boldsymbol{x}\right)\right\Vert _{L^{2}\left(\Omega\right)}\leqslant C_{\mathrm{emb}}C_{\mathrm{stab}}.
\]
Consequently,
\[
c_{h}\left(\boldsymbol{x},\boldsymbol{x}\right)\leqslant C_{\mathrm{emb}}^{2}C_{\mathrm{stab}}^{2}.
\]
Since the constants are independent of $\boldsymbol{x}$ and $h$,
\[
\sup_{\boldsymbol{x}\in\bar{\Omega}}c_{h}\left(\boldsymbol{x},\boldsymbol{x}\right)\leqslant C_{\mathrm{emb}}^{2}C_{\mathrm{stab}}^{2}.
\]
This completes the proof.\end{proofof}

The preceding uniform diagonal bound now yields the claimed uniform
decay of the posterior variance.

\begin{proofof}{cor:Decay-Predictive-Sigma2}

By the representation of the posterior variance and Lemma \ref{lem:Uniform-Posterior-Kernel-Diagonal-Bound},
\[
\sup_{\boldsymbol{x}\in\bar{\Omega}}\hat{\sigma}_{f,h}^{2}\left(\boldsymbol{x}\right)=\hat{\sigma}_{h}^{2}\sup_{\boldsymbol{x}\in\bar{\Omega}}c_{h}\left(\boldsymbol{x},\boldsymbol{x}\right)\lesssim\hat{\sigma}_{h}^{2}.
\]
Applying Eq. \eqref{eq:Sigma2-Interpretation} at refinement level
$h$ gives
\[
\hat{\sigma}_{h}^{2}=\frac{1}{N_{l,h}}\left\Vert \mathscr{D}\hat{f}_{h}\right\Vert _{L^{2}\left(\Omega\right)}^{2}.
\]
Corollary \ref{cor:PDE-Residual-Convergence-Rate} therefore yields
\[
\begin{aligned}\hat{\sigma}_{h}^{2} & \lesssim\frac{1}{N_{l,h}}h^{2m-4}\left\Vert f\right\Vert _{H^{m}\left(\Omega\right)}^{2}.\end{aligned}
\]
Combining the preceding estimates gives
\[
\sup_{\boldsymbol{x}\in\bar{\Omega}}\hat{\sigma}_{f,h}^{2}\left(\boldsymbol{x}\right)\lesssim\frac{h^{2m-4}}{N_{l,h}}\left\Vert f\right\Vert _{H^{m}\left(\Omega\right)}^{2}.
\]
 This completes the proof. \end{proofof}

\subsection{Proofs of Proposition \ref{Prop:PDE-Gram-Gauss+Matern-polycoeff}
and Corollary \ref{Cor:PDE-Gram-Gauss+Matern-conscoeff}\label{subsec:Proofs-PDE-Gram}}

Before proving Proposition \ref{Prop:PDE-Gram-Gauss+Matern-polycoeff}
and Corollary \ref{Cor:PDE-Gram-Gauss+Matern-conscoeff}, we establish
several auxiliary lemmas.
\begin{lem}
\label{lem:PDE-Gram-Residual-Criterion} (Finite-Dimensional PDE-Residual
Criterion) Suppose that 
\[
\mathscr{D}k\left(\cdot,\boldsymbol{z}_{j}\right)\in L^{2}\left(\Omega\right),\quad j=1,\cdots,p.
\]
Then the PDE-residual Gram matrix $\boldsymbol{A}$ is positive definite
if and only if $\left\{ \mathscr{D}k\left(\cdot,\boldsymbol{z}_{j}\right)\right\} _{j=1}^{p}$
is linearly independent in $L^{2}\left(\Omega\right)$.
\end{lem}
\begin{proofof}{lem:PDE-Gram-Residual-Criterion}

For any $\boldsymbol{c}=\left[c_{1},\cdots,c_{p}\right]^{\top}\in\mathbb{R}^{p}$,
\[
\boldsymbol{c}^{\top}\boldsymbol{Ac}=\sum_{i=1}^{p}\sum_{j=1}^{p}c_{i}c_{j}\left<\mathscr{D}k\left(\cdot,\boldsymbol{z}_{i}\right),\mathscr{D}k\left(\cdot,\boldsymbol{z}_{j}\right)\right>_{L^{2}\left(\Omega\right)}=\left\Vert \sum_{j=1}^{p}c_{j}\mathscr{D}k\left(\cdot,\boldsymbol{z}_{j}\right)\right\Vert _{L^{2}\left(\Omega\right)}^{2}.
\]
Thus the quadratic form $\boldsymbol{c}^{\top}\boldsymbol{Ac}$ vanishes
only at
\[
\sum_{j=1}^{p}c_{j}\mathscr{D}k\left(\cdot,\boldsymbol{z}_{j}\right)=0,\quad\text{in }L^{2}\left(\Omega\right).
\]
Therefore $\boldsymbol{A}\succ0$ if and only if the only such coefficient
vector is $\boldsymbol{c}=\boldsymbol{0}$, which is exactly the $L^{2}\left(\Omega\right)$-linear
independence of the derivative kernel sections $\left\{ \mathscr{D}k\left(\cdot,\boldsymbol{z}_{j}\right)\right\} _{j=1}^{p}$.\end{proofof}
\begin{lem}
\label{lem:PDE-Gram-Polynomial-Exponential-Independence}(Polynomial--Exponential
Linear Independence) Let $\boldsymbol{\lambda}_{1},\cdots,\boldsymbol{\lambda}_{p}\in\mathbb{C}^{d}$
be pairwise distinct, and let $Q_{1},\cdots,Q_{p}$ be polynomials
with complex coefficients. If 
\begin{equation}
\sum_{j=1}^{p}Q_{j}\left(\boldsymbol{\omega}\right)\exp\left(\boldsymbol{\lambda}_{j}^{\top}\boldsymbol{\omega}\right)=0\label{eq:PDE-Gram-Polynomial-Exponential-Identity}
\end{equation}
 for every $\boldsymbol{\omega}$ in a nonempty open subset of $\mathbb{R}^{d}$,
then 
\[
Q_{j}=0,\quad j=1,\cdots,p.
\]
\end{lem}
\begin{proofof}{lem:PDE-Gram-Polynomial-Exponential-Independence}

The function 
\[
F\left(\boldsymbol{\omega}\right)=\sum_{j=1}^{p}Q_{j}\left(\boldsymbol{\omega}\right)\exp\left(\boldsymbol{\lambda}_{j}^{\top}\boldsymbol{\omega}\right)
\]
is real analytic on $\mathbb{R}^{d}$. Since $F$ vanishes on a nonempty
open subset of $\mathbb{R}^{d}$, Eq. \eqref{eq:PDE-Gram-Polynomial-Exponential-Identity}
implies that 
\[
F\left(\boldsymbol{\omega}\right)=0,\quad\forall\boldsymbol{\omega}\in\mathbb{R}^{d}.
\]
Choose $\boldsymbol{v}\in\mathbb{R}^{d}$ such that the scalars 
\[
\mu_{j}=\boldsymbol{\lambda}_{j}^{\top}\boldsymbol{v},\quad j=1,\cdots,p,
\]
are pairwise distinct. This is possible because, for each pair $i\ne j$,
the condition 
\[
\mu_{i}=\mu_{j}
\]
is equivalent to 
\[
\left(\boldsymbol{\lambda}_{i}-\boldsymbol{\lambda}_{j}\right)^{\top}\boldsymbol{v}=0,
\]
and hence the set of $\boldsymbol{v}$ for which $\mu_{1},\cdots,\mu_{p}$
fail to be pairwise distinct is contained in a finite union of hyperplanes. 

Now fix an arbitrary $\boldsymbol{\omega}_{0}\in\mathbb{R}^{d}$,
and restrict Eq. \eqref{eq:PDE-Gram-Polynomial-Exponential-Identity}
to the affine line
\[
\boldsymbol{\omega}=\boldsymbol{\omega}_{0}+t\boldsymbol{v},\quad t\in\mathbb{R}.
\]
This yields
\begin{equation}
\sum_{j=1}^{p}q_{j}\left(t\right)\mathrm{e}^{\mu_{j}t}=0,\quad t\in\mathbb{R},\label{eq:eq:poly-exp-identity-q}
\end{equation}
where $q_{j}\left(t\right)=Q_{j}\left(\boldsymbol{\omega}_{0}+t\boldsymbol{v}\right)\exp\left(\boldsymbol{\lambda}_{j}^{\top}\boldsymbol{\omega}_{0}\right)$
is a univariate polynomial.

Let $M$ be larger than all degrees of the polynomials $\left\{ q_{j}\right\} $.
Fix $j\in\left\{ 1,\cdots,p\right\} $. Apply the differential operator
\[
\mathcal{L}_{j}=\prod_{\ell\ne j}\left(\frac{\mathrm{d}}{\mathrm{d}t}-\mu_{\ell}\right)^{M}
\]
to Eq. \eqref{eq:eq:poly-exp-identity-q}. For each $\ell\ne j$,
the factor $\left(\frac{\mathrm{d}}{\mathrm{d}t}-\mu_{\ell}\right)^{M}$
annihilates the term $q_{\ell}\left(t\right)\mathrm{e}^{\mu_{\ell}t}$,
because
\[
\left(\frac{\mathrm{d}}{\mathrm{d}t}-\mu_{\ell}\right)q_{\ell}\left(t\right)\mathrm{e}^{\mu_{\ell}t}=q_{\ell}^{\prime}\left(t\right)\mathrm{e}^{\mu_{\ell}t}
\]
and therefore 
\[
\left(\frac{\mathrm{d}}{\mathrm{d}t}-\mu_{\ell}\right)^{M}q_{\ell}\left(t\right)\mathrm{e}^{\mu_{\ell}t}=q_{\ell}^{\left(M\right)}\left(t\right)\mathrm{e}^{\mu_{\ell}t}=0.
\]
Hence all terms with index $\ell\ne j$ vanish, and we obtain 

\[
\mathrm{e}^{\mu_{j}t}\prod_{\ell\ne j}\left(\frac{\mathrm{d}}{\mathrm{d}t}+\mu_{j}-\mu_{\ell}\right)^{M}q_{j}\left(t\right)=0.
\]
Since $\mathrm{e}^{\mu_{j}t}\ne0$, it follows that 
\[
\prod_{\ell\ne j}\left(\frac{\mathrm{d}}{\mathrm{d}t}+\mu_{j}-\mu_{\ell}\right)^{M}q_{j}\left(t\right)=0.
\]
For each $\ell\ne j$, we have $\mu_{j}-\mu_{\ell}\ne0$. Therefore
the operator
\[
\frac{\mathrm{d}}{\mathrm{d}t}+\mu_{j}-\mu_{\ell}
\]
is injective on the space of polynomials. It follows that each factor
\[
\left(\frac{\mathrm{d}}{\mathrm{d}t}+\mu_{j}-\mu_{\ell}\right)^{M}
\]
is injective on the space of polynomials, and hence so is their product.
Consequently, $q_{j}=0$. Since $\exp\left(\boldsymbol{\lambda}_{j}^{\top}\boldsymbol{\omega}_{0}\right)\ne0$,
it follows that 
\[
Q_{j}\left(\boldsymbol{\omega}_{0}+t\boldsymbol{v}\right)=0,\quad\forall t\in\mathbb{R}.
\]
Because $\boldsymbol{\omega}_{0}$ was arbitrary, this shows that
$Q_{j}=0$. Since $j$ was arbitrary, the conclusion holds for all
$j=1,\cdots,p$, which completes the proof.\end{proofof}

The next three lemmas are needed only for a one-dimensional Matérn
kernel with a general real smoothness order. The fact that all centers
are interior points is used explicitly.
\begin{lem}
\label{lem:PDE-Gram-Local-Singularity} (Local Singularity Criterion
for Interior Centers) Let $a<z_{1}<\cdots<z_{p}<b$, and let $\left\{ g_{j}\right\} _{j=1}^{p}\subset L^{2}\left(a,b\right)$.
Suppose that $g_{j}$ is real analytic on $\left(a,b\right)\setminus\left\{ z_{j}\right\} $
but has no real-analytic extension across $z_{j}$. Then $\left\{ g_{j}\right\} _{j=1}^{p}$
are linearly independent in $L^{2}\left(a,b\right)$.
\end{lem}
\begin{proofof}{lem:PDE-Gram-Local-Singularity}

Suppose that 
\[
\sum_{j=1}^{p}c_{j}g_{j}=0.
\]
On every open interval avoiding the centers $\left\{ z_{j}\right\} _{j=1}^{p}$,
the sum is real analytic and vanishes almost everywhere, and hence
vanishes identically. Fix $z_{k}$. In a neighborhood of $z_{k}$,
every $g_{j}$ with $j\neq k$ extends analytically across $z_{k}$.
If $c_{k}\neq0$, the relation would express $g_{k}$ as a real-analytic
function across $z_{k}$, a contradiction. Therefore $c_{k}=0$. Since
$k$ is arbitrary, all coefficients vanish.\end{proofof}
\begin{lem}
\label{lem:PDE-Gram-Matern-Noncancellation} (One-Sided Noncancellation
for Non-Half-Integer Matérn Smoothness) Let $\nu>0$ be non-half-integer,
let $K_{\nu}\left(\cdot\right)$ denote the modified Bessel function
of the second kind of order $\nu$, and let $\mathscr{D}$ be a polynomial-coefficient
differential operator. Define $u_{\nu}\left(t\right)=t^{\nu}K_{\nu}\left(t\right)$
for $t>0$. Then, if $\mathscr{D}u_{\nu}$ has a real-analytic extension
across $t=0$, then $\mathscr{D}u_{\nu}=0$ on $\left(0,\infty\right)$.
\end{lem}
\begin{proofof}{lem:PDE-Gram-Matern-Noncancellation}

Let $v_{\nu}\left(t\right)=t^{\nu}I_{\nu}\left(t\right)$, where $I_{\nu}\left(\cdot\right)$
denotes the modified Bessel function of the first kind of order $\nu$.
Since both $I_{\nu}$ and $K_{\nu}$ satisfy the modified Bessel equation
\citep[Section 10]{olverNISTHandbook2010}, both $u_{\nu}$ and $v_{\nu}$
satisfy
\begin{equation}
\mathscr{L}_{\nu}u_{\nu}=0,\quad\mathscr{L}_{\nu}=t^{2}\frac{\mathrm{d}^{2}}{\mathrm{d}t^{2}}+\left(1-2\nu\right)t\frac{\mathrm{d}}{\mathrm{d}t}-t^{2}.\label{eq:PDE-Gram-Bessel-ODE}
\end{equation}
In particular, $\mathscr{L}_{\nu}u_{\nu}=\mathscr{L}_{\nu}v_{\nu}=0$.
We first show that the analyticity of $\mathscr{D}u_{\nu}$ implies
$\mathscr{D}v_{\nu}=0$.

\textbf{Step 1:} $\mathscr{D}v_{\nu}=0$. We distinguish the cases
$\nu\notin\mathbb{Z}$ and $\nu\in\mathbb{Z}$. 

\textbf{Step 1-1:} $\nu\notin\mathbb{Z}$. By the standard power series
for $I_{\nu}$ and the connection formula between $I_{\nu}$ and $K_{\nu}$
\citep[Sections 10.25 and 10.27]{olverNISTHandbook2010}, near $t=0$,
we have 
\begin{align}
u_{\nu}\left(t\right) & =A_{\nu}\left(t^{2}\right)+c_{\nu}t^{2\nu}B_{\nu}\left(t^{2}\right),\nonumber \\
 & =A_{\nu}\left(t^{2}\right)+c_{\nu}v_{\nu}\left(t\right),\label{eq:PDE-Gram-noninteger-expansion}
\end{align}
where
\begin{align*}
A_{\nu}\left(t\right) & =\frac{\pi}{2^{1-\nu}\sin\left(\pi\nu\right)}\sum_{k=0}^{\infty}\frac{1}{k!\Gamma\left(-\nu+k+1\right)}\left(\frac{1}{4}t\right)^{k},\\
B_{\nu}\left(t\right) & =\frac{1}{2^{\nu}}\sum_{k=0}^{\infty}\frac{1}{k!\Gamma\left(\nu+k+1\right)}\left(\frac{1}{4}t\right)^{k},\\
c_{\nu} & =-\frac{\pi}{2\sin\left(\pi\nu\right)}.
\end{align*}
Both $A_{\nu}$ and $B_{\nu}$ are analytic near zero, and 
\[
B_{\nu}\left(0\right)=\frac{1}{2^{\nu}\Gamma\left(\nu+1\right)}\neq0.
\]
Since $\nu$ is not a half-integer and $\nu\notin\mathbb{Z}$, $2\nu\notin\mathbb{Z}$.
Suppose, for contradiction, that $\mathscr{D}v_{\nu}\ne0$. Since
every derivative of $v_{\nu}$ has powers whose exponents differ from
$2\nu$ by integers, there exist $L\in\mathbb{Z}$ and a function
$H$ analytic near zero such that
\[
\mathscr{D}v_{\nu}=t^{2\nu+L}H\left(t\right),\quad H\left(t\right)\ne0.
\]
Let $q\in\mathbb{N}_{0}$ be the order of the zero of $H$ at zero.
Then 
\[
H\left(t\right)=t^{q}\tilde{H}\left(t\right),\quad\tilde{H}\left(0\right)\neq0,
\]
and hence
\[
\mathscr{D}v_{\nu}=t^{2\nu+L+q}\tilde{H}\left(t\right).
\]
Since $2\nu+L+q\notin\mathbb{Z}$, this function cannot have a real-analytic
extension across $t=0$. On the other hand, applying $\mathscr{D}$
to Eq. \eqref{eq:PDE-Gram-noninteger-expansion} gives
\[
\mathscr{D}u_{\nu}=\mathscr{D}\left[A_{\nu}\left(t^{2}\right)\right]+c_{\nu}\mathscr{D}v_{\nu}\left(t\right).
\]
The first term on the right-hand side is analytic near zero. Since
$c_{\nu}\neq0$, it follows that $\mathscr{D}v_{\nu}$ must also have
a real-analytic extension across zero, which is a contradiction. Therefore
\[
\mathscr{D}v_{\nu}=0.
\]

\textbf{Step 1-2: }$\nu\in\mathbb{Z}$. Since $\nu>0$, write $\nu=n$
with $n\in\mathbb{N}$. By the integer-order expansion of $K_{n}$
\citep[Section 10.31]{olverNISTHandbook2010}, near zero we can write
\begin{align*}
u_{n}\left(t\right) & =A_{n}\left(t^{2}\right)+c_{n}t^{2n}\log\left(t\right)B_{n}\left(t^{2}\right),\quad B_{n}\left(0\right)\ne0,\\
 & =A_{n}\left(t^{2}\right)+c_{n}\log\left(t\right)v_{n}\left(t\right)
\end{align*}
where 
\begin{align*}
A_{n}\left(t\right) & =\frac{1}{2}\left(\frac{1}{2}\right)^{-n}\sum_{k=0}^{n-1}\frac{\left(n-k-1\right)!}{k!}\left(-\frac{1}{4}t\right)^{k}+\left(-1\right)^{n+1}\left(-\log2\right)\cdot\left(\frac{1}{2}t\right)^{n}\sum_{k=0}^{\infty}\frac{1}{k!\Gamma\left(n+k+1\right)}\left(\frac{1}{4}t\right)^{k}\\
 & \quad+\left(-1\right)^{n}\frac{1}{2}\left(\frac{1}{2}t\right)^{n}\sum_{k=0}^{\infty}\left[\psi_{\mathrm{dig}}\left(k+1\right)+\psi_{\mathrm{dig}}\left(n+k+1\right)\right]\frac{1}{k!\left(n+k\right)!}\left(\frac{1}{4}t\right)^{k},\\
B_{n}\left(t\right) & =\frac{1}{2^{n}}\sum_{k=0}^{\infty}\frac{1}{k!\Gamma\left(n+k+1\right)}\left(\frac{1}{4}t\right)^{k},\\
c_{n} & =\left(-1\right)^{n+1}.
\end{align*}
Here, $\psi_{\mathrm{dig}}\left(x\right)=\Gamma^{\prime}\left(x\right)/\Gamma\left(x\right)$
denotes the digamma function, and both $A_{n}$ and $B_{n}$ are analytic
near zero. In this case, $v_{n}\left(t\right)$ is also analytic at
zero. It can be derived that
\begin{equation}
\mathscr{D}u_{n}\left(t\right)=\mathscr{D}\left[A_{n}\left(t^{2}\right)\right]+c_{n}\left(\mathscr{D}v_{n}\left(t\right)\right)\log\left(t\right)+c_{n}H_{\mathscr{D}}\left(t\right),\label{eq:PDE-Gram-Dun-decomposition}
\end{equation}
where $H_{\mathscr{D}}$ is meromorphic near zero and contains no
logarithmic term. Suppose again, for contradiction, that $\mathscr{D}v_{n}\neq0$.
Since $v_{n}$ is analytic and $\mathscr{D}$ has polynomial coefficients,
$\mathscr{D}v_{n}$ is analytic. Hence there exist $m\in\mathbb{N}_{0}$
and an analytic function $G$ such that 
\[
\mathscr{D}v_{n}\left(t\right)=t^{m}G\left(t\right),\quad G\left(0\right)\neq0.
\]
The logarithmic term in \eqref{eq:PDE-Gram-Dun-decomposition} therefore
has a first nonzero term of the form $c_{n}G\left(0\right)t^{m}\log\left(t\right)$.
Since $H_{\mathscr{D}}$ has a Laurent expansion containing no logarithmic
terms, its negative-power terms cannot cancel a logarithmic term.
Because $\mathscr{D}u_{n}$ is assumed to have a real-analytic extension
across zero, the principal part of $H_{\mathscr{D}}$ must therefore
vanish. Thus $H_{\mathscr{D}}$ is in fact analytic at zero, and $\left(\mathscr{D}v_{n}\left(t\right)\right)\log\left(t\right)$
must also be analytic there. This is impossible when $\mathscr{D}v_{n}\neq0$,
since the $m$-th derivative of $t^{m}G\left(t\right)\log\left(t\right)$
contains the nonzero term $m!G\left(0\right)\log\left(t\right)$ and
therefore cannot extend continuously to zero. This contradiction shows
that $\mathscr{D}v_{n}=0$. 

Hence, in either case, $\mathscr{D}v_{\nu}=0$.

\textbf{Step 2:} $\mathscr{D}=Q\mathscr{L}_{\nu}$ for some $Q\in\mathbb{C}\left(t\right)\left[\mathrm{d}/\mathrm{d}t\right]$
. The differential-operator ring $\mathbb{C}\left(t\right)\left[\mathrm{d}/\mathrm{d}t\right]$
admits right Euclidean division with respect to the order of a differential
operator \citep[Lemma 2.1]{vanDerPutSinger2003Galois}. Hence we may
divide $\mathscr{D}$ by $\mathscr{L}_{\nu}$ on the right and write
\begin{equation}
\mathscr{D}=Q\mathscr{L}_{\nu}+R,\quad\mathrm{ord}\left(R\right)<2,\label{eq:PDE-Gram-Ore-division}
\end{equation}
for some $Q,R\in\mathbb{C}\left(t\right)\left[\mathrm{d}/\mathrm{d}t\right]$.
Since $\mathscr{D}v_{\nu}=\mathscr{L}_{\nu}v_{\nu}=0$, Eq. \eqref{eq:PDE-Gram-Ore-division}
gives 
\[
0=\mathscr{D}v_{\nu}=Q\mathscr{L}_{\nu}v_{\nu}+Rv_{\nu}=Rv_{\nu}.
\]
Suppose, for contradiction, that $R\ne0$. We show that $R=0$. We
distinguish the cases $\mathrm{ord}\left(R\right)=0$ and $\mathrm{ord}\left(R\right)=1$. 

\textbf{Step 2-1:} $\mathrm{ord}\left(R\right)=0$. Suppose $R=a\left(t\right)\in\mathbb{C}\left(t\right)$.
Then, 
\[
Rv_{\nu}\left(t\right)=a\left(t\right)v_{\nu}\left(t\right)=0.
\]
Since $v_{\nu}$ is not identically zero, this implies $a\left(t\right)=0$,
contradicting $R\neq0$. 

\textbf{Step 2-2:} $\mathrm{ord}\left(R\right)=1$. Suppose 
\[
R=a\left(t\right)\frac{\mathrm{d}}{\mathrm{d}t}+b\left(t\right),\quad a,b\in\mathbb{C}\left(t\right),\quad a\neq0.
\]
Define 
\[
r\left(t\right)=-\frac{b\left(t\right)}{a\left(t\right)}\in\mathbb{C}\left(t\right).
\]
Then $Rv_{\nu}=0$ yields 
\[
v_{\nu}^{\prime}\left(t\right)-r\left(t\right)v_{\nu}\left(t\right)=0,
\]
and hence
\begin{equation}
r\left(t\right)=\frac{v_{\nu}^{\prime}\left(t\right)}{v_{\nu}\left(t\right)}.\label{eq:PDE-Gram-logarithmic-derivative}
\end{equation}
Dividing $\mathscr{L}_{\nu}v_{\nu}=0$ by $t^{2}v_{\nu}$ and using
\[
\frac{v_{\nu}^{\prime\prime}}{v_{\nu}}=r^{\prime}+r^{2},
\]
we obtain the Riccati equation
\begin{equation}
r^{\prime}+r^{2}+\frac{1-2\nu}{t}r-1=0.\label{eq:PDE-Gram-Bessel-Riccati}
\end{equation}
We now show that Eq. \eqref{eq:PDE-Gram-Bessel-Riccati} admits no
rational solution. A standard pole analysis shows that any rational
solution must have the form
\[
r=\varepsilon+\frac{\rho}{t}+\frac{P^{\prime}}{P},\quad\varepsilon\in\left\{ -1,1\right\} ,\rho\in\left\{ 0,2\nu\right\} ,
\]
where $\ensuremath{P\in\mathbb{C}\left[t\right]}$ is a nonzero polynomial.
Let $N=\deg\left(P\right)\in\mathbb{N}_{0}.$ As $t\rightarrow\infty$,
\[
\frac{P^{\prime}\left(t\right)}{P\left(t\right)}=\frac{N}{t}+O\left(t^{-2}\right),
\]
 and hence 
\[
r\left(t\right)=\varepsilon+\frac{\rho+N}{t}+O\left(t^{-2}\right).
\]
Therefore, 
\begin{align*}
r^{\prime}\left(t\right) & =O\left(t^{-2}\right),\\
r^{2}\left(t\right) & =1+\frac{2\varepsilon\left(\rho+N\right)}{t}+O\left(t^{-2}\right),\\
\frac{1-2\nu}{t}r\left(t\right) & =\frac{\varepsilon\left(1-2\nu\right)}{t}+O\left(t^{-2}\right).
\end{align*}
Substituting these expansions into Eq. \eqref{eq:PDE-Gram-Bessel-Riccati}
and comparing the coefficient of $t^{-1}$ gives
\[
2\varepsilon\left(\rho+N\right)+\varepsilon\left(1-2\nu\right)=0.
\]
Since $\varepsilon\neq0$, 
\[
N=\nu-\rho-\frac{1}{2}\notin\mathbb{N}_{0},
\]
which contradicts $N\in\mathbb{N}_{0}$. Therefore, the assumption
$\mathrm{ord}\left(R\right)=1$ is impossible. 

Together with Step 2-1, this shows that the remainder in Eq. \eqref{eq:PDE-Gram-Ore-division}
must satisfy $R=0$. Consequently, 
\[
\mathscr{D}=Q\mathscr{L}_{\nu}.
\]
Finally, since $\mathscr{L}_{\nu}u_{\nu}=0$, we obtain
\[
\mathscr{D}u_{\nu}=Q\mathscr{L}_{\nu}u_{\nu}=0
\]
on $\left(0,\infty\right)$.\end{proofof}
\begin{lem}
\label{lem:PDE-Gram-Matern-1D}(One-Dimensional Matérn Sections) Let
$\Omega=\left(a,b\right)$, let $\boldsymbol{Z}=\left\{ \boldsymbol{z}_{1},\cdots,\boldsymbol{z}_{p}\right\} $
be pairwise distinct points such that $a<z_{1}<\cdots<z_{p}<b$, and
suppose
\[
\mathscr{F}\left[\Phi_{m}\right]\left(\omega\right)=C_{m}\left(1+\omega^{2}\right){}^{-m},\quad m>s+\frac{1}{2}.
\]
Let $\mathscr{D}$ be a polynomial-coefficient operator of order at
most $s$. If every derivative kernel section $\mathscr{D}\Phi_{m}\left(\cdot-z_{j}\right)$
is nonzero in $L^{2}\left(a,b\right)$, then these derivative kernel
sections are linearly independent.
\end{lem}
\begin{proofof}{lem:PDE-Gram-Matern-1D}

Write $\nu=m-1/2>0$. First suppose that $m\notin\mathbb{N}$, so
that $\nu$ is not a half-integer. Each derivative kernel section
is real analytic away from its center. If the PDE residual associated
with the center $z_{j}$ admitted an analytic extension across $z_{j}$,
translate $t=z_{j}+h$, rescale the positive half-line by the Matérn
inverse length scale, and apply Lemma \ref{lem:PDE-Gram-Matern-Noncancellation}
to $h>0$. The PDE residual would vanish on the entire right half-line.
Its assumed analytic extension would then vanish in a two-sided neighborhood
of the center, and analytic continuation on the left branch would
show that it vanishes in $L^{2}\left(a,b\right)$, contradicting the
nonzero-residual assumption. Thus each PDE residual has a nonremovable
analytic singularity at its own center. Lemma \ref{lem:PDE-Gram-Local-Singularity}
gives linear independence.

Now suppose that $m\in\mathbb{N}$. Up to a positive constant,
\begin{equation}
\Phi_{m}\left(h\right)=\mathrm{e}^{-\left|h\right|}p_{m-1}\left(\left|h\right|\right),\label{eq:PDE-Gram-Matern-Half-Integer-Form}
\end{equation}
where $p_{m-1}$ is a polynomial of degree $m-1$. Order the centers
as $a<z_{1}<\cdots<z_{p}<b$, and suppose
\begin{equation}
\sum_{j=1}^{p}c_{j}\mathscr{D}\Phi_{m}\left(\cdot-z_{j}\right)=0\quad\text{in }L^{2}\left(a,b\right).\label{eq:PDE-Gram-Matern-1D-Relation}
\end{equation}
On either side of a center, multiplication by polynomial coefficients
and differentiation preserve the form exponential times polynomial.
Hence, for suitable polynomials $P_{j}^{-}$ and $P_{j}^{+}$,
\[
\mathscr{D}\Phi_{m}\left(t-z_{j}\right)=\mathrm{e}^{-t}P_{j}^{-}\left(t\right),\quad t>z_{j},\quad\mathscr{D}\Phi_{m}\left(t-z_{j}\right)={\rm e}^{t}P_{j}^{+}\left(t\right),\quad t<z_{j}.
\]
Fix $k\in\left\{ 1,\cdots,p\right\} $. On the nonempty interval immediately
to the left of $z_{k}$, Eq. \eqref{eq:PDE-Gram-Matern-1D-Relation}
becomes
\[
\mathrm{e}^{-t}\sum_{j<k}c_{j}P_{j}^{-}\left(t\right)+{\rm e}^{t}\sum_{j\geqslant k}c_{j}P_{j}^{+}\left(t\right)=0.
\]
Polynomial--exponential independence of Lemma \ref{lem:PDE-Gram-Polynomial-Exponential-Independence}
gives
\begin{equation}
\sum_{j<k}c_{j}P_{j}^{-}=0,\quad\sum_{j\geqslant k}c_{j}P_{j}^{+}=0.\label{eq:PDE-Gram-Matern-Left-Polynomials}
\end{equation}
On the nonempty interval immediately to the right of $z_{k}$, the
same argument gives
\begin{equation}
\sum_{j\leqslant k}c_{j}P_{j}^{-}=0,\quad\sum_{j>k}c_{j}P_{j}^{+}=0.\label{eq:PDE-Gram-Matern-Right-Polynomials}
\end{equation}
Subtracting Eqs. \eqref{eq:PDE-Gram-Matern-Left-Polynomials} and
\eqref{eq:PDE-Gram-Matern-Right-Polynomials} yields $c_{k}P_{k}^{-}=c_{k}P_{k}^{+}=0$.
If $c_{k}\neq0$, the derivative kernel section vanishes on nonempty
intervals on both sides of $z_{k}$, and analytic continuation on
each side shows that it vanishes in $L^{2}\left(a,b\right)$, a contradiction.
Hence $c_{k}=0$. Since every center is interior and $k$ is arbitrary,
all coefficients vanish.\end{proofof}

With these auxiliary results in place, we now prove Proposition \ref{Prop:PDE-Gram-Gauss+Matern-polycoeff},
and then derive Corollary \ref{Cor:PDE-Gram-Gauss+Matern-conscoeff}
as its constant-coefficient consequence.

\begin{proofof}{Prop:PDE-Gram-Gauss+Matern-polycoeff}

\textbf{Step 1: Well-definedness of the PDE-residual Gram matrix.
}We first verify that 
\[
\mathscr{D}k\left(\cdot,\boldsymbol{z}_{j}\right)\in L^{2}\left(\Omega\right),\qquad j=1,\cdots,p,
\]
which ensures that  the entries of $\boldsymbol{A}$ are well defined.

For the Gaussian kernel, every derivative of $k\left(\cdot,\boldsymbol{z}_{j}\right)$
is a polynomial times a Gaussian. Since $\Omega$ is bounded and the
coefficients of $\mathscr{D}$ are polynomials, it follows immediately
that
\[
\mathscr{D}k\left(\cdot,\boldsymbol{z}_{j}\right)\in L^{2}\left(\Omega\right),\qquad j=1,\cdots,p.
\]

For the Matérn kernel, let $\left|\boldsymbol{\alpha}\right|\leqslant s$.
By Plancherel's theorem and Eq. \eqref{eq:standard-Matern-spectrum},
\[
\left\Vert \partial^{\boldsymbol{\alpha}}\Phi_{m}\right\Vert _{L^{2}\left(\mathbb{R}^{d}\right)}^{2}=\frac{C_{m,d}^{2}}{\left(2\pi\right)^{d}}\int_{\mathbb{R}^{d}}\left|\boldsymbol{\omega}^{\boldsymbol{\alpha}}\right|^{2}\left(1+\left\Vert \boldsymbol{\omega}\right\Vert _{2}^{2}\right)^{-2m}\mathrm{d}\boldsymbol{\omega}.
\]
As $\left\Vert \boldsymbol{\omega}\right\Vert _{2}\rightarrow\infty$,
after passing to polar coordinates, the radial integrand behaves like
\[
\left\Vert \boldsymbol{\omega}\right\Vert ^{2\left|\boldsymbol{\alpha}\right|-4m+d-1}.
\]
Hence the integral is finite whenever 
\[
2\left|\boldsymbol{\alpha}\right|-4m+d-1<-1\iff4m>2\left|\boldsymbol{\alpha}\right|+d.
\]
Since $\left|\boldsymbol{\alpha}\right|\leqslant s$ and $m>s+d/2$,
this condition holds for every $\left|\boldsymbol{\alpha}\right|\leqslant s$.
Therefore,
\[
\partial^{\boldsymbol{\alpha}}\Phi_{m}\left(\cdot-\boldsymbol{z}_{j}\right)\in L^{2}\left(\mathbb{R}^{d}\right),\quad\left|\boldsymbol{\alpha}\right|\leqslant s.
\]
Because $\Omega$ is bounded, every polynomial coefficient $a_{\boldsymbol{\alpha}}$
is bounded on $\bar{\Omega}$. It follows that
\[
\mathscr{D}\Phi_{m}\left(\cdot-\boldsymbol{z}_{j}\right)\in L^{2}\left(\Omega\right),\qquad j=1,\cdots,p.
\]

Since $\boldsymbol{A}$ is the $L^{2}$-Gram matrix of $\left\{ \mathscr{D}k\left(\cdot,\boldsymbol{z}_{j}\right):j=1,\cdots,p\right\} $,
it is positive definite if and only if $\left\{ \mathscr{D}k\left(\cdot,\boldsymbol{z}_{j}\right):j=1,\cdots,p\right\} $
is linearly independent in $L^{2}\left(\Omega\right)$. 

It remains to prove that
\[
\mathscr{D}k\left(\cdot,\boldsymbol{z}_{j}\right)\ne0\quad\text{for every }j=1,\cdots,p
\]
 is equivalent to the linear independence of $\left\{ \mathscr{D}k\left(\cdot,\boldsymbol{z}_{j}\right):j=1,\cdots,p\right\} $. 

The necessity is immediate. Let $\boldsymbol{e}_{j}\in\mathbb{R}^{p}$
denote the $j$-th standard basis vector, that is 
\[
\boldsymbol{e}_{j}=\left[0,\cdots,0,1,0,\cdots,0\right]^{\top}
\]
with the $1$ in the $j$-th position. If 
\[
\mathscr{D}k\left(\cdot,\boldsymbol{z}_{j_{0}}\right)=0,\quad\text{in }L^{2}\left(\Omega\right)
\]
for some $j_{0}$, then 
\[
\boldsymbol{e}_{j_{0}}^{\top}\boldsymbol{Ae}_{j}=\left<\mathscr{D}k\left(\cdot,\boldsymbol{z}_{j_{0}}\right),\mathscr{D}k\left(\cdot,\boldsymbol{z}_{j}\right)\right>=0,\quad j=1,\cdots,p,
\]
so the $j_{0}$-th row of $\boldsymbol{A}$ is zero. Hence $\boldsymbol{A}$
cannot be positive definite. 

We now prove sufficiency. Suppose that
\begin{equation}
\sum_{j=1}^{p}c_{j}\mathscr{D}k\left(\cdot,\boldsymbol{z}_{j}\right)=0\quad\text{in }L^{2}\left(\Omega\right)\label{eq:polycoeff-residual-relation}
\end{equation}
for some $\boldsymbol{c}=\left[c_{1},\cdots,c_{p}\right]^{\top}\in\mathbb{R}^{p}$.
We shall show that $c_{1}=\cdots=c_{p}=0$. 

\textbf{Step 2: Gaussian kernel.} Define
\[
F\left(\boldsymbol{t}\right)=\sum_{j=1}^{p}c_{j}\mathscr{D}_{\boldsymbol{t}}k\left(\boldsymbol{t},\boldsymbol{z}_{j}\right).
\]
Each term $c_{j}\mathscr{D}_{\boldsymbol{t}}k\left(\boldsymbol{t},\boldsymbol{z}_{j}\right)$
is an entire real-analytic function of $\boldsymbol{t}$, and hence
so is $F$. Since Eq. \eqref{eq:polycoeff-residual-relation} holds
in $L^{2}\left(\Omega\right)$, we have $F=0$ almost everywhere on
$\Omega$. A real-analytic function on a connected open set that vanishes
on a set of positive measure must vanish identically. Therefore
\[
F=0,\quad\text{on }\mathbb{R}^{d}.
\]
Using the Fourier transform identities
\[
\mathscr{F}\left[\partial^{\boldsymbol{\alpha}}u\right]\left(\boldsymbol{\omega}\right)=\left(\mathrm{i}\boldsymbol{\omega}\right)^{\boldsymbol{\alpha}}\hat{u}\left(\boldsymbol{\omega}\right),\quad\mathscr{F}\left[\boldsymbol{t}^{\boldsymbol{\beta}}u\right]\left(\boldsymbol{\omega}\right)=\mathrm{i}^{\left|\boldsymbol{\beta}\right|}\partial^{\boldsymbol{\beta}}\hat{u}\left(\boldsymbol{\omega}\right)
\]
together with
\[
\mathscr{F}\left[k\left(\cdot,\boldsymbol{z}_{j}\right)\right]\left(\boldsymbol{\omega}\right)=\pi^{d/2}\left(\prod_{i=1}^{d}\theta_{i}^{-1}\right)\exp\left(-\frac{1}{4}\left\Vert \boldsymbol{\varTheta}^{-1}\boldsymbol{\omega}\right\Vert _{2}^{2}\right)\mathrm{e}^{-\mathrm{i}\boldsymbol{\omega}^{\top}\boldsymbol{z}_{j}},
\]
we conclude that for each $j$ there exists a polynomial $Q_{j}$
such that
\[
\mathscr{F}\left[\mathscr{D}k\left(\cdot,\boldsymbol{z}_{j}\right)\right]\left(\boldsymbol{\omega}\right)=Q_{j}\left(\boldsymbol{\omega}\right)\exp\left(-\frac{1}{4}\left\Vert \boldsymbol{\varTheta}^{-1}\boldsymbol{\omega}\right\Vert _{2}^{2}\right)\mathrm{e}^{-\mathrm{i}\boldsymbol{\omega}^{\top}\boldsymbol{z}_{j}}.
\]
Taking Fourier transforms in Eq. \eqref{eq:polycoeff-residual-relation}
and dividing by the common nonvanishing Gaussian factor, we obtain
\begin{equation}
\sum_{j=1}^{p}c_{j}Q_{j}\left(\boldsymbol{\omega}\right)\mathrm{e}^{-\mathrm{i}\boldsymbol{\omega}^{\top}\boldsymbol{z}_{j}}=0,\quad\forall\boldsymbol{\omega}\in\mathbb{R}^{d}.\label{eq:gaussian-polyexp-identity}
\end{equation}
Moreover, the assumption $\mathscr{D}k\left(\cdot,\boldsymbol{z}_{j}\right)\ne0$
implies $Q_{j}\ne0$ for every $j$. Since the vectors $-\mathrm{i}\boldsymbol{z}_{j}\in\mathbb{C}^{d}$
are pairwise distinct, Lemma \ref{lem:PDE-Gram-Polynomial-Exponential-Independence}
applied to Eq. \eqref{eq:gaussian-polyexp-identity} yields 
\[
c_{1}=\cdots=c_{p}=0.
\]

\textbf{Step 3: Matérn kernel with $d\geqslant2$.} Define
\[
F\left(\boldsymbol{t}\right)=\sum_{j=1}^{p}c_{j}\mathscr{D}\Phi_{m}\left(\boldsymbol{t}-\boldsymbol{z}_{j}\right).
\]
Each PDE residual $\mathscr{D}\Phi_{m}\left(\cdot-\boldsymbol{z}_{j}\right)$
is real analytic on $\mathbb{R}^{d}\setminus\left\{ \boldsymbol{z}_{j}\right\} $,
and hence $F$ is real analytic on $\mathbb{R}^{d}\setminus\boldsymbol{Z}$.
Since $d\geqslant2$ and $\boldsymbol{Z}$ is finite, both $\Omega\setminus\boldsymbol{Z}$
and $\mathbb{R}^{d}\setminus\boldsymbol{Z}$ are connected. Since
Eq. \eqref{eq:polycoeff-residual-relation} holds in $L^{2}\left(\Omega\right)$,
we have $F=0$ almost everywhere on $\Omega\setminus\boldsymbol{Z}$.
Real analyticity implies 
\[
F=0,\quad\text{on }\Omega\setminus\boldsymbol{Z}.
\]
Since $F$ is real analytic on the connected set $\mathbb{R}^{d}\setminus\boldsymbol{Z}$
and vanishes on the nonempty open set $\Omega\setminus\boldsymbol{Z}$,
the identity theorem implies
\[
F=0,\quad\text{a.e. on }\mathbb{R}^{d}.
\]
Let 
\[
L=\underset{\left|\boldsymbol{\alpha}\right|\leqslant s}{\max}\mathrm{deg}\left(a_{\boldsymbol{\alpha}}\right).
\]
Using again the Fourier transform identities for differentiation and
multiplication by polynomial coefficients, together with Eq. \eqref{eq:standard-Matern-spectrum},
one finds that for each $j$ there exists a polynomial $Q_{j}$ such
that 
\[
\left(1+\left\Vert \boldsymbol{\omega}\right\Vert _{2}^{2}\right)^{m+L}\mathscr{F}\left[\mathscr{D}\Phi_{m}\left(\cdot-\boldsymbol{z}_{j}\right)\right]\left(\boldsymbol{\omega}\right)=Q_{j}\left(\boldsymbol{\omega}\right)\mathrm{e}^{-\mathrm{i}\boldsymbol{\omega}^{\top}\boldsymbol{z}_{j}}.
\]
Summing over $j$ and using $\mathscr{F}\left[F\right]=0$, we obtain
\begin{equation}
\sum_{j=1}^{p}c_{j}Q_{j}\left(\boldsymbol{\omega}\right)\mathrm{e}^{-\mathrm{i}\boldsymbol{\omega}^{\top}\boldsymbol{z}_{j}}=0,\quad\forall\boldsymbol{\omega}\in\mathbb{R}^{d}.\label{eq:matern-polyexp-identity}
\end{equation}
Again, the assumption $\mathscr{D}k\left(\cdot,\boldsymbol{z}_{j}\right)\ne0$
implies $Q_{j}\ne0$ for every $j$. Since the vectors $-\mathrm{i}\boldsymbol{z}_{j}\in\mathbb{C}^{d}$
are pairwise distinct, Lemma \ref{lem:PDE-Gram-Polynomial-Exponential-Independence}
applied to Eq. \eqref{eq:matern-polyexp-identity} yields 
\[
c_{1}=\cdots=c_{p}=0.
\]

\textbf{Step 4: Matérn kernel with $d=1$.} Finally, if $d=1$, then
$\Omega=\left(a,b\right)$ for some $a<b$. After relabeling the pairwise
distinct centers if necessary, we may assume that $a<z_{1}<\cdots<z_{p}<b$.
In this case, the required linear independence follows directly from
Lemma \ref{lem:PDE-Gram-Matern-1D}. Hence again
\[
c_{1}=\cdots=c_{p}=0.
\]

Combining Steps 2--4, the nonvanishing condition implies that the
derivative kernel sections 
\[
\left\{ \mathscr{D}k\left(\cdot,\boldsymbol{z}_{j}\right):j=1,\cdots,p\right\} 
\]
are linearly independent in $L^{2}\left(\Omega\right)$. By Lemma
\ref{lem:PDE-Gram-Residual-Criterion}, the PDE-residual Gram matrix
therefore satisfies $\boldsymbol{A}\succ0$. This proves sufficiency
and completes the proof.\end{proofof}

\begin{proofof}{Cor:PDE-Gram-Gauss+Matern-conscoeff}

By Proposition \ref{Prop:PDE-Gram-Gauss+Matern-polycoeff}, it suffices
to prove that
\[
\mathscr{D}k\left(\cdot,\boldsymbol{z}_{j}\right)\ne0\quad\text{for every }j=1,\cdots,p.
\]
Fix $j\in\left\{ 1,\cdots,p\right\} $. Since $\mathscr{D}$ is constant-coefficient,
we may write
\[
\mathscr{D}=\sum_{\left|\boldsymbol{\alpha}\right|\leqslant s}a_{\boldsymbol{\alpha}}\partial^{\boldsymbol{\alpha}}=p_{\mathscr{D}}\left(\partial\right),
\]
where 
\[
p_{\mathscr{D}}\left(\boldsymbol{\xi}\right)=\sum_{\left|\boldsymbol{\alpha}\right|\leqslant s}a_{\boldsymbol{\alpha}}\boldsymbol{\xi}^{\boldsymbol{\alpha}},\quad\boldsymbol{\xi}^{\boldsymbol{\alpha}}=\xi_{1}^{\alpha_{1}}\cdots\xi_{d}^{\alpha_{d}},
\]
and $\partial=\left[\partial_{1},\cdots,\partial_{d}\right]^{\top}$.
Since $\mathscr{D}\ne0$, the polynomial $p_{\mathscr{D}}$ is not
identically zero.

We first consider the Gaussian kernel. Suppose that
\[
p_{\mathscr{D}}\left(\partial\right)k\left(\cdot,\boldsymbol{z}_{j}\right)=0\quad\text{in }L^{2}\left(\Omega\right).
\]
The function $p_{\mathscr{D}}\left(\partial\right)k\left(\cdot,\boldsymbol{z}_{j}\right)$
is real analytic on $\mathbb{R}^{d}$. Since it vanishes almost everywhere
on the open set $\Omega$, it must vanish identically on $\mathbb{R}^{d}$.
Taking Fourier transforms and using 
\[
\mathscr{F}\left[k\left(\cdot,\boldsymbol{z}_{j}\right)\right]\left(\boldsymbol{\omega}\right)=\pi^{d/2}\left(\prod_{i=1}^{d}\theta_{i}^{-1}\right)\exp\left(-\frac{1}{4}\left\Vert \boldsymbol{\varTheta}^{-1}\boldsymbol{\omega}\right\Vert _{2}^{2}\right)\mathrm{e}^{-\mathrm{i}\boldsymbol{\omega}^{\top}\boldsymbol{z}_{j}},
\]
we obtain 
\[
\mathscr{F}\left[\mathscr{D}k\left(\cdot,\boldsymbol{z}_{j}\right)\right]\left(\boldsymbol{\omega}\right)=\pi^{d/2}\left(\prod_{i=1}^{d}\theta_{i}^{-1}\right)p_{\mathscr{D}}\left(\mathrm{i}\boldsymbol{\omega}\right)\exp\left(-\frac{1}{4}\left\Vert \boldsymbol{\varTheta}^{-1}\boldsymbol{\omega}\right\Vert _{2}^{2}\right)\mathrm{e}^{-\mathrm{i}\boldsymbol{\omega}^{\top}\boldsymbol{z}_{j}}=0,\quad\forall\boldsymbol{\omega}\in\mathbb{R}^{d}.
\]
The Gaussian factor and the exponential factor are both nonzero for
every $\boldsymbol{\omega}$, so
\[
p_{\mathscr{D}}\left(\mathrm{i}\boldsymbol{\omega}\right)=0\quad\forall\boldsymbol{\omega}\in\mathbb{R}^{d}.
\]
Hence $p_{\mathscr{D}}=0$, a contradiction. Therefore, 
\[
\mathscr{D}k\left(\cdot,\boldsymbol{z}_{j}\right)\ne0
\]
for the Gaussian kernel.

We next consider the Matérn kernel $k\left(\cdot,\cdot\right)=\Phi_{m}\left(\cdot-\cdot\right)$.
Suppose first that $d\geqslant2$, and assume that
\[
p_{\mathscr{D}}\left(\partial\right)\Phi_{m}\left(\cdot-\boldsymbol{z}_{j}\right)=0\quad\text{in }L^{2}\left(\Omega\right).
\]
The function $p_{\mathscr{D}}\left(\partial\right)\Phi_{m}\left(\cdot-\boldsymbol{z}_{j}\right)$
is real analytic on $\mathbb{R}^{d}\setminus\left\{ \boldsymbol{z}_{j}\right\} $.
Since it vanishes almost everywhere on the open set $\Omega\setminus\left\{ \boldsymbol{z}_{j}\right\} $,
it vanishes identically on $\Omega\setminus\left\{ \boldsymbol{z}_{j}\right\} $.
Because $\mathbb{R}^{d}\setminus\left\{ \boldsymbol{z}_{j}\right\} $
is connected when $d\geqslant2$, the identity theorem for real-analytic
functions implies that
\[
p_{\mathscr{D}}\left(\partial\right)\Phi_{m}\left(\cdot-\boldsymbol{z}_{j}\right)=0\quad\text{on }\mathbb{R}^{d}\setminus\left\{ \boldsymbol{z}_{j}\right\} .
\]
As $\left\{ \boldsymbol{z}_{j}\right\} $ has Lebesgue measure zero,
it follows that
\[
p_{\mathscr{D}}\left(\partial\right)\Phi_{m}\left(\cdot-\boldsymbol{z}_{j}\right)=0\quad\text{a.e. on }\mathbb{R}^{d}.
\]
Taking Fourier transforms and using Eq. \eqref{eq:standard-Matern-spectrum},
we obtain
\[
p_{\mathscr{D}}\left(\mathrm{i}\boldsymbol{\omega}\right)C_{m,d}\left(1+\left\Vert \boldsymbol{\omega}\right\Vert _{2}^{2}\right)^{-m}\mathrm{e}^{-\mathrm{i}\boldsymbol{\omega}^{\top}\boldsymbol{z}_{j}}=0,\quad\forall\boldsymbol{\omega}\in\mathbb{R}^{d}.
\]
Again, the last two factors are nonzero for every $\boldsymbol{\omega}$,
so 
\[
p_{\mathscr{D}}\left(\mathrm{i}\boldsymbol{\omega}\right)=0\quad\forall\boldsymbol{\omega}\in\mathbb{R}^{d}.
\]
Hence $p_{\mathscr{D}}=0$, a contradiction. Therefore, 
\[
\mathscr{D}\Phi_{m}\left(\cdot-\boldsymbol{z}_{j}\right)\ne0
\]
for $d\geqslant2$ .

It remains to consider the case $d=1$. Then $\Omega=\left(a,b\right)$,
and the interior-center assumption gives $a<z_{j}<b$. Suppose that
\[
p_{\mathscr{D}}\left(\partial\right)\Phi_{m}\left(\cdot-z_{j}\right)=0,\quad\text{in }L^{2}\left(a,b\right).
\]
The function $p_{\mathscr{D}}\left(\partial\right)\Phi_{m}\left(\cdot-z_{j}\right)$
is real analytic on each of the two connected components $\left(-\infty,z_{j}\right)$
and $\left(z_{j},\infty\right)$. Since $z_{j}\in\left(a,b\right)$,
the intervals $\left(a,z_{j}\right)$ and $\left(z_{j},b\right)$
are both nonempty. The assumed $L^{2}\left(a,b\right)$ vanishing
therefore implies that the PDE residual vanishes almost everywhere
on a nonempty open interval in each connected component. By real analyticity
and the identity theorem, 
\[
p_{\mathscr{D}}\left(\partial\right)\Phi_{m}\left(\cdot-z_{j}\right)=0,\quad\text{on }\left(-\infty,z_{j}\right)\cup\left(z_{j},\infty\right).
\]
As $\left\{ z_{j}\right\} $ has Lebesgue measure zero, it follows
that
\[
p_{\mathscr{D}}\left(\partial\right)\Phi_{m}\left(\cdot-z_{j}\right)=0,\quad\text{a.e. on }\mathbb{R}.
\]
Taking Fourier transforms and using Eq. \eqref{eq:standard-Matern-spectrum}
gives
\[
p_{\mathscr{D}}\left(\mathrm{i}\omega\right)C_{m,1}\left(1+\omega^{2}\right)^{-m}\mathrm{e}^{-\mathrm{i}\omega z_{j}}=0,\quad\forall\omega\in\mathbb{R}.
\]
Since $C_{m,1}\left(1+\omega^{2}\right)^{-m}\mathrm{e}^{-\mathrm{i}\omega z_{j}}\ne0$
for every $\omega\in\mathbb{R}$, we obtain
\[
p_{\mathscr{D}}\left(\mathrm{i}\omega\right)=0,\quad\forall\omega\in\mathbb{R}.
\]
Hence $p_{\mathscr{D}}=0$, again contradicting $\mathscr{D}\ne0$.
Therefore, 
\[
\mathscr{D}\Phi_{m}\left(\cdot-\boldsymbol{z}_{j}\right)\ne0
\]
for $d=1$.

Since $j$ was arbitrary, $\mathscr{D}k\left(\cdot,\boldsymbol{z}_{j}\right)\ne0$
holds for all $j=1,\cdots,p$.  Proposition \ref{Prop:PDE-Gram-Gauss+Matern-polycoeff}
implies $\boldsymbol{A}\succ0$, which completes the proof.\end{proofof}

\subsection{Proof of Proposition \ref{Prop:PDE-Gram-Formula-Gauss} \label{subsec:Proof-of-A-Formulas}}

\begin{proofof}{Prop:PDE-Gram-Formula-Gauss}

By the definition of the PDE-residual Gram matrix, 
\begin{equation}
A_{ij}=\sum_{\left|\boldsymbol{\alpha}\right|\leqslant s}\sum_{\left|\boldsymbol{\beta}\right|\leqslant s}a_{\boldsymbol{\alpha}}a_{\boldsymbol{\beta}}\int_{\Omega}\partial^{\boldsymbol{\alpha}}k\left(\boldsymbol{t},\boldsymbol{z}_{i}\right)\partial^{\boldsymbol{\beta}}k\left(\boldsymbol{t},\boldsymbol{z}_{j}\right)\mathrm{d}\boldsymbol{t}.\label{eq:Aij-def}
\end{equation}

First consider the case $\Omega=\mathbb{R}^{d}$. Since every derivative
of the Gaussian kernel is a polynomial multiplied by a Gaussian density,
integration by parts over increasing boxes and passage to the limit
give no boundary contribution. Hence
\begin{equation}
A_{ij}=\sum_{\left|\boldsymbol{\alpha}\right|\leqslant s}\sum_{\left|\boldsymbol{\beta}\right|\leqslant s}\left(-1\right)^{\left|\boldsymbol{\alpha}\right|}a_{\boldsymbol{\alpha}}a_{\boldsymbol{\beta}}\int_{\mathbb{R}^{d}}k\left(\boldsymbol{t},\boldsymbol{z}_{i}\right)\partial^{\boldsymbol{\alpha}+\boldsymbol{\beta}}k\left(\boldsymbol{t},\boldsymbol{z}_{j}\right)\mathrm{d}\boldsymbol{t}.\label{eq:proof-Gauss-fullspace-ibp}
\end{equation}
Let $\boldsymbol{h}=\boldsymbol{z}_{i}-\boldsymbol{z}_{j}$ and $\boldsymbol{u}=\boldsymbol{t}-\boldsymbol{z}_{i}$.
Then
\[
k\left(\boldsymbol{t},\boldsymbol{z}_{i}\right)=\exp\left(-\left\Vert \boldsymbol{\varTheta}\boldsymbol{u}\right\Vert _{2}^{2}\right),\quad k\left(\boldsymbol{t},\boldsymbol{z}_{j}\right)=\exp\left(-\left\Vert \boldsymbol{\varTheta}\left(\boldsymbol{u}+\boldsymbol{h}\right)\right\Vert _{2}^{2}\right).
\]
Differentiation with respect to $\boldsymbol{t}$ in the second kernel
is equivalently differentiation with respect to $\boldsymbol{h}$,
and dominated convergence permits the derivative to pass through the
integral. Hence Eq. \eqref{eq:proof-Gauss-fullspace-ibp} becomes
\begin{equation}
A_{ij}=\sum_{\left|\boldsymbol{\alpha}\right|\leqslant s}\sum_{\left|\boldsymbol{\beta}\right|\leqslant s}\left(-1\right)^{\left|\boldsymbol{\alpha}\right|}a_{\boldsymbol{\alpha}}a_{\boldsymbol{\beta}}\partial_{\boldsymbol{h}}^{\boldsymbol{\alpha}+\boldsymbol{\beta}}F\left(\boldsymbol{h}\right),\label{eq:proof-Gauss-Aij-Fh}
\end{equation}
where 
\begin{equation}
F\left(\boldsymbol{h}\right)=\int_{\mathbb{R}^{d}}\exp\left(-\left\Vert \boldsymbol{\varTheta}\boldsymbol{u}\right\Vert _{2}^{2}\right)\exp\left(-\left\Vert \boldsymbol{\varTheta}\left(\boldsymbol{u}+\boldsymbol{h}\right)\right\Vert _{2}^{2}\right)\mathrm{d}\boldsymbol{u}.\label{eq:proof-Gauss-Fh-def}
\end{equation}
Completing the square in the exponent gives
\begin{equation}
\left\Vert \boldsymbol{\varTheta}\boldsymbol{u}\right\Vert _{2}^{2}+\left\Vert \boldsymbol{\varTheta}\left(\boldsymbol{u}+\boldsymbol{h}\right)\right\Vert _{2}^{2}=2\left\Vert \boldsymbol{\varTheta}\left(\boldsymbol{u}+\frac{\boldsymbol{h}}{2}\right)\right\Vert _{2}^{2}+\frac{1}{2}\left\Vert \boldsymbol{\varTheta}\boldsymbol{h}\right\Vert _{2}^{2}.\label{eq:proof-Gauss-complete-square-fullspace}
\end{equation}
Substituting Eq. \eqref{eq:proof-Gauss-complete-square-fullspace}
into Eq. \eqref{eq:proof-Gauss-Fh-def} and using the change of variables
\[
\boldsymbol{v}=\boldsymbol{u}+\frac{\boldsymbol{h}}{2}
\]
yields
\begin{equation}
\begin{aligned}F\left(\boldsymbol{h}\right) & =\exp\left(-\frac{1}{2}\left\Vert \boldsymbol{\varTheta}\boldsymbol{h}\right\Vert _{2}^{2}\right)\int_{\mathbb{R}^{d}}\exp\left(-2\left\Vert \boldsymbol{\varTheta}\boldsymbol{v}\right\Vert _{2}^{2}\right)\mathrm{d}\boldsymbol{v}\\
 & =C_{\boldsymbol{\theta}}\exp\left(-\frac{1}{2}\left\Vert \boldsymbol{\varTheta}\boldsymbol{h}\right\Vert _{2}^{2}\right),
\end{aligned}
\label{eq:proof-Gauss-Fh-closed}
\end{equation}
where
\[
C_{\boldsymbol{\theta}}=\left(\frac{\pi}{2}\right)^{d/2}\prod_{r=1}^{d}\theta_{r}^{-1}.
\]
Combining Eqs. \eqref{eq:proof-Gauss-Aij-Fh} and \eqref{eq:proof-Gauss-Fh-closed}
gives 

\begin{equation}
A_{ij}=C_{\boldsymbol{\theta}}\sum_{\left|\boldsymbol{\alpha}\right|\leqslant s}\sum_{\left|\boldsymbol{\beta}\right|\leqslant s}\left(-1\right)^{\left|\boldsymbol{\alpha}\right|}a_{\boldsymbol{\alpha}}a_{\boldsymbol{\beta}}\partial_{\boldsymbol{h}}^{\boldsymbol{\alpha}+\boldsymbol{\beta}}\exp\left(-\frac{1}{2}\left\Vert \boldsymbol{\varTheta}\boldsymbol{h}\right\Vert _{2}^{2}\right).\label{eq:proof-Gauss-before-Hermite}
\end{equation}
Let $\mathrm{He}_{n}$ denote the probabilists' Hermite polynomial,
defined by
\[
\frac{\mathrm{d}^{n}}{\mathrm{d}x^{n}}\exp\left(-\frac{x^{2}}{2}\right)=\left(-1\right)^{n}\mathrm{He}_{n}\left(x\right)\exp\left(-\frac{x^{2}}{2}\right).
\]
For a multi-index $\boldsymbol{\nu}$, set
\[
\mathrm{He}_{\boldsymbol{\nu}}\left(\boldsymbol{t}\right)=\prod_{r=1}^{d}\mathrm{He}_{\nu_{r}}\left(t_{r}\right).
\]
Then
\begin{equation}
\partial_{\boldsymbol{h}}^{\boldsymbol{\alpha}+\boldsymbol{\beta}}\exp\left(-\frac{1}{2}\left\Vert \boldsymbol{\varTheta}\boldsymbol{h}\right\Vert _{2}^{2}\right)=\left(-1\right)^{\left|\boldsymbol{\alpha}+\boldsymbol{\beta}\right|}\boldsymbol{\theta}^{\boldsymbol{\alpha}+\boldsymbol{\beta}}\mathrm{He}_{\boldsymbol{\alpha}+\boldsymbol{\beta}}\left(\boldsymbol{\varTheta}\boldsymbol{h}\right)\exp\left(-\frac{1}{2}\left\Vert \boldsymbol{\varTheta}\boldsymbol{h}\right\Vert _{2}^{2}\right).\label{eq:proof-Gauss-Hermite-derivative}
\end{equation}
Substituting Eq. \eqref{eq:proof-Gauss-Hermite-derivative} into Eq.
\eqref{eq:proof-Gauss-before-Hermite} and using
\[
\left(-1\right)^{\left|\boldsymbol{\alpha}\right|}\left(-1\right)^{\left|\boldsymbol{\alpha}+\boldsymbol{\beta}\right|}=\left(-1\right)^{2\left|\boldsymbol{\alpha}\right|+\left|\boldsymbol{\beta}\right|}=\left(-1\right)^{\left|\boldsymbol{\beta}\right|}
\]
gives

\begin{equation}
A_{ij}=C_{\boldsymbol{\theta}}\Pi_{\mathscr{D}}\left(\boldsymbol{h}\right)\exp\left(-\frac{1}{2}\left\Vert \boldsymbol{\varTheta}\boldsymbol{h}\right\Vert _{2}^{2}\right),\label{eq:proof-Gauss-fullspace-final-h}
\end{equation}
where
\begin{equation}
\Pi_{\mathscr{D}}\left(\boldsymbol{h}\right)=\sum_{\left|\boldsymbol{\alpha}\right|\leqslant s}\sum_{\left|\boldsymbol{\beta}\right|\leqslant s}a_{\boldsymbol{\alpha}}a_{\boldsymbol{\beta}}\left(-1\right)^{\left|\boldsymbol{\beta}\right|}\boldsymbol{\theta}^{\boldsymbol{\alpha}+\boldsymbol{\beta}}\mathrm{He}_{\boldsymbol{\alpha}+\boldsymbol{\beta}}\left(\boldsymbol{\varTheta}\boldsymbol{h}\right).\label{eq:proof-Gauss-PiD-def}
\end{equation}
Since $\left|\boldsymbol{\alpha}+\boldsymbol{\beta}\right|\leqslant2s$,
$\Pi_{\mathscr{D}}$ is a multivariate polynomial of degree at most
$2s$. The first formula in Eq. \eqref{eq:PDE-Gram-Formula-Gauss}
follows from Eq. \eqref{eq:proof-Gauss-fullspace-final-h} by substituting
$\boldsymbol{h}=\boldsymbol{z}_{i}-\boldsymbol{z}_{j}$.

Next consider the hyper-rectangular case $\Omega=\prod_{r=1}^{d}\left[a_{r},b_{r}\right]$.
No integration by parts is used in this case, because boundary terms
would generally appear. Let $H_{n}$ denote the physicists' Hermite
polynomial, defined by
\[
\frac{\mathrm{d}^{n}}{\mathrm{d}x^{n}}\exp\left(-x^{2}\right)=\left(-1\right)^{n}H_{n}\left(x\right)\exp\left(-x^{2}\right),
\]
and set
\[
H_{\boldsymbol{\nu}}\left(\boldsymbol{t}\right)=\prod_{r=1}^{d}H_{\nu_{r}}\left(t_{r}\right).
\]
Direct differentiation of the Gaussian kernel gives
\begin{equation}
\partial^{\boldsymbol{\alpha}}k\left(\boldsymbol{t},\boldsymbol{z}\right)=\left(-1\right)^{\left|\boldsymbol{\alpha}\right|}\boldsymbol{\theta}^{\boldsymbol{\alpha}}H_{\boldsymbol{\alpha}}\left(\boldsymbol{\varTheta}\left(\boldsymbol{t}-\boldsymbol{z}\right)\right)k\left(\boldsymbol{t},\boldsymbol{z}\right).\label{eq:proof-Gauss-rect-derivative}
\end{equation}
Substituting Eq. \eqref{eq:proof-Gauss-rect-derivative} into Eq.
\eqref{eq:Aij-def} yields
\begin{equation}
\partial^{\boldsymbol{\alpha}}k\left(\boldsymbol{t},\boldsymbol{z}_{i}\right)\partial^{\boldsymbol{\beta}}k\left(\boldsymbol{t},\boldsymbol{z}_{j}\right)=\left(-1\right)^{\left|\boldsymbol{\alpha}\right|+\left|\boldsymbol{\beta}\right|}\boldsymbol{\theta}^{\boldsymbol{\alpha}+\boldsymbol{\beta}}H_{\boldsymbol{\alpha}}\left(\boldsymbol{\varTheta}\left(\boldsymbol{t}-\boldsymbol{z}_{i}\right)\right)H_{\boldsymbol{\beta}}\left(\boldsymbol{\varTheta}\left(\boldsymbol{t}-\boldsymbol{z}_{j}\right)\right)k\left(\boldsymbol{t},\boldsymbol{z}_{i}\right)k\left(\boldsymbol{t},\boldsymbol{z}_{j}\right).\label{eq:proof-Gauss-rect-product-derivative}
\end{equation}
The domain $\Omega$, the Gaussian kernel, and the multivariate Hermite
polynomials all factorize over coordinates. Therefore Fubini's theorem
gives
\begin{equation}
A_{ij}=\sum_{\left|\boldsymbol{\alpha}\right|\leqslant s}\sum_{\left|\boldsymbol{\beta}\right|\leqslant s}a_{\boldsymbol{\alpha}}a_{\boldsymbol{\beta}}\left(-1\right)^{\left|\boldsymbol{\alpha}\right|+\left|\boldsymbol{\beta}\right|}\boldsymbol{\theta}^{\boldsymbol{\alpha}+\boldsymbol{\beta}}\prod_{r=1}^{d}J_{\alpha_{r},\beta_{r}}^{\left(r\right)}\left(z_{ir},z_{jr}\right),\label{eq:proof-Gauss-rect-factorized}
\end{equation}
where, for $p,q\in\mathbb{N}_{0}$, 
\begin{equation}
J_{p,q}^{\left(r\right)}\left(x,y\right)=\int_{a_{r}}^{b_{r}}H_{p}\left(\theta_{r}\left(t-x\right)\right)H_{q}\left(\theta_{r}\left(t-y\right)\right)\exp\left\{ -\theta_{r}^{2}\left(t-x\right)^{2}-\theta_{r}^{2}\left(t-y\right)^{2}\right\} \mathrm{d}t.\label{eq:proof-Gauss-Jpq-def}
\end{equation}
Eq. \eqref{eq:proof-Gauss-rect-factorized} is the second structural
formula in Eq. \eqref{eq:PDE-Gram-Formula-Gauss}. It remains to give
an explicit evaluation of the one-dimensional moment in Eq. \eqref{eq:proof-Gauss-Jpq-def}.

Fix $r$ and suppress the superscript $\left(r\right)$ by writing
\[
a=a_{r},\quad b=b_{r},\quad\theta=\theta_{r}.
\]
Write
\[
h=x-y,\quad m=\frac{x+y}{2},\quad\eta=\frac{\theta h}{2},\quad L=\sqrt{2}\theta\left(a-m\right),\quad U=\sqrt{2}\theta\left(b-m\right).
\]
The exponent in Eq. \eqref{eq:proof-Gauss-Jpq-def} satisfies 

\begin{equation}
-\theta^{2}\left(t-x\right)^{2}-\theta^{2}\left(t-y\right)^{2}=-2\theta^{2}\left(t-m\right)^{2}-\frac{\theta^{2}h^{2}}{2}.\label{eq:proof-Gauss-rect-complete-square}
\end{equation}
Using Eq. \eqref{eq:proof-Gauss-rect-complete-square} in Eq. \eqref{eq:proof-Gauss-Jpq-def}
and then changing variables by
\[
w=\sqrt{2}\theta\left(t-m\right),
\]
one obtains
\begin{equation}
J_{p,q}\left(x,y\right)=\frac{\exp\left(-\theta^{2}h^{2}/2\right)}{\sqrt{2}\theta}\int_{L}^{U}H_{p}\left(\frac{w}{\sqrt{2}}-\eta\right)H_{q}\left(\frac{w}{\sqrt{2}}+\eta\right)\exp\left(-w^{2}\right)\mathrm{d}w.\label{eq:proof-Gauss-Jpq-w}
\end{equation}
Since the product of the two Hermite polynomials in Eq. \eqref{eq:proof-Gauss-Jpq-w}
is a polynomial in $w$, write
\begin{equation}
H_{p}\left(\frac{w}{\sqrt{2}}-\eta\right)H_{q}\left(\frac{w}{\sqrt{2}}+\eta\right)=\sum_{\ell=0}^{p+q}c_{\ell}^{p,q}\left(\eta\right)w^{\ell}.\label{eq:proof-Gauss-Hermite-product-expansion}
\end{equation}
Combining Eqs. \eqref{eq:proof-Gauss-Jpq-w} and \eqref{eq:proof-Gauss-Hermite-product-expansion}
gives
\begin{equation}
J_{p,q}\left(x,y\right)=\frac{\exp\left(-\theta^{2}h^{2}/2\right)}{\sqrt{2}\theta}\sum_{\ell=0}^{p+q}c_{\ell}^{p,q}\left(\eta\right)I_{\ell}\left(L,U\right),\label{eq:proof-Gauss-Jpq-explicit}
\end{equation}
where 
\begin{equation}
I_{\ell}\left(L,U\right)=\int_{L}^{U}w^{\ell}\exp\left(-w^{2}\right)\mathrm{d}w.\label{eq:proof-Gauss-Iell-def}
\end{equation}
The moments $\left\{ I_{\ell}\left(L,U\right)\right\} $ in Eq. \eqref{eq:proof-Gauss-Iell-def}
are explicit. In particular, 
\[
I_{0}\left(L,U\right)=\frac{\sqrt{\pi}}{2}\left\{ \mathrm{erf}\left(U\right)-\mathrm{erf}\left(L\right)\right\} ,\quad I_{1}\left(L,U\right)=\frac{1}{2}\left\{ \exp\left(-L^{2}\right)-\exp\left(-U^{2}\right)\right\} ,
\]
and, for $\ell\geqslant2$, 
\begin{equation}
I_{\ell}\left(L,U\right)=\frac{1}{2}\left\{ L^{\ell-1}\exp\left(-L^{2}\right)-U^{\ell-1}\exp\left(-U^{2}\right)\right\} +\frac{\ell-1}{2}I_{\ell-2}\left(L,U\right).\label{eq:proof-Gauss-Iell-recursion}
\end{equation}
Here, $\mathrm{erf}\left(x\right)=2\pi^{-1/2}\int_{0}^{x}\exp\left(-t^{2}\right)\mathrm{d}t$
denotes the error function. Thus each $J_{p,q}^{(r)}\left(z_{ir},z_{jr}\right)$
in Eq. \eqref{eq:proof-Gauss-rect-factorized} is reduced to a finite
linear combination of truncated Gaussian moments, with coefficients
obtained from the Hermite-polynomial expansion in Eq. \eqref{eq:proof-Gauss-Hermite-product-expansion}.
Substituting Eqs. \eqref{eq:proof-Gauss-Jpq-explicit} and \eqref{eq:proof-Gauss-Iell-recursion}
into Eq. \eqref{eq:proof-Gauss-rect-factorized} completes the proof.\end{proofof}

\bibliographystyle{asa}
\bibliography{PRIK}

\begin{thebibliography}{20}
\newcommand{\enquote}[1]{``#1''}
\expandafter\ifx\csname natexlab\endcsname\relax\def\natexlab#1{#1}\fi

\bibitem[{Arcang{\'e}li et~al.(2012)Arcang{\'e}li, L{\'o}pez~de Silanes, and Torrens}]{arcangeliExtensionSamplingInequalities2012}
Arcang{\'e}li, R., L{\'o}pez~de Silanes, M.~C., and Torrens, J.~J. (2012), \enquote{Extension of sampling inequalities to Sobolev semi-norms of fractional order and derivative data,} \textit{Numerische Mathematik}, 121, 587--608.

\bibitem[{Arcang{\'e}li and Torrens(2014)}]{ArcangeliTorrens2014Sampling}
Arcang{\'e}li, R. and Torrens, J.~J. (2014), \enquote{Sampling inequalities in Sobolev spaces,} \textit{Journal of Approximation Theory}, 182, 18--28.

\bibitem[{Cheung et~al.(2018)Cheung, Ling, and Schaback}]{cheung$H^2$ConvergenceLeastSquaresKernel2018}
Cheung, K.~C., Ling, L., and Schaback, R. (2018), \enquote{{$H^2$}-Convergence of Least-Squares Kernel Collocation Methods,} \textit{SIAM Journal on Numerical Analysis}, 56, 614--633.

\bibitem[{Dalton et~al.(2026)Dalton, Gao, and Husmeier}]{daltonFiniteelementGaussianProcesses2026}
Dalton, D., Gao, H., and Husmeier, D. (2026), \enquote{Finite-Element Gaussian Processes for the Machine Learning of Steady-State Linear Partial Differential Equations,} \textit{Computer Methods in Applied Mechanics and Engineering}, 451, 118580.

\bibitem[{Dalton et~al.(2024)Dalton, Lazarus, Gao, and Husmeier}]{dalton2024boundary}
Dalton, D., Lazarus, A., Gao, H., and Husmeier, D. (2024), \enquote{Boundary constrained Gaussian processes for robust physics-informed machine learning of linear partial differential equations,} \textit{Journal of Machine Learning Research}, 25, 1--61.

\bibitem[{Davis and Rabinowitz(2007)}]{davis2007methods}
Davis, P.~J. and Rabinowitz, P. (2007), \textit{Methods of numerical integration}, Courier Corporation.

\bibitem[{De~Marchi and Schaback(2010)}]{deMarchiSchabackStability2010}
De~Marchi, S. and Schaback, R. (2010), \enquote{Stability of Kernel-Based Interpolation,} \textit{Advances in Computational Mathematics}, 32, 155--161.

\bibitem[{Doum{\`e}che et~al.(2025)Doum{\`e}che, Bach, Biau, and Boyer}]{doumechePhysicsinformedKernelLearning2025}
Doum{\`e}che, N., Bach, F., Biau, G., and Boyer, C. (2025), \enquote{Physics-Informed Kernel Learning,} \textit{Journal of Machine Learning Research}, 26, 1--39.

\bibitem[{Evans(2010)}]{evansPartialDifferentialEquations2010}
Evans, L.~C. (2010), \textit{Partial Differential Equations}, no. v. 19 in Graduate Studies in Mathematics, Providence, R.I: American Mathematical Society, 2nd ed.

\bibitem[{Fasshauer and McCourt(2012)}]{fasshauer2012stable}
Fasshauer, G.~E. and McCourt, M.~J. (2012), \enquote{Stable evaluation of Gaussian radial basis function interpolants,} \textit{SIAM Journal on Scientific Computing}, 34, A737--A762.

\bibitem[{Fasshauer and Zhang(2007)}]{fasshauer2007choosing}
Fasshauer, G.~E. and Zhang, J.~G. (2007), \enquote{On choosing "optimal" shape parameters for RBF approximation,} \textit{Numerical Algorithms}, 45, 345--368.

\bibitem[{Gilbarg and Trudinger(1998)}]{gilbarg1998elliptic}
Gilbarg, D. and Trudinger, N.~S. (1998), \textit{Elliptic partial differential equations of second order}, vol.~2, Springer.

\bibitem[{H{\"a}rk{\"o}nen et~al.(2023)H{\"a}rk{\"o}nen, {Lange-Hegermann}, and Rai{\c t}{\u a}}]{harkonenGaussianProcessPriors2023}
H{\"a}rk{\"o}nen, M., {Lange-Hegermann}, M., and Rai{\c t}{\u a}, B. (2023), \enquote{Gaussian Process Priors for Systems of Linear Partial Differential Equations with Constant Coefficients,} in \textit{Proceedings of the 40th International Conference on Machine Learning}, PMLR, pp. 12587--12615.

\bibitem[{Li et~al.(2024)Li, Yang, and Wu}]{liParameterInferenceBased2024}
Li, Z., Yang, S., and Wu, C. F.~J. (2024), \enquote{Parameter Inference Based on Gaussian Processes Informed by Nonlinear Partial Differential Equations,} \textit{SIAM/ASA Journal on Uncertainty Quantification}, 12, 964--1004.

\bibitem[{Narcowich et~al.(2005)Narcowich, Ward, and Wendland}]{narcowichSobolevBoundsScatteredZeros2005}
Narcowich, F., Ward, J., and Wendland, H. (2005), \enquote{Sobolev bounds on functions with scattered zeros, with applications to radial basis function surface fitting,} \textit{Mathematics of Computation}, 74, 743--763.

\bibitem[{Narcowich et~al.(2006)Narcowich, Ward, and Wendland}]{narcowichSobolevErrorEstimates2006}
Narcowich, F.~J., Ward, J.~D., and Wendland, H. (2006), \enquote{Sobolev Error Estimates and a Bernstein Inequality for Scattered Data Interpolation via Radial Basis Functions,} \textit{Constructive Approximation}, 24, 175--186.

\bibitem[{Olver et~al.(2010)Olver, Lozier, Boisvert, and Clark}]{olverNISTHandbook2010}
Olver, F. W.~J., Lozier, D.~W., Boisvert, R.~F., and Clark, C.~W. (eds.) (2010), \textit{{NIST} Handbook of Mathematical Functions}, New York: Cambridge University Press.

\bibitem[{Sun and Ling(2025)}]{sunInverseInequalitiesKernel2025}
Sun, Z. and Ling, L. (2025), \enquote{Inverse inequalities for kernel-based approximation on bounded domains and Riemannian manifolds,} \textit{arXiv preprint arXiv:2508.05376}.

\bibitem[{van~der Put and Singer(2003)}]{vanDerPutSinger2003Galois}
van~der Put, M. and Singer, M.~F. (2003), \textit{Galois Theory of Linear Differential Equations}, vol. 328 of \textit{Grundlehren der mathematischen Wissenschaften}, Berlin: Springer.

\bibitem[{Wendland(2004)}]{wendlandScatteredDataApproximation2004}
Wendland, H. (2004), \textit{Scattered Data Approximation}, Cambridge, UK: Cambridge University Press.

\end{thebibliography}


\begin{thebibliography}{40}
\newcommand{\enquote}[1]{``#1''}
\expandafter\ifx\csname natexlab\endcsname\relax\def\natexlab#1{#1}\fi

\bibitem[{Azzimonti et~al.(2015)Azzimonti, Sangalli, Secchi, Domanin, and Nobile}]{azzimonti2015blood}
Azzimonti, L., Sangalli, L.~M., Secchi, P., Domanin, M., and Nobile, F. (2015), \enquote{Blood flow velocity field estimation via spatial regression with PDE penalization,} \textit{Journal of the American Statistical Association}, 110, 1057--1071.

\bibitem[{Batlle et~al.(2025)Batlle, Chen, Hosseini, Owhadi, and Stuart}]{batlleErrorAnalysisKernel2025}
Batlle, P., Chen, Y., Hosseini, B., Owhadi, H., and Stuart, A.~M. (2025), \enquote{Error Analysis of Kernel/GP Methods for Nonlinear and Parametric PDEs,} \textit{Journal of Computational Physics}, 520, 113488.

\bibitem[{Besginow and {Lange-Hegermann}(2022)}]{besginowConstrainingGaussianProcesses2022}
Besginow, A. and {Lange-Hegermann}, M. (2022), \enquote{Constraining Gaussian Processes to Systems of Linear Ordinary Differential Equations,} in \textit{Advances in Neural Information Processing Systems}, vol.~35, pp. 29386--29399.

\bibitem[{Buffa et~al.(2022)Buffa, Gantner, Giannelli, Praetorius, and V{\'a}zquez}]{buffa2022mathematical}
Buffa, A., Gantner, G., Giannelli, C., Praetorius, D., and V{\'a}zquez, R. (2022), \enquote{Mathematical Foundations of Adaptive Isogeometric Analysis: A. Buffa et al.} \textit{Archives of Computational Methods in Engineering}, 29, 4479--4555.

\bibitem[{Cai et~al.(2021)Cai, Wang, Wang, Perdikaris, and Karniadakis}]{cai2021physics}
Cai, S., Wang, Z., Wang, S., Perdikaris, P., and Karniadakis, G.~E. (2021), \enquote{Physics-informed neural networks for heat transfer problems,} \textit{Journal of Heat Transfer}, 143, 060801.

\bibitem[{Chen et~al.(2022)Chen, Chen, Zhang, and Jeff~Wu}]{chenAPIKActivePhysicsInformed2022}
Chen, J., Chen, Z., Zhang, C., and Jeff~Wu, C.~F. (2022), \enquote{APIK: Active Physics-Informed Kriging Model with Partial Differential Equations,} \textit{SIAM/ASA Journal on Uncertainty Quantification}, 10, 481--506.

\bibitem[{Chen et~al.(2021)Chen, Hosseini, Owhadi, and Stuart}]{chenSolvingLearningNonlinear2021}
Chen, Y., Hosseini, B., Owhadi, H., and Stuart, A.~M. (2021), \enquote{Solving and Learning Nonlinear PDEs with Gaussian Processes,} \textit{Journal of Computational Physics}, 447, 110668.

\bibitem[{Davis and Rabinowitz(2007)}]{davis2007methods}
Davis, P.~J. and Rabinowitz, P. (2007), \textit{Methods of numerical integration}, Courier Corporation.

\bibitem[{Doum{\`e}che et~al.(2024)Doum{\`e}che, Bach, Biau, and Boyer}]{doumechePhysicsinformedMachineLearning2024}
Doum{\`e}che, N., Bach, F., Biau, G., and Boyer, C. (2024), \enquote{Physics-Informed Machine Learning as a Kernel Method,} in \textit{Proceedings of Thirty Seventh Conference on Learning Theory}, PMLR, pp. 1399--1450.

\bibitem[{Doum{\`e}che et~al.(2025)Doum{\`e}che, Bach, Biau, and Boyer}]{doumechePhysicsinformedKernelLearning2025}
--- (2025), \enquote{Physics-Informed Kernel Learning,} \textit{Journal of Machine Learning Research}, 26, 1--39.

\bibitem[{Drake et~al.(2021)Drake, Fuselier, and Wright}]{drake2021partition}
Drake, K.~P., Fuselier, E.~J., and Wright, G.~B. (2021), \enquote{A partition of unity method for divergence-free or curl-free radial basis function approximation,} \textit{SIAM Journal on Scientific Computing}, 43, A1950--A1974.

\bibitem[{Evans(2010)}]{evansPartialDifferentialEquations2010}
Evans, L.~C. (2010), \textit{Partial Differential Equations}, no. v. 19 in Graduate Studies in Mathematics, Providence, R.I: American Mathematical Society, 2nd ed.

\bibitem[{Fan et~al.(2018)Fan, Paul, Lee, and Matsuo}]{fan2018modeling}
Fan, M., Paul, D., Lee, T.~C., and Matsuo, T. (2018), \enquote{Modeling tangential vector fields on a sphere,} \textit{Journal of the American Statistical Association}, 113, 1625--1636.

\bibitem[{Farrell and Wendland(2013)}]{farrellRBFMultiscaleCollocation2013}
Farrell, P. and Wendland, H. (2013), \enquote{RBF Multiscale Collocation for Second Order Elliptic Boundary Value Problems,} \textit{SIAM Journal on Numerical Analysis}, 51, 2403--2425.

\bibitem[{Fuselier(2007)}]{fuselierRefinedErrorEstimates2007}
Fuselier, E.~J. (2007), \enquote{Refined Error Estimates for Matrix-Valued Radial Basis Functions,} Ph.D. thesis, Texas A\&M University.

\bibitem[{Graepel(2003)}]{graepelSolvingNoisyLinear2003}
Graepel, T. (2003), \enquote{Solving Noisy Linear Operator Equations by Gaussian Processes: Application to Ordinary and Partial Differential Equations,} in \textit{Proceedings of the Twentieth International Conference on International Conference on Machine Learning}, Washington, DC, USA: AAAI Press, ICML'03, pp. 234--241.

\bibitem[{H{\"a}rk{\"o}nen et~al.(2023)H{\"a}rk{\"o}nen, {Lange-Hegermann}, and Rai{\c t}{\u a}}]{harkonenGaussianProcessPriors2023}
H{\"a}rk{\"o}nen, M., {Lange-Hegermann}, M., and Rai{\c t}{\u a}, B. (2023), \enquote{Gaussian Process Priors for Systems of Linear Partial Differential Equations with Constant Coefficients,} in \textit{Proceedings of the 40th International Conference on Machine Learning}, PMLR, pp. 12587--12615.

\bibitem[{Hughes et~al.(2005)Hughes, Cottrell, and Bazilevs}]{hughes2005isogeometric}
Hughes, T.~J., Cottrell, J.~A., and Bazilevs, Y. (2005), \enquote{Isogeometric analysis: CAD, finite elements, NURBS, exact geometry and mesh refinement,} \textit{Computer methods in applied mechanics and engineering}, 194, 4135--4195.

\bibitem[{Jidling et~al.(2017)Jidling, Wahlstr{\"o}m, Wills, and Sch{\"o}n}]{jidlingLinearlyConstrainedGaussian2017}
Jidling, C., Wahlstr{\"o}m, N., Wills, A., and Sch{\"o}n, T.~B. (2017), \enquote{Linearly Constrained Gaussian Processes,} in \textit{Advances in Neural Information Processing Systems}, Curran Associates, Inc., vol.~30.

\bibitem[{Karniadakis et~al.(2021)Karniadakis, Kevrekidis, Lu, Perdikaris, Wang, and Yang}]{karniadakis2021physics}
Karniadakis, G.~E., Kevrekidis, I.~G., Lu, L., Perdikaris, P., Wang, S., and Yang, L. (2021), \enquote{Physics-informed machine learning,} \textit{Nature Reviews Physics}, 3, 422--440.

\bibitem[{Krishnapriyan et~al.(2021)Krishnapriyan, Gholami, Zhe, Kirby, and Mahoney}]{krishnapriyanCharacterizingPossibleFailure2021}
Krishnapriyan, A., Gholami, A., Zhe, S., Kirby, R., and Mahoney, M. (2021), \enquote{Characterizing Possible Failure Modes in Physics-Informed Neural Networks,} in \textit{Advances in Neural Information Processing Systems}, Curran Associates, Inc., vol.~34, pp. 26548--26560.

\bibitem[{{Lange-Hegermann}(2018)}]{lange-hegermannAlgorithmicLinearlyConstrained2018}
{Lange-Hegermann}, M. (2018), \enquote{Algorithmic Linearly Constrained Gaussian Processes,} in \textit{Advances in Neural Information Processing Systems}, Curran Associates, Inc., vol.~31.

\bibitem[{{Lange-Hegermann}(2021)}]{lange-hegermannLinearlyConstrainedGaussian2021}
--- (2021), \enquote{Linearly Constrained Gaussian Processes with Boundary Conditions,} in \textit{Proceedings of The 24th International Conference on Artificial Intelligence and Statistics}, PMLR, pp. 1090--1098.

\bibitem[{Li et~al.(2024)Li, Yang, and Wu}]{liParameterInferenceBased2024}
Li, Z., Yang, S., and Wu, C. F.~J. (2024), \enquote{Parameter Inference Based on Gaussian Processes Informed by Nonlinear Partial Differential Equations,} \textit{SIAM/ASA Journal on Uncertainty Quantification}, 12, 964--1004.

\bibitem[{Narcowich and Ward(1994)}]{narcowichGeneralizedHermiteInterpolation1994}
Narcowich, F.~J. and Ward, J.~D. (1994), \enquote{Generalized Hermite Interpolation via Matrix-Valued Conditionally Positive Definite Functions,} \textit{Mathematics of Computation}, 63, 661--661.

\bibitem[{Ni et~al.(2025)Ni, Xing, Li, Wang, and Chen}]{ni2025representing}
Ni, X., Xing, J., Li, X., Wang, B., and Chen, B. (2025), \enquote{Representing Flow Fields with Divergence-Free Kernels for Reconstruction,} \textit{Proceedings of the ACM on Computer Graphics and Interactive Techniques}, 8, 1--21.

\bibitem[{Pf{\"o}rtner et~al.(2022)Pf{\"o}rtner, Steinwart, Hennig, and Wenger}]{pfortner2022physics}
Pf{\"o}rtner, M., Steinwart, I., Hennig, P., and Wenger, J. (2022), \enquote{Physics-informed Gaussian process regression generalizes linear PDE solvers,} \textit{arXiv preprint arXiv:2212.12474}.

\bibitem[{Raissi et~al.(2018)Raissi, Perdikaris, and Karniadakis}]{NumericalGaussianProcesses}
Raissi, M., Perdikaris, P., and Karniadakis, G.~E. (2018), \enquote{Numerical Gaussian Processes for Time-Dependent and Nonlinear Partial Differential Equations,} \textit{SIAM Journal on Scientific Computing}, 40, A172--A198.

\bibitem[{Raissi et~al.(2019)Raissi, Perdikaris, and Karniadakis}]{raissi2019physics}
--- (2019), \enquote{Physics-informed neural networks: A deep learning framework for solving forward and inverse problems involving nonlinear partial differential equations,} \textit{Journal of Computational physics}, 378, 686--707.

\bibitem[{Raissi et~al.(2020)Raissi, Yazdani, and Karniadakis}]{raissi2020hidden}
Raissi, M., Yazdani, A., and Karniadakis, G.~E. (2020), \enquote{Hidden fluid mechanics: Learning velocity and pressure fields from flow visualizations,} \textit{Science}, 367, 1026--1030.

\bibitem[{Rasht-Behesht et~al.(2022)Rasht-Behesht, Huber, Shukla, and Karniadakis}]{rasht2022physics}
Rasht-Behesht, M., Huber, C., Shukla, K., and Karniadakis, G.~E. (2022), \enquote{Physics-informed neural networks (PINNs) for wave propagation and full waveform inversions,} \textit{Journal of Geophysical Research: Solid Earth}, 127, e2021JB023120.

\bibitem[{Sangalli(2021)}]{sangalli2021spatial}
Sangalli, L.~M. (2021), \enquote{Spatial regression with partial differential equation regularisation,} \textit{International Statistical Review}, 89, 505--531.

\bibitem[{Santner et~al.(2018)Santner, Williams, and Notz}]{santnerDesignAnalysisComputer2018}
Santner, T.~J., Williams, B.~J., and Notz, W.~I. (2018), \textit{The Design and Analysis of Computer Experiments}, Springer Series in Statistics, New York, NY: Springer New York.

\bibitem[{S{\"a}rkk{\"a}(2011)}]{sarkkaLinearOperatorsStochastic2011}
S{\"a}rkk{\"a}, S. (2011), \enquote{Linear Operators and Stochastic Partial Differential Equations in Gaussian Process Regression,} in \textit{Artificial Neural Networks and Machine Learning -- ICANN 2011}, eds. Honkela, T., Duch, W., Girolami, M., and Kaski, S., Berlin, Heidelberg: Springer, pp. 151--158.

\bibitem[{Sun et~al.(2023)Sun, Sengupta, and Juniper}]{sun2023physics}
Sun, Y., Sengupta, U., and Juniper, M. (2023), \enquote{Physics-informed deep learning for simultaneous surrogate modeling and PDE-constrained optimization of an airfoil geometry,} \textit{Computer Methods in Applied Mechanics and Engineering}, 411, 116042.

\bibitem[{Swiler et~al.(2020)Swiler, Gulian, Frankel, Safta, and Jakeman}]{swilerSURVEYCONSTRAINEDGAUSSIAN2020}
Swiler, L.~P., Gulian, M., Frankel, A.~L., Safta, C., and Jakeman, J.~D. (2020), \enquote{A Survey of Constrained Gaussian Process Regression: Approaches and Implementation Challenges,} \textit{Journal of Machine Learning for Modeling and Computing}, 1.

\bibitem[{Wang et~al.(2022)Wang, Yu, and Perdikaris}]{wang2022and}
Wang, S., Yu, X., and Perdikaris, P. (2022), \enquote{When and why PINNs fail to train: A neural tangent kernel perspective,} \textit{Journal of Computational Physics}, 449, 110768.

\bibitem[{Wendland(2004)}]{wendlandScatteredDataApproximation2004}
Wendland, H. (2004), \textit{Scattered Data Approximation}, Cambridge, UK: Cambridge University Press.

\bibitem[{Xu and Pan(2024)}]{HHDGP2024XuHao}
Xu, H. and Pan, J. (2024), \enquote{HHD-GP: Incorporating Helmholtz-Hodge Decomposition into Gaussian Processes for Learning Dynamical Systems,} in \textit{Advances in Neural Information Processing Systems}, eds. Globerson, A., Mackey, L., Belgrave, D., Fan, A., Paquet, U., Tomczak, J., and Zhang, C., Curran Associates, Inc., vol.~37, pp. 67282--67318.

\bibitem[{Yang et~al.(2019)Yang, Barajas-Solano, Tartakovsky, and Tartakovsky}]{yang2019physics}
Yang, X., Barajas-Solano, D., Tartakovsky, G., and Tartakovsky, A.~M. (2019), \enquote{Physics-informed CoKriging: A Gaussian-process-regression-based multifidelity method for data-model convergence,} \textit{Journal of Computational Physics}, 395, 410--431.

\end{thebibliography}

\end{document}